\documentclass[preprint,12pt]{elsarticle}

\usepackage{amssymb}

\usepackage{color}
\usepackage{bm}
\usepackage{amsmath,amssymb,amsfonts}
\usepackage{multirow}
\usepackage{eepic,mathtips}
\usepackage{lscape}
\usepackage{booktabs}
\usepackage[ruled,vlined]{algorithm2e}
\usepackage{setspace}
\usepackage{subfig}
\usepackage{mathtools}
\DeclareMathOperator*{\argmin}{arg\,min}

\newtheorem{theorem}{Theorem}
\newproof{proof}{Proof}
\newtheorem{corollary}{Corollary}
\newdefinition{remark}{Remark}
\newdefinition{assumption}{Assumption}
\newtheorem{lemma}{Lemma}

\newcommand{\upbm}[2]{{\bm{{#1}}}^{#2}}
\newcommand{\udrbm}[2]{{\bm{{#1}}}_{#2}}
\newcommand{\upbracudr}[3]{{#1}_{{#2}}^{{(#3)}}}
\newcommand{\autonami}[1]{\left\{{#1}\right\}}
\newcommand{\autobrac}[1]{\left({#1}\right)}
\newcommand{\vecbregman}[4]{d_\phi \autobrac{\udrbm{#1}{#2},\udrbm{#3}{#4}}}
\newcommand{\tildebregman}[4]{d_\phi \autobrac{\udrbm{#1}{#2},{\tilde{\bm{{#3}}}}_{#4}}}

\journal{Neurocomputing}

\begin{document}

\begin{frontmatter}



\title{Generalized Dirichlet-process-means for $f$-separable distortion measures}


\author[label1]{Masahiro Kobayashi\fnref{fn1}\corref{cor1}}
\author[label1]{Kazuho Watanabe\fnref{fn2}}
\address[label1]{Department of Computer Science and Engineering, Toyohashi University of Technology, 1-1, Hibarigaoka, Tempaku-cho, Toyohashi 441-8580, Japan}
\cortext[cor1]{Corresponding author}
\fntext[fn1]{Email:kobayashi@lisl.cs.tut.ac.jp}
\fntext[fn2]{Email:wkazuho@cs.tut.ac.jp}
\begin{abstract}
DP-means clustering was obtained as an extension of $K$-means clustering. 
While it is implemented with a simple and efficient algorithm, 
it can estimate the number of clusters simultaneously. 
However, DP-means is specifically designed for the average distortion measure. 
Therefore, it is vulnerable to outliers in data, and can cause large maximum distortion in clusters.
In this work, we extend the objective function of the DP-means to $f$-separable distortion measures and propose a unified learning algorithm to overcome the above problems by selecting the function $f$.
Further, the influence function of the estimated cluster center is analyzed to evaluate the robustness against outliers.
We demonstrate the performance of the generalized method by numerical experiments using real datasets.
\end{abstract}


\begin{highlights}
\item The objective function of DP-means is generalized to $f$-separable distortion measures.
\item It achieves maximum distortion minimization or obtains robustness.
\item Two functions families which have real parameter $\beta$ are introduced.
\item Monotonic decreasing property of the objective function is guaranteed.
\item The influence function is derived, which investigates robustness against outliers.
\end{highlights}


\end{frontmatter}


\section{Introduction \label{sec:intro}}
$K$-means is one of the most popular clustering methods.
This is because its algorithm is simple and can be executed at high speed in linear time with respect to the number of data.
However, it is necessary to specify the number of clusters in advance.
Therefore, it is necessary to apply some heuristics or examine results with multiple cluster numbers.
Although clustering methods that can estimate the number of clusters from given data have been proposed so far, they have problems such as long computation time and many parameters to be tuned.
Affinity propagation \cite{affinity} and Mean shift clustering \cite{mean_shift} are difficult to apply to large scale data because they require the squared order of the number of data to execute the algorithms.
Gamma-clust based on a $\gamma$-divergence has been proposed as a clustering method that is robust against scatted outliers \cite{gamma-clust}. 
It learns means and covariance matrices of $q$-Gaussian mixtures based on $\gamma$-divergence. 
It requires four or five hyper parameters to be set.

In this paper, we focus on the Dirichlet-Process-means (DP-means) clustering \cite{dpmeans} which is a simple algorithm, and an extension of $K$-means that can be performed with a linear order of the number of data.
Historically, to estimate the number of clusters, a learning method of Gaussian mixtures by the nonparametric Bayes approach was proposed \cite{non-parabayse}.
DP-means was obtained as an extension of $K$-means capable of estimating the number of clusters in the limit where the variances of the Gaussian components approach $0$.
DP-means retains the advantages of $K$-means.
It can be executed in linear time with respect to the number of data, is easy to apply to large scale data, and is can be implemented using a simple algorithm.

In an attempt to further speed up DP-means, parallelization can be applied with optimistic concurrency control when a new cluster is created \cite{occ}.
In addition, computation time has been significantly reduced by dividing data into weighted subsets called coresets, although at the expense of accuracy \cite{coreset-dpmeans}.
DP-means specialized for application to large scale genetic data has been devised, and shown to be superior to existing methods from the aspect both of accuracy and efficiency \cite{dace}.
To improve the accuracy of clustering, studies have been made to avoid local minimum solutions \cite{split-dp}.
An extension using Bregman divergence was also given, which introduces an appropriate distance measure when data has a special type such as binary or non-negative integer value \cite{clustering-bregman,dp-bregman}.

From the viewpoint of information theory, the algorithm of DP-means monotonically decreases the average distortion of the training data, whereas the penalty parameter, which controls the number of clusters, has been interpreted as the maximum distortion of the data \cite{mypaper1}.
This motivated us to consider modifying DP-means, where the maximum distortion would be minimized instead of the average distortion.
This problem is also known as the $K$-center problem \cite{k-center}. 
It is also demonstrated in the smallest enclosing ball problem when the number of clusters is one \cite{seb}, and a method to calculate the smallest enclosing Bregman ball with the radius of the smallest enclosing ball measured by Bregman divergence has been studied \cite{bregman-ball}.

DP-means has a problem that it is prone to the influence of outliers because of the nature of the objective function, the average distortion.
Therefore, we extended the objective function of DP-means and invented two objective functions, either of which could bridge maximum distortion and robust distortion measures, and constructed the algorithms to minimize them \cite{mypaper2}.
However, the degree of the robustness against outliers induced by these objective functions has yet to be clarified.

Further, in order to extend the linear distortion measure as the average distortion to nonlinear distortion measures with respect to the distortion of each data point, $f$-separable distortion measures using $f$-mean has been proposed.
In particular, for this distortion measure, the rate-distortion function showing the limit of lossy compression was elucidated \cite{f-separable}.

In this paper, we further capitalize on the above previous work and extend the objective function of DP-means to $f$-separable distortion measure using a monotonically increasing function $f$.
We derive a cluster center update rule for a sufficiently wide class of the function $f$, and show that the objective function monotonically decreases if $f$ is concave.
As concrete examples of the function $f$, we show that two kinds of functions $f$ including a parameter $\beta$ can unify the minimization of robust distortion measures and the minimization of the maximum distortion by adjusting the parameter $\beta$.
Furthermore, we derive the influence function and evaluate the robustness against outliers.
Experiments using real datasets demonstrate that DP-means generalized by the function $f$ improve the performance of the original DP-means.

The paper is organized as follows.
Section \ref{sec:dpm} introduces DP-means.
Section \ref{sec:fgene} generalizes the objective function of DP-means to $f$-separable distortion measures and explains the behavior of the objective function corresponding to the selection of the function $f$.
In Section \ref{sec:const_algorithm}, the generalized DP-means algorithm is constructed based on the generalized objective function.
In Section \ref{sec:if}, we derive the influence function and evaluate robustness against outliers.
In Section \ref{sec:experiments}, we present the results of numerical experiments using real datasets to demonstrate the effectiveness of the generalized DP-means.
In Section \ref{sec:tbd}, we discuss further modification of the objective function in terms of the pseudo-distance.
Finally, Section \ref{sec:conclusion} concludes this paper.
\begin{figure*}[htbp]
\centering
	\includegraphics[scale=0.4]{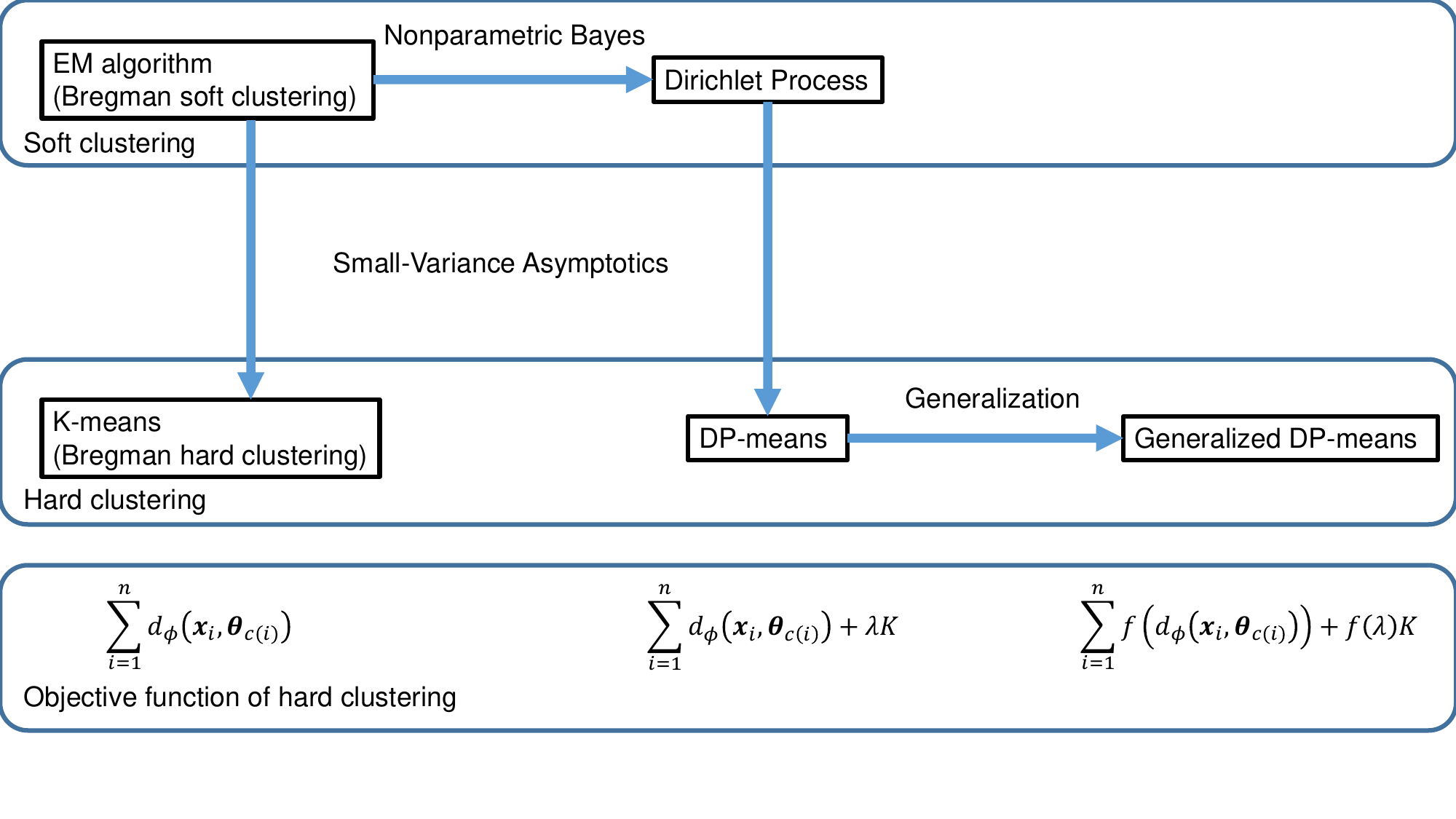}
		\caption{Relations between related clustering methods and the generalized DP-means \label{fig:graphical_abst} }
\vspace{-10pt}
\end{figure*}
\section{DP-means \label{sec:dpm}}
$K$-means is one of the most popular clustering methods, is a simple algorithm, and can be executed at high speed in linear time with respect to the number of data.
However, it is necessary to specify the number of clusters in advance.
DP-means can estimate the number of clusters and retains the advantages of $K$-means.
Figure \ref{fig:graphical_abst} schematically explains the difference between $K$-means, original DP-means and generalized DP-means.
The generalized DP-means will be explained in next section.
$K$-means, a.k.a. Bregman hard clustering, is obtained in the limit of Expectation-Maximization algorithm for the mixture of regular exponential family distributions, a.k.a. Bregman soft clustering, as the variances of component distributions tend to $0$ \cite{clustering-bregman}. 
Precisely, while $K$-means minimizes the sum of squared distance, Bregman hard clustering minimizes the sum of Bregman divergence, the resulting algorithm is identical to $K$-means \cite{clustering-bregman}. 
Thus, we refer to the Bregman hard clustering as $K$-means. 
On the other hand, introducing the Dirichlet-process (DP) prior for the cluster assignments in the mixture of regular exponential family distributions, we obtain the DP mixture model \cite{non-parabayse}. 
The DP-means is obtained as the same small-variance limit as above from the DP mixture model \cite{dpmeans, dp-bregman}.

DP-means requires data $\upbm{x}{n}=\autonami{\udrbm{x}{1},\dots,\udrbm{x}{n}}$ and penalty parameter $\lambda$ as inputs.
Suppose that each data point is $L$-dimensional, $\udrbm{x}{i}=(\upbracudr{x}{i}{1}, \cdots, \upbracudr{x}{i}{L})^{\rm T}\in \mathbb{R}^L$.
The algorithm of DP-means is basically the same as $K$-means.
Let $\autonami{\udrbm{\theta}{1}, \dots, \udrbm{\theta}{K}}$ be cluster centers.
DP-means executes a calculation of the cluster centers and assigns the data point to clusters until the following objective function converges:
\begin{align}
L(\autonami{\udrbm{\theta}{k}}_{k=1}^K, \autonami{c(i)}_{i=1}^n) = \sum_{i=1}^n \vecbregman{x}{i}{\theta}{c(i)} + \lambda K. \label{eq:original}
\end{align}
Here, $c\left(i\right) \triangleq \argmin_k {\vecbregman{x}{i}{\theta}{k}}$ denotes the cluster label of the data point $\udrbm{x}{i}$.
However, the following two points are different from $K$-means.
DP-means is initialized with one cluster, $K=1$.
When the pseudo-distance between a data point and its nearest cluster center is greater than the penalty parameter $\lambda$, a new cluster is created.
In other words, a new cluster is generated when the following is satisfied:
\begin{align}
\vecbregman{x}{i}{\theta}{c(i)} > \lambda . \label{eq:clusterAdd}
\end{align}
Also, the cluster center $\udrbm{\theta}{k}$ is calculated as the average of data points assigned to the $k$-th cluster,
\begin{align*}
\udrbm{\theta}{k} = \frac{\sum_{i=1}^n r_{ik}\bm{x}_i}{\sum_{j=1}^n r_{jk}}  .
\end{align*}
Here, $r_{ik}$ is given by
\[
r_{ik} = \begin{cases}
           1 & (c(i) = k), \\
           0 & (c(i) \neq k) .
         \end{cases}
\]

In this paper, we assume the Bregman divergence as the pseudo-distance, which generalizes the squared distance.
Specifically, the Bregman divergence is the pseudo-distance determined from a differentiable strictly convex function $\phi: \mathbb{R}^L\to\mathbb{R}$ as
\begin{align}
d_\phi \autobrac{\bm{x},\bm{\theta}} \triangleq \phi(\bm{x}) - \phi(\bm{\theta}) -  \langle \bm{x}-\bm{\theta},\bm{\nabla}\phi(\bm{\theta})\rangle, \label{eq:def_bregman}
\end{align}
where $\bm{\nabla}\phi$ represents the gradient vector of $\phi$ and $\langle \cdot , \cdot \rangle$ is the inner product.
The Bregman divergence satisfies non-negativity and identity of indiscernibles in distance axioms.
Moreover, the Bregman divergences are related to the probability distributions in the regular exponential family, bijectively.
For example, if data are of a particular type such as a binary or non-negative integer, the Bregman divergences corresponding to the Bernoulli and Poisson distributions are used as the more suitable distance measures than the usual squared distance for real numbers \cite{clustering-bregman,dp-bregman}.

Under a fixed number of clusters, the only difference between the $K$-means and original DP-means is the initialization.
$K$-means is initialized with random assignment of clusters while in the DP-means, the order of data can be considered as the random initialization.

The average distortion and the maximum distortion are defined by
\begin{align*}
\frac{1}{n} \sum_{i=1}^n \vecbregman{x}{i}{\theta}{c(i)},  \\
\max_{1\leq i \leq n} \vecbregman{x}{i}{\theta}{c(i)}, \nonumber 
\end{align*}
respectively.
Note that the minimization of the objective function \eqref{eq:original} with respect to $\udrbm{\theta}{k}$ is equivalent to that of the average distortion.
On the other hand, the penalty parameter $\lambda$, which determines the number of clusters, can be interpreted as the maximum distortion \cite{mypaper1}.
Therefore, we can consider the maximum distortion as the measure for cluster increment.
\section{Generalized objective function \label{sec:fgene}}
\subsection{Generalization with $f$-separable distortion measures \label{sec:fmean}}
In this paper, we propose an objective function that generalizes the objective function of DP-means to $f$-separable distortion measures as follows:
\begin{align}
L_f(\autonami{\udrbm{\theta}{k}}_{k=1}^K, \autonami{c(i)}_{i=1}^n) = \sum_{i=1}^n f\autobrac{\vecbregman{x}{i}{\theta}{c(i)}} + f\autobrac{\lambda}K . \label{eq:generalized_f}
\end{align}
As we will discuss in Section \ref{sec:monotonically}, this objective function is guaranteed to decrease monotonically with respect to $\autonami{\udrbm{\theta}{k}}_{k=1}^K$ and $\autonami{c(i)}_{i=1}^n$.
In this paper, we assume that the function $f: \mathbb{R}_+ \rightarrow \mathbb{R}$ is a differentiable and continuous monotonically increasing, where $\mathbb{R}_+$ is the set of non-negative real numbers.
The argument is given by the Bregman divergence or the penalty parameter $\lambda$.
In particular, we consider the following three types:
\begin{itemize}
 \setlength{\leftskip}{1.0cm}
 \item[$\nearrow$ \ :] linear,
 \item[\uzarrow \ :] concave,
 \item[\szarrow \ :] convex.
\end{itemize}
If the function $f (z)$ is a differentiable and strictly monotonically increasing function, an inverse function $f^{-1}(z)$ exists, and \eqref{eq:generalized_f} can be normalized to a distortion measure as $f$-mean \cite{f-mean}.
The $f$-separable distortion measures using this inverse function $f^{-1}(z)$ correspond to that in the literature \cite{f-separable}.
The generalized objective function \eqref{eq:generalized_f} can be monotonically transformed with the inverse function $f^{-1}$.
Therefore, minimizing \eqref{eq:generalized_f} is equivalent to minimizing the $f$-mean,
\begin{align*}
f^{-1}\left( \frac{1}{n+K} \left[\sum_{i=1}^n f\autobrac{\vecbregman{x}{i}{\theta}{c(i)}} + f\autobrac{\lambda}K\right]  \right).  
\end{align*}

Table \ref{table:func_property} summarizes the behavior of the objective function, its monotonic improvement property, and the calculation order required to execute the learning algorithm.
If $f$ is linear ($\nearrow$), it becomes the original objective function \eqref{eq:original}, which corresponds to the average distortion in the original DP-means.
If $f$ is convex (\szarrow), a distortion with a larger pseudo-distance value tends to be minimized. 
In particular, the faster the function $f$ diverges to infinity, the more the objective function approaches the maximum distortion.
Conversely, If $f$ is concave (\uzarrow), a distortion with a smaller pseudo-distance value will be prioritized to be minimized.
That is, the influence of data points far from other data points, such as outliers, is weakened.
In other words, there is a trade-off between the maximum distortion, which is the maximum radius of the cluster, and the robustness against outliers.
Robustness against outliers is explained more in detail in Section \ref{sec:if}.

\begin{table}[tbp]\scriptsize
\begin{center} 
\caption{Algorithm behavior corresponding to function $f$.}
  \begin{tabular}{lllll} \toprule
  $f(z)$ & $f^{-1}(z)$ & behavior & monotonic decrease & calculation order\\ \midrule
  $\nearrow$ & $\nearrow$ &average distortion minimization & yes &  $\bm{O}(Ln)$\\
  \uzarrow & \szarrow & robustness against outliers & yes &  $\bm{O}(Ln)$\\ 
  \szarrow & \uzarrow & approaches the maximum distortion minimization & *1 & *2 \\ \bottomrule
  \end{tabular}
\\ *1: A gradient descent optimization method is required. \\ *2: The order depends on the applied gradient descent optimization.
\label{table:func_property}
\end{center}
\end{table}

Eguchi and Kano generalized the likelihood of a probabilistic model to ${\rm \Psi}$-likelihood using a convex function like \eqref{eq:generalized_f}, and devised a ${\rm \Psi}$-estimator as a robust estimator against outliers \cite{psi-div}.
The ${\rm \Psi}$-estimator focuses on the following two points.
The first is to obtain robustness against outliers.
We consider a wide class of functions, not only the robustness against outliers but also the maximum distortion minimization within our focus.
The second point is to guarantee the unbiased estimation equation.
For this, a bias correction term, whose calculation is complicated in general, is included in the objective function.
The ${\rm \Psi}$-likelihood assumes a probabilistic model, whereas in this study only the Bregman divergence is assumed.
As we will discuss, the update rule of the cluster center derived from the combination of the function $f$ and Bregman divergence enables us to execute the learning algorithm in the linear order on the number of the training data as the original DP-means.

\subsection{Examples of function $f$}
\setcounter{footnote}{0}
In this subsection, we show two concrete examples of functions with a parameter $\beta$.
When the parameter $\beta$ is changed, the generalized objective function changes its behavior as average distortion, maximum distortion, or robust distortion measures.
\subsubsection{Power mean objective}
For the function
\begin{align}
f(z)=\frac{1}{\beta}\left[(z+a)^\beta-1\right],  \label{eq:f_pow}
\end{align}
the corresponding $f$-mean is given by\footnotemark
\begin{align}
\left[\frac{1}{n}\sum_{i=1}^n \autobrac{z_i+a}^\beta\right]^{\frac{1}{\beta}}-a . \label{eq:pow}
\end{align}
The first term of \eqref{eq:pow} is called power mean.
The parameters are $\beta \in \mathbb{R}$, $a\geq0$.
Parameter $\beta$ determines the effect of the objective function, and parameter $a$ is introduced to avoid an algorithmic disadvantage.
The disadvantage is that the cluster center is not updated when it overlaps the nearest data point.
If $a>0$, this problem does not occur. 
If $a=0$, some algorithmic modification is required.
We will explain the details in Section \ref{sec:problem}.
Table \ref{table:pow} shows the characteristics of the objective function \eqref{eq:pow} and the corresponding function $f$ for different choices of $\beta$.

\footnotetext{
It can also be expressed as $\ln_{1-\beta} (z+a)=\frac{1}{\beta}[(z+a)^\beta-1]$ by using Tsallis $q$-function $\ln_q (z) \triangleq \frac{z^{1-q}-1}{1-q}$, for which $\ln(z) = \lim_{q\to1} \ln_q (z)$ \cite{tsallis_book}.}
As shown in Table \ref{table:pow}, the behavior of the objective function varies around $\beta = 1$: it shows robust characteristics when $\beta <1$, and the smaller the $\beta$, the smaller the influence of outliers.
When $\beta>1$, the larger the value of $\beta$, the closer the objective function approaches the maximum distortion. In particular, $\beta\to\infty$ implies maximum distortion minimization.
When the Bregman divergence is the squared distance, the objective function using \eqref{eq:pow} with $\beta>0$ and $a=0$ corresponds to the objective function derived in the framework of MAP-based Asymptotic Derivations (MAD-Bayes) \cite{mad-bayes}, in which the generalized Gaussian distribution is assumed as the component of the nonparametric mixture model.
The proof for this is in \ref{sec:proof_gene_gauss}.
Similarly, assuming a deformed $t$-distribution as the component of a nonparametric mixture model, the same objective function when $\beta=0$ and $a> 0$ is obtained.
The proof is in \ref{sec:proof_t_dist}.

\subsubsection{Log-sum-exp objective}
For the function
\begin{align}
f(z)=\frac{1}{\beta-1}\left[\exp\autobrac{(\beta-1)z}-1\right],   \label{eq:f_exp}
\end{align}
the corresponding $f$-mean is given by\footnotemark
\begin{align}
\frac{1}{\beta-1}\ln\left[\frac{1}{n} \sum_{i=1}^n \exp\autobrac{(\beta-1)z_i}\right] . \label{eq:lse}
\end{align}
Equation \eqref{eq:lse} is a differentiable approximation of the maximum value function when $\beta=2$, known as the log-sum-exp function \cite{convex}.
As in the case of the power mean, $\beta \in \mathbb{R}$ determines the characteristics of the objective function as a parameter.
Table \ref{table:lse} shows the characteristics of the objective function using \eqref{eq:lse} and the corresponding function $f$ for different choices of $\beta$.

Table \ref{table:lse} also shows how the objective function behaves differently around $\beta=1$, as in the case of the power mean.
It becomes robust when $\beta <1$, and approaches the maximum distortion when $\beta> 1$.
In particular, when $\beta\to\infty$, its limit is maximum distortion.
In addition, \eqref{eq:lse} corresponds to the objective function for the estimation of mixture models in \cite{entropic_risk} when the variance of each component approaches $0$.

\footnotetext{
It can also be expressed as $\ln_{2-\beta} \autobrac{\exp(z)}=\frac{\exp\autobrac{(\beta-1)z}-1}{\beta-1}$.}
\begin{landscape}
\begin{table*}[tb]
\begin{center}
\caption{Behavior of the power mean objective and corresponding function $f$.}
\scalebox{0.8}[1]{
\begin{tabular}{|c|c|c|c|c|c|}
  \hline \hline
  $\beta$ & $f^{-1}\autobrac{\frac{1}{n}\sum_{i=1}^n f(z_i)}$ & \multicolumn{2}{|c|}{$f(z)$} & $f{\,'}(z)$ & behavior \\
  \hline \hline
  $\beta=1$ & $\frac{1}{n} \sum_{i=1}^n z_i$ & $z+a-1$ & $\nearrow$ & 1 & average distortion minimization \\ \hline
  $\beta=0$ &  $\autobrac{\prod_{i=1}^n (z_i+a)}^{\frac{1}{n}}-a$ &$\ln (z+a)$ & \multirow{3}{*}{\uzarrow}  & $\frac{1}{z+a}$ & \multirow{3}{*}{robustness against outliers} \\  \cline{1-3} \cline{5-5}
  $-\infty<\beta<0$ & \multirow{3}{*}{$\autobrac{\frac{1}{n}\sum_{i=1}^n (z_i+a)^\beta}^\frac{1}{\beta}-a$}  & \multirow{3}{*}{$\frac{1}{\beta}[(z+a)^\beta-1]$} &  & \multirow{3}{*}{$(z+a)^{\beta-1}$} &  \\ \cline{1-1}  
  $0<\beta<1$ &  & &  & &  \\ \cline{1-1} \cline{4-4} \cline{6-6}
  $1<\beta<\infty$ &  & & \szarrow &  & approaches the maximum distortion minimization \\ \hline
  $\beta\to\infty$ & $\max_{1\leq i \leq n} z_i$ &  &  &  & maximum distortion minimization \\ \hline
\end{tabular}}
\label{table:pow}
\end{center}
\begin{center}
\caption{Behavior of the log-sum-exp objective and corresponding function $f$.}
\scalebox{0.75}[1]{
\begin{tabular}{|c|c|c|c|c|c|}
  \hline \hline
  $\beta$ & $f^{-1}\autobrac{\frac{1}{n}\sum_{i=1}^n f(z_i)}$ & \multicolumn{2}{|c|}{$f(z)$} & $f{\,'}(z)$ & behavior \\
  \hline \hline
  $\beta=1$ & $\frac{1}{n} \sum_{i=1}^n z_i$ & $z$ & $\nearrow$ & 1 & average distortion minimization \\ \hline
  $-\infty<\beta<1$ & \multirow{2}{*}{$\frac{1}{\beta-1}\ln\autobrac{\frac{1}{n} \sum_{i=1}^n \exp\autobrac{(\beta-1)z_i}}$} & \multirow{2}{*}{$\frac{\exp\autobrac{(\beta-1)z}-1}{\beta-1}$} & \uzarrow & \multirow{2}{*}{$\exp\autobrac{(\beta-1)z}$} & robustness against outliers \\ \cline{1-1} \cline{4-4} \cline{6-6}
  $1<\beta<\infty$ &  & &  \szarrow & & approaches the maximum distortion minimization \\ \hline
  $\beta\to\infty$ & $\max_{1\leq i \leq n} z_i$ &  &  &  & maximum distortion minimization \\ \hline
\end{tabular}}
\label{table:lse}
\end{center}
\end{table*}
\end{landscape}

\section{Construction of generalized algorithm \label{sec:const_algorithm}}
In this section, we construct a generalized DP-means algorithm based on the objective function proposed in Section \ref{sec:fgene}.
First, we derive the update rule of the cluster center in Section \ref{sec:update_rule}.
Second, we show that the objective function decreases monotonically with the derived update rule in Section \ref{sec:monotonically}.
Third, we discuss minor problems in the execution of the algorithm and offer solutions to them in Section \ref{sec:problem}.
Finally, we discuss the execution time of the generalized algorithm in Section \ref{sec:comp2}.
The generalized algorithms are constructed from the original DP-means by replacing the update rule of cluster centers and the objective function used for convergence diagnosis (Algorithm \ref{extAlg}).
This algorithm differs only to the original algorithm in the update rule of the cluster center, and the computation time required for execution is of linear order with respect to the number of data.

\subsection{Derivation of update rules \label{sec:update_rule}}
The updated equations for the cluster centers are derived from the stationary conditions when the gradient of the cluster center $\udrbm{\theta}{k}$ of the generalized objective function \eqref{eq:generalized_f} is $\bm{0}$. 
Here, $f{\,'}$ represents the derivative of the function $f$.
Thus, the update rule of the cluster center is 
\begin{align}
\udrbm{\theta}{k} = \frac{\sum_{i=1}^n w_{ik} \udrbm{x}{i}}{\sum_{j=1}^n w_{jk}}, \label{eq:update_rule}
\end{align}
\begin{align}
w_{ik} = r_{ik}f{\,'}\autobrac{\vecbregman{x}{i}{\theta}{k}} , \label{eq:weight}
\end{align}
which is the weighted mean of $\udrbm{x}{i}$ weighted by $f{\,'}$ with the Bregman divergence as its argument.
However, the objective function monotonically decreases by this update rule \eqref{eq:update_rule} only when the function $f$ is concave (or linear).
If function $f$ is convex, the cluster center is updated by gradient descent optimization such as the steepest gradient or Newton's method and so on.
The update rule with Newton's method is put in \ref{sec:newton_update}.

\begin{algorithm}[tb]
\caption{Generalized DP-means for $f$-separable distortion measures}
\label{extAlg}
\setstretch{0.9}
\DontPrintSemicolon

\nl \KwIn{$\upbm{x}{n}=\autonami{\udrbm{x}{1},\dots,\udrbm{x}{n}},\:\: \lambda, \:\:$ generic function $f$}
\nl \KwOut{$\{\bm{\theta}_k\}_{k=1}^K, \:\: \{c(i)\}_{i=1}^n, \:\: K$}\nl$K=1, \:\: \bm{\theta}_1=\frac{1}{n} \sum_{i=1}^{n} \bm{x}_i, \:\: c\left(i\right)=1~(i=1,\dots, n)$\;
\nl\Repeat{the decrease of $\bar{L}_f\left(\bm{\theta}_1\right)$ becomes smaller than the threshold $\delta$}{
  \nl calculate $w_{ik}$ by \eqref{eq:weight} ~$({i}=1,\dots, n)$\;
  \nl update $\bm{\theta}_1$ by \eqref{eq:update_rule}\;
}
\Repeat{\eqref{eq:generalized_f} converges}{ 
  \nl\For{$i=1$ \KwTo $n$}{
    \nl $d_{k} =  {d_\phi\left(\bm{x}_i, \bm{\theta}_{k}\right)}~({k}=1,\dots, K)$ \;
    \nl \If{$\min_k d_k>\lambda$ }{
      \nl $K=K+1$ \;
      \nl $c(i)=K, \:\: \bm{\theta}_K=\bm{x}_i$ \;
    }
    \nl \Else{
       \nl $c(i)=\argmin_{k} d_k$ \;
    }
  }
  \nl\For{$k=1$ \KwTo $K$}{
    \nl\Repeat{the decrease of $\bar{L}_f\left(\bm{\theta}_k\right)$ becomes smaller than the threshold $\delta$}{
      \nl calculate $w_{ik}$ by \eqref{eq:weight} ~$({i}=1,\dots, n)$\;
      \nl update $\bm{\theta}_k$ by \eqref{eq:update_rule} \;
    }
  }
}
\end{algorithm}

\subsection{Guarantee of monotonic decreasing property \label{sec:monotonically}}
The DP-means algorithm iterates the cluster center updating step and the assignments of data points to clusters.
Because of the monotonic decreasing property of the objective function in each step, the algorithm converges within finite iterations.
Even in the generalized objective function \eqref{eq:generalized_f}, the cluster assignment step is same as the original DP-means.
Therefore, only the monotonic decreasing property of the objective function in the cluster center updating step is considered.
In the following, we explain the two cases, concave and convex.

\subsubsection{Concave case}
The following theorem applies when the function $f$ is concave.
\begin{theorem}\label{teiri:monot}
If the function $f$ is concave, the updating of the cluster center  using \eqref{eq:update_rule} monotonically decreases the objective function for general Bregman divergence.
\end{theorem}
\begin{proof}
{\rm
We show that the objective function \eqref{eq:generalized_f} monotonically decreases when the $k$-th cluster center $\udrbm{\theta}{k}$ is newly updated to $\tilde{\bm{\theta}}_k$ by \eqref{eq:update_rule}.
More specifically, we prove that, $\bar{L}_f(\udrbm{\theta}{k})\geq \bar{L}_f(\tilde{\bm{\theta}}_k) $,  where $\bar{L}_f(\udrbm{\theta}{k})$ is the sum of the terms related to $\udrbm{\theta}{k}$ in \eqref{eq:generalized_f}:
\begin{align*}
\bar{L}_f({\bm{\theta}}_k) = \sum_{i=1}^n r_{ik} f\autobrac{\vecbregman{x}{i}{\theta}{k}} .
\end{align*}
Here, the tangent line $y$ of $f(z)$ at the point $a$ is expressed by the following equation:
\begin{align*}
y=f{\,'}(a)(z-a)+f(a) .
\end{align*}
Furthermore, since $y\geq f(z)$ follows from the concavity of the function $f$, the following inequality holds:
\begin{align}
f(a)-f(z)\geq f{\,'}(a)(a-z) . \label{eq:tangent_inquality}
\end{align}
From this inequality, the following holds:
\begin{align}
\MoveEqLeft[1]
\bar{L}_f(\udrbm{\theta}{k})-\bar{L}_f(\tilde{\bm{\theta}}_k) \nonumber
\\={}& \sum_{i=1}^n r_{ik} \left[f\autobrac{\vecbregman{x}{i}{\theta}{k}}-f\autobrac{\tildebregman{x}{i}{\theta}{k}}\right]  \nonumber
\\ \geq{}& \sum_{i=1}^n r_{ik} f{\,'}\autobrac{\vecbregman{x}{i}{\theta}{k}} \left[\vecbregman{x}{i}{\theta}{k} - \tildebregman{x}{i}{\theta}{k}\right]  \nonumber
\\={}& \sum_{i=1}^n r_{ik} f{\,'}\autobrac{\vecbregman{x}{i}{\theta}{k}} \left[ \phi\autobrac{\tilde{\bm{\theta}}_k} - \phi\autobrac{\bm{\theta}_k} -\bm{\nabla}\phi\autobrac{\bm{\theta}_k}\autobrac{\bm{x}_i-\bm{\theta}_k} +  \bm{\nabla}\phi\autobrac{\tilde{\bm{\theta}}_k}\autobrac{\bm{x}_i-\tilde{\bm{\theta}}_k} \right] \label{eq:use_update}
\\={}& d_\phi \autobrac{\tilde{\bm{\theta}}_k, \bm{\theta}_k} \sum_{i=1}^n r_{ik}f{\,'}\autobrac{\vecbregman{x}{i}{\theta}{k}}  \geq 0 , \nonumber
\end{align}
where
\begin{align*}
\tilde{\bm{\theta}}_k \sum_{i=1}^n r_{ik} f{\,'}\autobrac{\vecbregman{x}{i}{\theta}{k}} = \sum_{i=1}^n r_{ik} f{\,'}\autobrac{\vecbregman{x}{i}{\theta}{k}} \bm{x}_i
\end{align*}
derived from \eqref{eq:update_rule} and \eqref{eq:weight} was used in \eqref{eq:use_update}.
\qed
}
\end{proof}
The following corollary immediately follows from Theorem \ref{teiri:monot}.
\begin{corollary}
When the objective function is constructed by the power mean \eqref{eq:pow} or the log-sum-exp function \eqref{eq:lse}, the following holds.
When $\beta\leq1$, the updating of the cluster center  using \eqref{eq:update_rule} monotonically decreases the objective function for general Bregman divergence.
\end{corollary}
If the function $f$ satisfies $f(0)>-\infty$, the algorithm converges within finite iterations.
Let $t$ be the number of times the $k$-th cluster center has been updated, and denote it by $\bm{\theta}_k^{(t)}$.
The corresponding objective function is $\bar{L}_f\left(\bm{\theta}_k^{(t)}\right)$.
The objective function sequence $\left\{\bar{L}_f\left(\bm{\theta}_k^{(t)}\right)\right\}_{t=0}^\infty$ converges to a finite value, because it has monotonic decreasing property (Theorem \ref{teiri:monot} ) and lower bounded, that is, $\bar{L}_f\autobrac{\bm{\theta}_k^{(t)}}\geq \sum_{i=1}^n r_{ik} f(0)>-\infty$.
Therefore, the following holds:
\begin{align*}
\lim_{t\rightarrow\infty} \bar{L}_f\left(\bm{\theta}_k^{(t)}\right)-\bar{L}_f\left(\bm{\theta}_k^{(t+1)}\right)=0.
\end{align*}
In other words, 
\begin{align*}\forall\delta>0, \exists t^*\in \mathbb{N}: \bar{L}_f\autobrac{\bm{\theta}_k^{(t^*)}}- \bar{L}_f\autobrac{\bm{\theta}_k^{(t^*+1)}} < \delta.
\end{align*}
That is, the algorithm converges in finite iterations for threshold $\delta$.
The proof of Theorem \ref{teiri:monot} is a generalization of the monotonic decreasing property of ei-means ($\varepsilon=0$) proposed in \cite{ei-means}, which corresponds to the case where $f(z)=\sqrt{z}$ and $\vecbregman{x}{}{\theta}{}=\|\bm{x}-\bm{\theta}\|^2$.
The proof of Theorem \ref{teiri:monot} is also interpreted by the Majorization-Minimization algorithm \cite{mm-alg}.

\subsubsection{Convex case \label{sec:convex_case}}
The function $f$ is convex when the objective function \eqref{eq:generalized_f} exhibits a characteristic close to the maximum distortion minimization.
If the cluster center is updated by \eqref{eq:update_rule}, the value of the objective function can oscillate and may not decrease monotonically.
Therefore, it is necessary to apply a gradient descent optimization such as the steepest descent method or Newton's method.
When the pseudo-distance is the general Bregman divergence, the problem of calculating the cluster centers is not generally a convex optimization problem, so the gradient is not necessarily in the descent direction.
However, monotonic decreasing property is also guaranteed for the Bregman divergence in general by using the algorithm that updates to the descending direction of the gradient like the modified Newton's method.
In particular, when symmetry is satisfied among distance axioms such as the squared distance, the problem of calculating the cluster centers is reduced to a convex optimization problem, so the gradient direction is always in the descent direction \cite{bregman-ball}\cite[Section 3.2]{convex}.
When the function $f$ is convex, the calculation time in the case of the steepest descent method is $\bm{O}(Ln)$, and Newton's method with Cholesky decomposition is $\bm{O}(L^2n)$, where $L$ is the dimension of data points.
In this case, there is no change in the linear order with respect to the number of data.

\subsection{Problem and solution\label{sec:problem}}
When the objective function shows robustness against outliers, that is, when the function $f$ is concave, if a cluster center overlaps a data point, subsequent updates are not performed.
In the assignment step of DP-means, a new cluster center is generated exactly on the data point satisfying \eqref{eq:clusterAdd}.
We can see that a data point and a cluster center frequently overlap.
We will now show the condition where cluster center updating does not take place and offer a solution.

When the function $f$ is a concave and satisfies 
\begin{align}
\lim_{z\to0} f{\,'}(z)=\infty , \label{eq:no_update}
\end{align}
if a cluster center overlaps a data point, updating of the cluster center does not occur.
It is assumed that one point in $\upbm{x}{n}=\autonami{\udrbm{x}{1},\dots,\udrbm{x}{n}}$ overlaps the cluster center $\udrbm{\theta}{k}$.
That is, $\udrbm{x}{i^*}=\udrbm{\theta}{k} \iff \vecbregman{x}{i^*}{\theta}{k}=0$.
Here, if the function $f$ satisfies \eqref{eq:no_update}, 
\begin{align*}
\frac{f{\,'}\autobrac{\vecbregman{x}{i}{\theta}{k}}} {f{\,'}\autobrac{\vecbregman{x}{i^*}{\theta}{k}} } =
\begin{cases}
    1 & (i=i^*), \\
    0 & (i\neq i^*),
  \end{cases}
\end{align*}
holds. 
Therefore, we have
\begin{align*}
\udrbm{\theta}{k} = \frac{\sum_{i=1}^n \frac{w_{ik}} {f{\,'}\autobrac{\vecbregman{x}{i^*}{\theta}{k}} } \udrbm{x}{i} }{\sum_{j=1}^n \frac{w_{jk}}{f{\,'}\autobrac{\vecbregman{x}{i^*}{\theta}{k}}}}
=\udrbm{x}{i^*} ,
\end{align*}
which means that the cluster center does not move from the data point.

Next, we examine the case where function $f$ satisfies \eqref{eq:no_update} and the cluster center overlaps the nearest data point with new data points additionally assigned to the cluster.
Here, updating can be performed by shifting the cluster center from the data point to some point such as the average point, and applying the update rule.

In the case of the power mean \eqref{eq:f_pow}, when $a=0$, since \eqref{eq:no_update} is satisfied, updating of the cluster center does not occur.
Therefore, in this case, the procedure described above is required.
However, in the case of the power mean \eqref{eq:f_pow} with $a>0$, \eqref {eq:no_update} is not satisfied, so the update of the cluster center is performed without the above procedure.
However, it must be smaller than pseudo-distance values.
Similarly, in the case of the log-sum-exp function \eqref{eq:f_exp}, the cluster center can be updated without the above procedure because \eqref{eq:no_update} is not satisfied.

\subsection{Computational time \label{sec:comp2}}
Although the computational complexity of the generalized DP-means is $\bm{O}(n)$ as discussed in Section \ref{sec:convex_case}, its actual clustering time is proportional to the number of outer loops times the number of inner loops. Furthermore, the former depends on the estimated number of clusters. The latter is required for non-linear $f$ and it generally increases as $f$ goes away from the linear function. Hence, the generalized DP-means requires more clustering time than the original DP-means in general. This means that there is a trade-off between the clustering time and robustness.
\section{Analysis of influence function \label{sec:if}}
The influence function is one of the indicators of the robustness against outliers and shows how much the estimator is influenced by the contamination of a small number of outliers.
In this section, we derive the influence function and evaluate how robust the generalized objective function is against outliers.
\subsection{Influence function of general function $f$ \label{sec:influence}}
Derivation of the influence function in this subsection is based on the derivation of the influence function of the total Bregman divergence \cite{total-bregman,information-geometry} (see also Section \ref{sec:tbd}).
We consider the influence function when an outlier is mixed in the $k$-th cluster.
However, since the influence of mixed outliers is independent for each cluster, in the derivation of subsequent influence functions, the subscript of $\udrbm{\theta}{k}$ is omitted and expressed as $\bm{\theta}$.
Suppose that the $k$-th cluster contains $m$ samples of data $\upbm{x}{m}=\autonami{\udrbm{x}{1}, \dots, \udrbm{x}{m}}$, and the cluster center estimated from $\upbm{x}{m}$ is $\bm{\theta}$.
When an outlier $\upbm{x}{*}$ is mixed into the data $\upbm{x}{m}$, a new cluster center $\tilde{\bm{\theta}}$ is calculated for the data including outliers $\upbm{x}{m+1}=\autonami{\udrbm{x}{1}, \dots, \udrbm{x}{m}, \upbm{x}{*}}$.
If we let $\delta\bm{\eta}$ be the difference between the estimator $\tilde{\bm{\theta}}$ including outliers and the estimator $\bm{\theta}$ without outliers,
\begin{align*}
\tilde{\bm{\theta}} -\bm{\theta} =  \delta\bm{\eta} .
\end{align*}
The influence function is defined by
\begin{align}
\bm{{\rm IF}}(\upbm{x}{*}) = m\cdot\delta \bm{\eta} . \label{eq:influence}
\end{align}
The influence function \eqref{eq:influence} defined by a finite sample is also specifically called a sensitivity curve in the field of robust statistics \cite{robust}.
Then, the new cluster center $\tilde{\bm{\theta}}$ minimizes
\begin{align}
\frac{1}{m+1} \sum_{i=1}^m f\autobrac{\tildebregman{x}{i}{\theta}{}} + \frac{1}{m+1} f\autobrac{d_\phi \autobrac{\bm{x}^*, \tilde{\bm{\theta}}}} . \label{eq:outlier_loss}
\end{align}
Now we compute the first three terms of the Taylor expansion around the old cluster center $\bm{\theta}$, which minimizes $\sum_{i=1}^m f\autobrac{\vecbregman{x}{i}{\theta}{}}$.
Therefore, the first derivative with respect to $\delta\bm{\eta}$ of the first term of \eqref{eq:outlier_loss} becomes $\sum_{i=1}^m \bm{\nabla}f\autobrac{\vecbregman{x}{i}{\theta}{}}=0$.
The second derivative with respect to $\delta\bm{\eta}$ of the second term of \eqref{eq:outlier_loss} becomes very small when $m$ is large because 
\begin{align*}
&\frac{1}{m+1}\delta\bm{\eta}^{\mathrm{T}} \bm{\nabla}\bm{\nabla} f\autobrac{d_\phi \autobrac{\bm{x}^*, \bm{\theta}}}\delta\bm{\eta}
\\=& \frac{1}{m+1}\frac{1}{m^2}\bm{{\rm IF}}^{\mathrm{T}}(\upbm{x}{*}) \bm{\nabla}\bm{\nabla}f\autobrac{d_\phi \autobrac{\bm{x}^*, \bm{\theta}}} \bm{{\rm IF}}(\upbm{x}{*}) 
=O(m^{-3}),
\end{align*}
where \eqref{eq:influence} was used.
Here, $\bm{\nabla}$ expresses the gradient with respect to $\bm{\theta}$.
From the above arguments, the term related to $\delta\bm{\eta}$ is given as follows:
\begin{align}
&\frac{1}{m+1} \left[\frac{1}{2}\delta\bm{\eta}^{\mathrm{T}} \bm{{\rm G}}\delta\bm{\eta}
+ \bm{\nabla}f\autobrac{d_\phi \autobrac{\bm{x}^*, \bm{\theta}}}\delta\bm{\eta}\right] , \label{eq:taylor}\\
&\bm{{\rm G}} = \sum_{i=1}^m \bm{\nabla}\bm{\nabla}f\autobrac{\vecbregman{x}{i}{\theta}{}} . \nonumber
\end{align}
The $\delta\bm{\eta}$ that minimizes \eqref{eq:taylor} is solved by completing the square with respect to $\delta\bm{\eta}$.
Then, the influence function in \eqref{eq:influence} is given by
\begin{align}
\bm{{\rm IF}}(\upbm{x}{*}) = - m\bm{{\rm G}}^{-1}\bm{\nabla}f\autobrac{d_\phi \autobrac{\bm{x}^*, \bm{\theta}}} . \label{eq:full_IF}
\end{align}
Essentially, \eqref{eq:full_IF} is the same as the influence function in M-estimation \cite{robust}.
Since the matrix $\bm{{\rm G}}$ does not depend on the outlier, in order to investigate the boundedness of the influence function, we should evaluate the following term :
\begin{align}
\begin{split}
\MoveEqLeft
-\bm{\nabla}f\autobrac{d_\phi \autobrac{\bm{x}^*, \bm{\theta}}} = -f{\,'}\autobrac{d_\phi \autobrac{\bm{x}^*, \bm{\theta}}} \bm{\nabla}d_\phi \autobrac{\bm{x}^*, \bm{\theta}} \\
= {}&f{\,'}\autobrac{d_\phi \autobrac{\bm{x}^*, \bm{\theta}}} \autobrac{\bm{\nabla}\bm{\nabla}\phi\autobrac{\bm{\theta}}(\bm{x}^* - \bm{\theta})}. \\
\end{split} \label{eq:if}
\end{align}
This shows that the boundedness of the influence function depends on the function $f$ and the strictly convex function $\phi$ constituting the Bregman divergence.
If \eqref{eq:if} is bounded, the estimated cluster center is robust against outliers.

\subsection{Necessary condition for boundedness}
\begin{theorem} \label{th:bound}
The following condition of the function $f$ is necessary for its influence function to be bounded for all $\bm{x}^*$:
\begin{align}\lim_{z\to\infty}f{\,'}(z)=0 . \label{eq:limz}\end{align}
\end{theorem}
\begin{proof}
As $\|\bm{\theta}\|<\infty$, when $\|\bm{x}^*\|$ is large, $d_\phi \autobrac{\bm{x}^*, \bm{\theta}}$ is large.
When the function $f(z)$ is linear ($\nearrow$) or convex (\szarrow), $f{\,'}(z)$ is constant or monotonically increasing with respect to $z$.
In such a case, $\|\bm{x}^*\|$ is increased, and the norm of \eqref{eq:if} becomes as large as possible, which means the influence function is not bounded.
In order to reduce the norm of \eqref{eq:if} when $\|\bm{x}^*\|$ is large, $f{\,'}(z)$ must be monotonically decreasing.
The type of the function $f(z)$ that satisfies this condition is only concave (\uzarrow).
When $f{\,'}(z)$ does not satisfy \eqref{eq:limz}, the norm of \eqref{eq:if} is divergent at $\|\bm{x}^*\|\to\infty$, and hence the influence function is not bounded.
Therefore, the function $f(z)$ must satisfy \eqref{eq:limz}.
\qed
\end{proof}
\begin{remark}
In this paper, the function $f$ is assumed to be one of three types: linear, convex, and concave.
Therefore, from the proof of this theorem, we know that $f(z)$ is restricted to a subclass of concave functions in order to obtain the robustness.
However, even for non-concave functions, if the condition of Theorem \ref{th:bound} is satisfied, robustness can be obtained.
For example, a sigmoid function is not a concave function although it can induce robustness.
\end{remark}
\begin{remark} \label{rmk:infl}
In some cases, the factor $f{\,'}\autobrac{d_\phi \autobrac{\bm{x}^*, \bm{\theta}}}$ in \eqref{eq:if} can diverge to infinity around $\bm{x}^*=\bm{\theta}$.
Even in such a case, if we consider the region of $\bm{x}^*$ satisfying $d_\phi \autobrac{\bm{x}^*, \bm{\theta}}>\delta$ for a constant $\delta$, the condition of Theorem \ref{th:bound} provides a necessary condition for the boundedness of the influence function on the region.
\end{remark}

Under the condition of Theorem \ref{th:bound}, the norm of \eqref{eq:if} is bounded.
In the following, we consider whether the norm of the influence function vanishes as $\|\bm{x}^*\|\to\infty$.
In particular, if 
\begin{align}
\lim_{\|\bm{x}^*\|\to\infty} \|\bm{{\rm IF}}(\upbm{x}{*})\| = 0 \label{eq:redescending}
\end{align}
holds, the influence function is said to be redescending, and an outlier that is too large is automatically ignored.

The following Assumption \ref{asm:asm} is assumed in the following discussion.
\begin{assumption} \label{asm:asm}
The input dimension, the norm of the cluster center $\bm{\theta}$, and that of $\bm{\nabla}\phi$ at $\bm{\theta}$ are finite, that is,
$L < \infty$, $\|\bm{\theta}\| < \infty$, and $\|\bm{\nabla}\phi(\bm{\theta})\|<\infty$.
The Bregman divergence $d_\phi$ satisfies the followings for $l\in \{1,\cdots, L\}$:
\begin{align}
&\quad |{x^{*}}^{(l)}| \to \infty \Rightarrow d_\phi\autobrac{\bm{x}^{*},\bm{\theta}} \to \infty, \label{eq:scal_infty} \\
&\tilde{\bm{x}}=(\theta^{(1)}, \cdots, \theta^{(l-1)}, {x^*}^{(l)}, \theta^{(l+1)}, \cdots, \theta^{(L)})^{\rm T} \Rightarrow d_\phi(\tilde{\bm{x}}, \bm{\theta})\leq d_\phi(\bm{x}^*, \bm{\theta}). \label{eq:addtiveBregman}
\end{align}
\end{assumption}
Equations \eqref{eq:scal_infty} and \eqref{eq:addtiveBregman} hold true if the Bregman divergence $d_\phi(\bm{x}, \bm{\theta})$ is defined additively with respect to $L$-dimensions.
Below, we discuss the situation where $\|\bm{x}^*\| \to \infty$ holds under these assumptions.
In some cases, such as the Bregman divergence corresponding to the binomial distribution, $\|\bm{x}^*\| \to \infty$ can not occur.
However we can investigate the behavior of the influence function for finite $\bm{x}^*$.
We illustrate the behavior of \eqref{eq:if} for such a case in \ref{sec:plot_if}.

\subsection{Power mean\label{sec:pow}}
From \eqref{eq:if} and \eqref{eq:f_pow}, to evaluate the influence function in the case of the power mean, we evaluate the following term:
\begin{align*}
\lim_{\|\bm{x}^*\|\to\infty} \left\| \frac{\bm{\nabla}\bm{\nabla}\phi\autobrac{\bm{\theta}}(\bm{x}^* - \bm{\theta})}{\left[d_\phi \autobrac{\bm{x}^*, \bm{\theta}}+a\right]^{1-\beta}}\right\|.  
\end{align*}
\begin{theorem} \label{th:pow}
For the function $f$ of the power mean \eqref{eq:f_pow} with $\beta<0$, the influence function is redescending for general Bregman divergences.
\end{theorem}
The proof of Theorem \ref{th:pow} is in \ref{sec:pow_redec}.
However, when $0\leq\beta<1$, the redescending or boundedness property of the influence function depends on the Bregman divergence.
In the rest of this subsection, we investigate the influence functions for concrete examples of the Bregman divergences.

\begin{figure}[tb]
\begin{center}
  \includegraphics[width=0.7\textwidth]{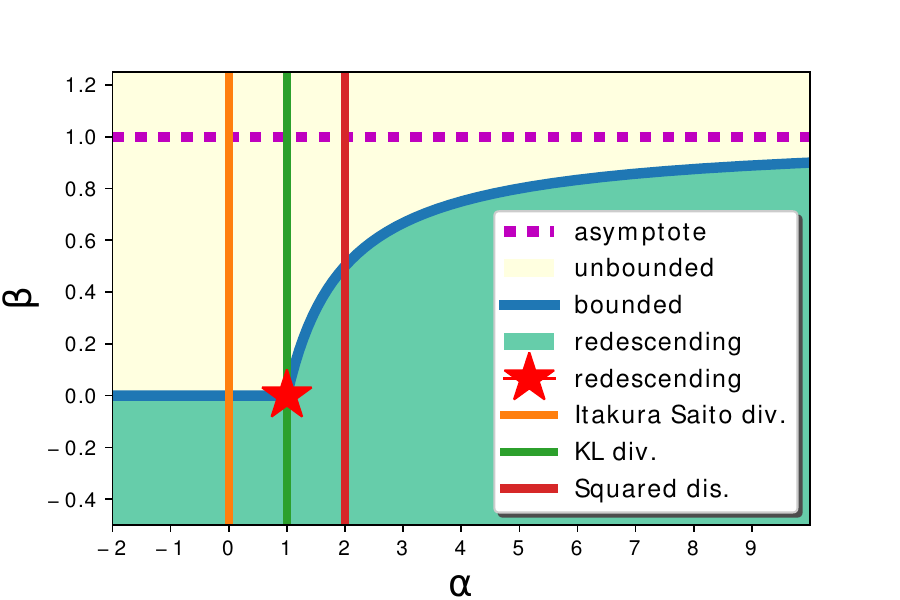}
\end{center}
\caption{Robustness of the $\alpha$-divergence in the case of generalization with the power mean.}
\label{fig:beta-div}
\end{figure}

$\alpha$-divergence is a subclass of Bregman divergences, including the Itakura Saito divergence ($\alpha=0$), the generalized KL divergence ($\alpha=1$), and the squared distance ($\alpha=2$) as special cases \cite{beta-divergence}.\footnotemark
\footnotetext{The $\alpha$-divergence here is usually termed as the $\beta$-divergence with the parameter $\beta$.
We denote the parameter by $\alpha$ in order not to be confused with the $\beta$ in \eqref{eq:f_pow}.}
The $\alpha$-divergence and corresponding convex functions are given by 
\begin{align*}
&d_\alpha\autobrac{x, \theta} = \begin{cases}
\frac{x}{\theta}-\ln\autobrac{\frac{x}{\theta}}-1 & (\alpha=0) , \\
x\ln\autobrac{\frac{x}{\theta}}-(x-\theta) & (\alpha=1) , \\
\frac{x^\alpha+(\alpha-1)\theta^\alpha-\alpha x\theta^{\alpha-1}}{\alpha(\alpha-1)} & ({\rm otherwise}) , \\
\end{cases} \\
&\phi_\alpha(x) = \begin{cases}
 -\ln x + x -1 & (\alpha = 0) ,\\
x\ln x - x + 1 & (\alpha = 1) ,\\
\frac{{x}^\alpha}{\alpha(\alpha-1)} - \frac{x}{\alpha-1} + \frac{1}{\alpha} & ({\rm otherwise}) ,
\end{cases} 
\end{align*}
respectively.
In the convex function, when the parameter $\alpha$ is a positive even number other than $0$, its domain is defined as $\mathbb{R}$, otherwise it is $\mathbb{R}_{+}\setminus\{0\}$.
If the data is multidimensional, the Bregman divergence and the corresponding convex function are defined additively with respect to dimensions as follows :
\begin{align*}
&d_\phi\autobrac{\bm{x}, \bm{\theta}} = \sum_{l=1}^L d_\alpha(x^{(l)}, \theta^{(l)}) ,\\
&\phi(\bm{x}) = \sum_{l=1}^L \phi_\alpha(x^{(l)}).
\end{align*}
We calculated the influence function for the $\alpha$-divergence and found that it can be classified into divergent, bounded, and redescending types with respect to $\alpha$ and $\beta$ according to the following conditions (the proof is in \ref{sec:beta_proof}) : 
\begin{align*}
\alpha<1 &:& \begin{cases}
\beta>0 & \Rightarrow \|\bm{{\rm IF}}(\upbm{x}{*})\| \; {\rm is} \; {\rm divergent}, \\
\beta = 0 & \Rightarrow \|\bm{{\rm IF}}(\upbm{x}{*})\| \; {\rm is} \; {\rm bounded}, \\
\beta <0 & \Rightarrow \|\bm{{\rm IF}}(\upbm{x}{*})\| \; {\rm is} \; {\rm redescending}, \\
\end{cases}\\
\alpha=1 &:& \begin{cases}
\beta>0 & \Rightarrow \|\bm{{\rm IF}}(\upbm{x}{*})\| \; {\rm is} \; {\rm divergent}, \\
\beta \leq 0 & \Rightarrow \|\bm{{\rm IF}}(\upbm{x}{*})\| \; {\rm is} \; {\rm redescending},
\end{cases}\\
\alpha>1 &:& \begin{cases}
\beta>1-\frac{1}{\alpha} & \Rightarrow \|\bm{{\rm IF}}(\upbm{x}{*})\| \; {\rm is} \; {\rm divergent}, \\
\beta = 1-\frac{1}{\alpha} &\Rightarrow \|\bm{{\rm IF}}(\upbm{x}{*})\| \; {\rm is} \; {\rm bounded}, \\
\beta <1-\frac{1}{\alpha} & \Rightarrow \|\bm{{\rm IF}}(\upbm{x}{*})\| \; {\rm is} \; {\rm redescending} . \\
\end{cases}
\end{align*}
Figure \ref{fig:beta-div} shows the regions of $(\alpha, \beta)$ corresponding to the three types.
Specifically, we see the boundedness property holds at the boundary line between the divergent and the redescending properties of the influence function.
This boundary line is continuous except for the point of $\alpha=1$, and gradually approaches $\beta=1$ when $\alpha\to\infty$.
The case of another divergence, the exp-loss is shown in \ref{sec:exp-loss}.

\subsection{Log-sum-exp\label{sec:lse}}
From \eqref{eq:if} and \eqref{eq:f_exp}, to evaluate the influence function in the case of the log-sum-exp function, we evaluate the following term:
\begin{align*}
\lim_{\|\bm{x}^*\|\to\infty} \left\|\frac{{\nabla\nabla\phi(\bm{\theta})(\upbm{x}{*}-\bm{\theta})}}{\exp\left((1-\beta)d_\phi \autobrac{\bm{x}^*, \bm{\theta}}\right)}\right\|. \end{align*}
\begin{theorem} \label{th:lse}
For the function $f$ of the log-sum-exp \eqref{eq:f_exp} with $\beta<1$, the influence function is redescending for general Bregman divergences.
\end{theorem}
Theorem \ref{th:lse} shows that when the estimated cluster center is robust against outliers, the redescending property always holds for any Bregman divergences.
(The proof of Theorem \ref{th:lse} is in \ref{sec:lse_redec}.)

From the above examples, it can be seen that the robustness property of $f$-mean strongly depends on the function $f$.

\subsection{Discussion on the influence function \label{sec:outlier}}
In Section \ref{sec:if}, we have discussed the boundedness of the influence function when the norm of the outlier $\left\|\bm{x}^*\right\|$ goes to infinity.
As the property of the DP-means algorithm, the data point which satisfies \eqref{eq:clusterAdd} generates a new cluster.
Hence, updating of the cluster center is not affected by such a data point.
However, data points which satisfy
\begin{align}d_\phi(\bm{x}, \bm{\theta})\leq \lambda \label{eq:no_outlier}\end{align} may effectively work as outliers.
Clustering performance may be badly affected by data points satisfying \eqref{eq:no_outlier} for the fixed penalty parameter $\lambda$ depending on the dataset.
In fact, the clustering performance of DP-means depends on the selection of function $f$ as we will examine numerically in Section \ref{sec:experiments}.
Therefore, it is important to consider the robustness against outliers.
On the other hand, efficiency and robustness trade-off: if priority is given to efficiency, lower robustness may be better.
In such a case, we may choose a concave function similar to the linear function which corresponds to an unbounded influence function.
In any case, it is important to examine the influence function, and it can be a criterion for the design of the function $f$ and the penalty parameter $\lambda$ from the dataset.
\section{Experiments \label{sec:experiments}}
\subsection{UCI experiment \label{sec:uci_experiment}}
We conducted experiments with benchmark datasets in UCI Machine Learning Repository\footnote{https://archive.ics.uci.edu/ml/datasets.html} to investigate the influence of the objective function generalized by the monotonically increasing function $f$.
We focused on the power mean \eqref{eq:f_pow} and the log-sum-exp function \eqref{eq:f_exp}.
It is difficult to fairly compare DP-means with the other clustering methods introduced in Section \ref{sec:intro} that can estimate the number of clusters because the calculation time of $\bm{O}(n^2)$ is necessary or many parameters are required, and the preconditions of the algorithms are different.
Also, as discussed in Section \ref{sec:comp2}, the clustering time of the generalized DP-means depends on the estimated number of clusters, which changes against $\beta$ for a fixed $\lambda$. 
Thus, we focused on the clustering performance of the generalized DP-means other than the computational time.

In Section \ref{sec:outlier}, we discussed that outliers satisfying \eqref{eq:no_outlier} and depending on $\lambda$ can influence the performance of DP-means. In the experiments below, we assume that real datasets contain such outliers implicitly. In other words, the improvement of the performance of the generalized DP-means for $\beta<1$ reflects the fact that datasets contain such outliers deteriorating the original DP-means.

\subsubsection{Dataset}
The datasets used for the experiment, where $n$, $K$, and $L$ denote the number of data, the number of clusters, and the number of dimensions, respectively are summarized in Table \ref{table:uci_data}, where links to the specific datasets are given as footnotes when there are multiple datasets.
Datasets for classification problems were used assuming classes as the true clusters.
Heart dataset consists of data on heart disease with five clusters. 
Four clusters represent heart disease and one cluster represents no heart disease.
HeartK2 dataset was made of Heart dataset by coarsening the cluster labels.
More specifically it was made from the two clusters, with and without heart disease.
We deleted data points with missing values beforehand.
\begin{table}[t]
\begin{center} 
\caption{UCI datasets. }
  \begin{tabular}{crrrr} \toprule
  dataset & $n$ & $K$ & $L$ \\ \midrule
  Breast Cancer Wisconsin\footnotemark[5] & $683$ & $2$ & $9$ \\
  Heart\footnotemark[6] & $297$ & $5$ & $13$ \\
  HeartK2 & $297$ & $2$ & $13$ \\
  HTRU2 & $17898$ & $2$ & $8$ \\
  Iris & $150$ & $3$ & $4$ \\
  Mice Protein Expression & 552 & 8 & 77 \\
  Pima & $768$ & $2$ & $8$ \\
  Seeds & $210$ & $3$ & $7$ \\
  Thyroid\footnotemark[7] & $215$ & $3$ & $5$ \\
  Wine & $178$ & $3$ & $13$ \\ 
  Yeast & $1484$ & $10$ & $8$ \\ \bottomrule
  \end{tabular}
\label{table:uci_data}
\end{center}
\end{table}
\footnotetext[5]{https://archive.ics.uci.edu/ml/machine-learning-databases/breast-cancer-wisconsin/breast-cancer-wisconsin.data}
\footnotetext[6]{https://archive.ics.uci.edu/ml/machine-learning-databases/heart-disease/processed.cleveland.data}
\footnotetext[7]{http://archive.ics.uci.edu/ml/machine-learning-databases/thyroid-disease/new-thyroid.data}

\subsubsection{Evaluation criteria}
We used the number of clusters and normalized mutual information (NMI) as the evaluation criteria to investigate the influence of the penalty parameter $\lambda$ and control parameter $\beta$ for the objective function.
In order to confirm the behavior of the objective function, we examined the behavior of the maximum distortion against the change of $\beta$.
NMI is criterion for evaluating the clustering result, and take values from $0$ to $1$.
The closer the NMI is to $1$, the better the result.
NMI is defined by the following equation:
\begin{align*}
{\rm NMI}(C, A) = \frac{I(C, A)}{\sqrt{H(C)H(A)}}
\end{align*}
for the label set $C$ of the clustering result and the label set $A$ of the correct cluster.
Here $I(\cdot,\cdot)$ and $H(\cdot)$ represent mutual information and entropy, respectively.

\subsubsection{Method \label{sec:method}}
For preprocessing of clustering, we standardized data so that  the each dimension is transformed as $x_i^{(l)}\leftarrow \frac{x_i^{(l)}}{\sqrt{\frac{1}{n}\sum_{i=1}^n (x_i^{(l)})^2}}$.
We chose the squared distance $d_\phi\autobrac{\bm{x}, \bm{\theta}}=\frac{1}{L} \|\bm{x}-\bm{\theta}\|^2$ as the distortion measure, which is averaged with respect to the dimensions.

In the experiment, we investigated the change of each evaluation criterion when changing the parameter $\beta$ for each case of the power mean \eqref{eq:pow} with $a=0$ and log-sum-exp function \eqref{eq:lse}.
The range of $\beta$ examined was $[-2,5]$.
DP-means returns a local minimum solution depending on the order of data.
Hence, the order of data was shuffled $100$ times, and for each evaluation criteria, the average value on the number of shuffles was calculated and used as the result.
The concrete procedure is shown below.
\begin{enumerate}[Step 1.]
  \item When the number of clusters is 1, find the maximum value of the maximum distortion $\hat{d}_m$ in the range of $\beta\in[-2, 5]$.
  \item Randomly rearrange the sequence of data. \label{item:first}
  \item Change $\beta$ from $-2$ to $5$ in $0.1$ increments. \label{item:second}
  \begin{enumerate}[Step 3-1.]
    \item Set $\lambda^{(t)} = \begin{cases} \hat{d}_m & (t=0), \\ \lambda^{(t-1)}/1.01 &  (t=1, 2, \cdots, ). \end{cases}$ \label{item:lambda}
    \item Algorithm is executed with the penalty parameter $\lambda^{(t)}$. \label{item:exe}
    \item Repeat Step 3-1 and Step 3-2 until $\lambda^{(t)}$ falls below the threshold.
  \end{enumerate}
  \item Repeat Step 2 and Step 3 100 times.
  \item Average the evaluation criteria over $100$ rearrangements of data for each $\beta$ and $\lambda$, and use them as the result.
\end{enumerate}
The threshold was set so that the number of clusters was within about three times the number of correct clusters.

Note that when the Bregman divergence is defined as the average with respect to the dimension $L$, in the case of the log-sum-exp function, the effective value of the parameter $\beta$ depends on $L$.
When the parameter to be given is $\beta^*$, the effective value of the parameter is $\beta=\frac{\beta^*-1}{L}+1$.
Thus, the range of $\beta$ is $[\frac{-3}{L}+1, \frac{4}{L}+1]$ and the step size is $\frac{0.1}{L}$.

\begin{figure*}[htbp]
\centering
		\subfloat[Number of clusters \label{fig:under_lim_min_ave}]{
 			 \includegraphics[width=47mm]{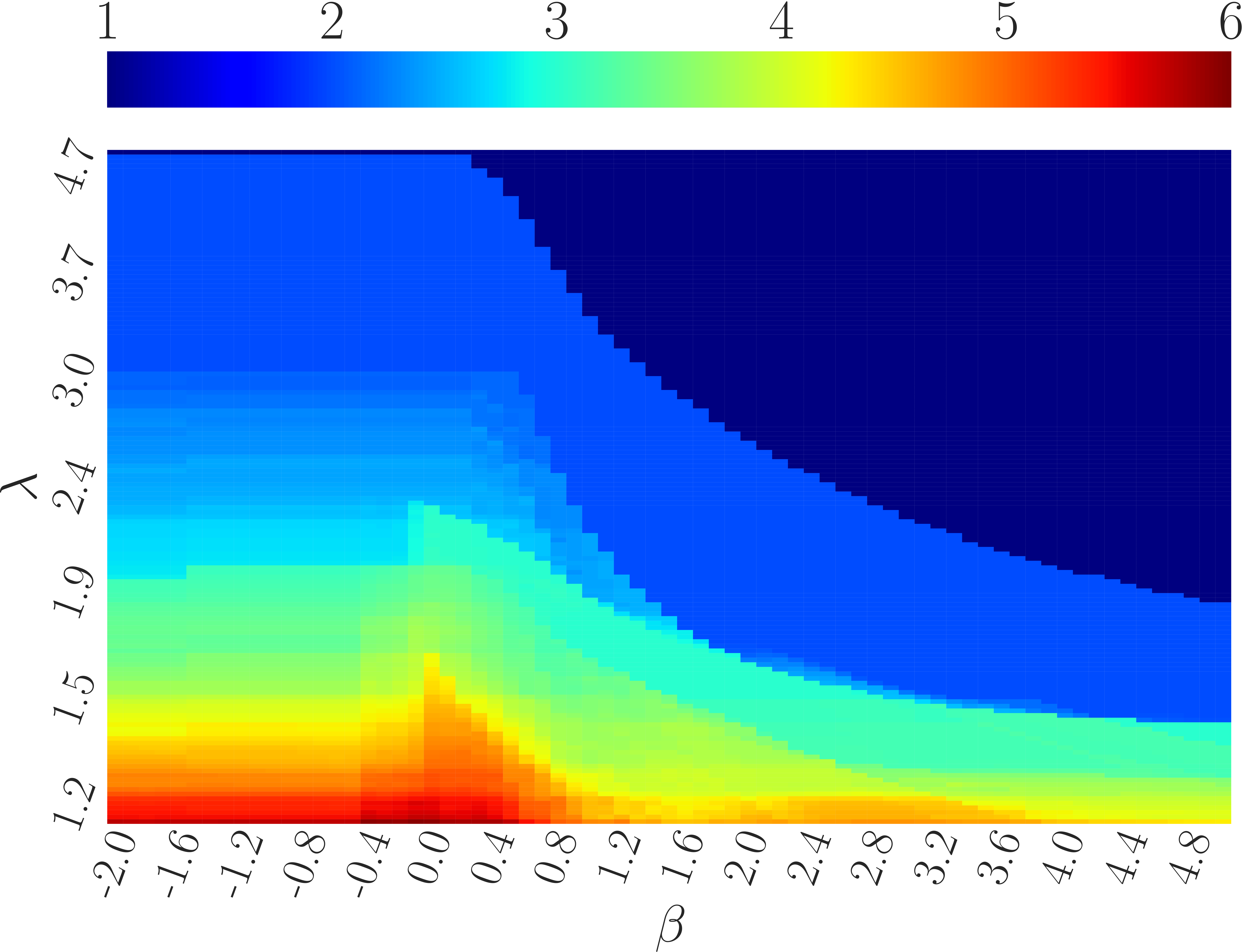}
		}
		\subfloat[NMI \label{fig:under_lim_min_max}]{
			 \includegraphics[width=47mm]{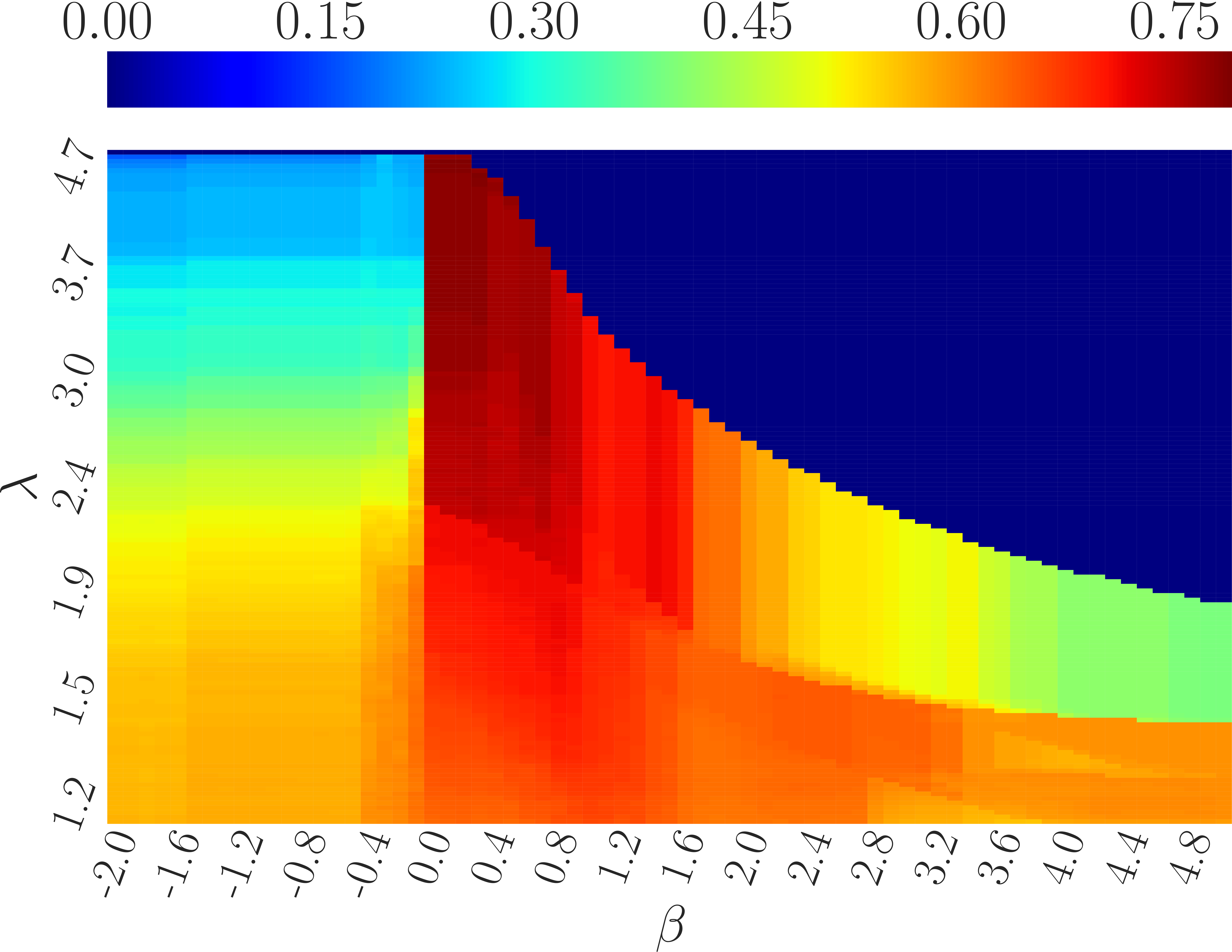}
		}
             \subfloat[Maximum distortion \label{fig:under_lim_min_max}]{
			 \includegraphics[width=47mm]{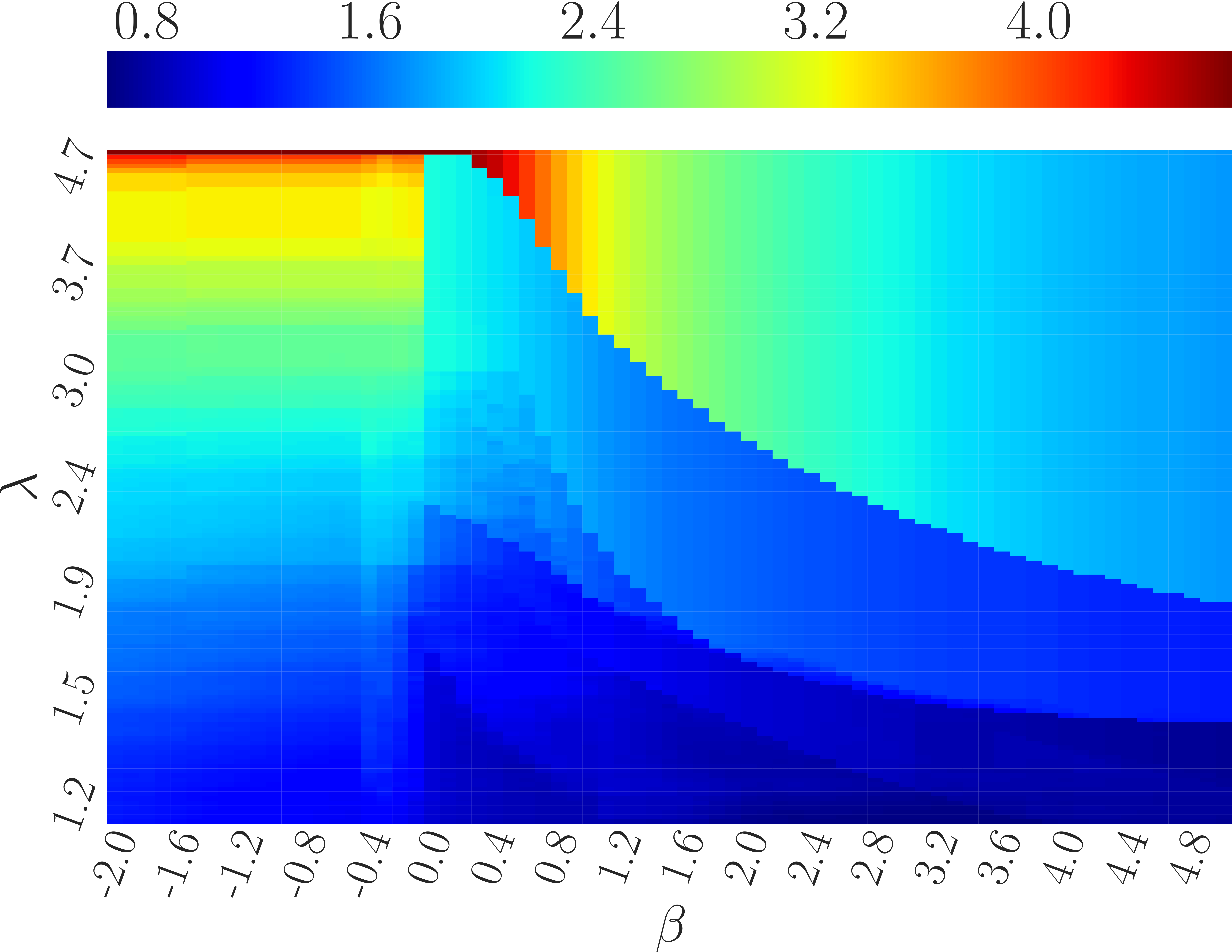}
		}
\vspace{-10pt}
		\caption{Breast Cancer Wisconsin, power mean \label{fig:bcw_pow} }
\vspace{-10pt}

\centering
		\subfloat[Number of clusters \label{fig:under_lim_min_ave}]{
 			 \includegraphics[width=47mm]{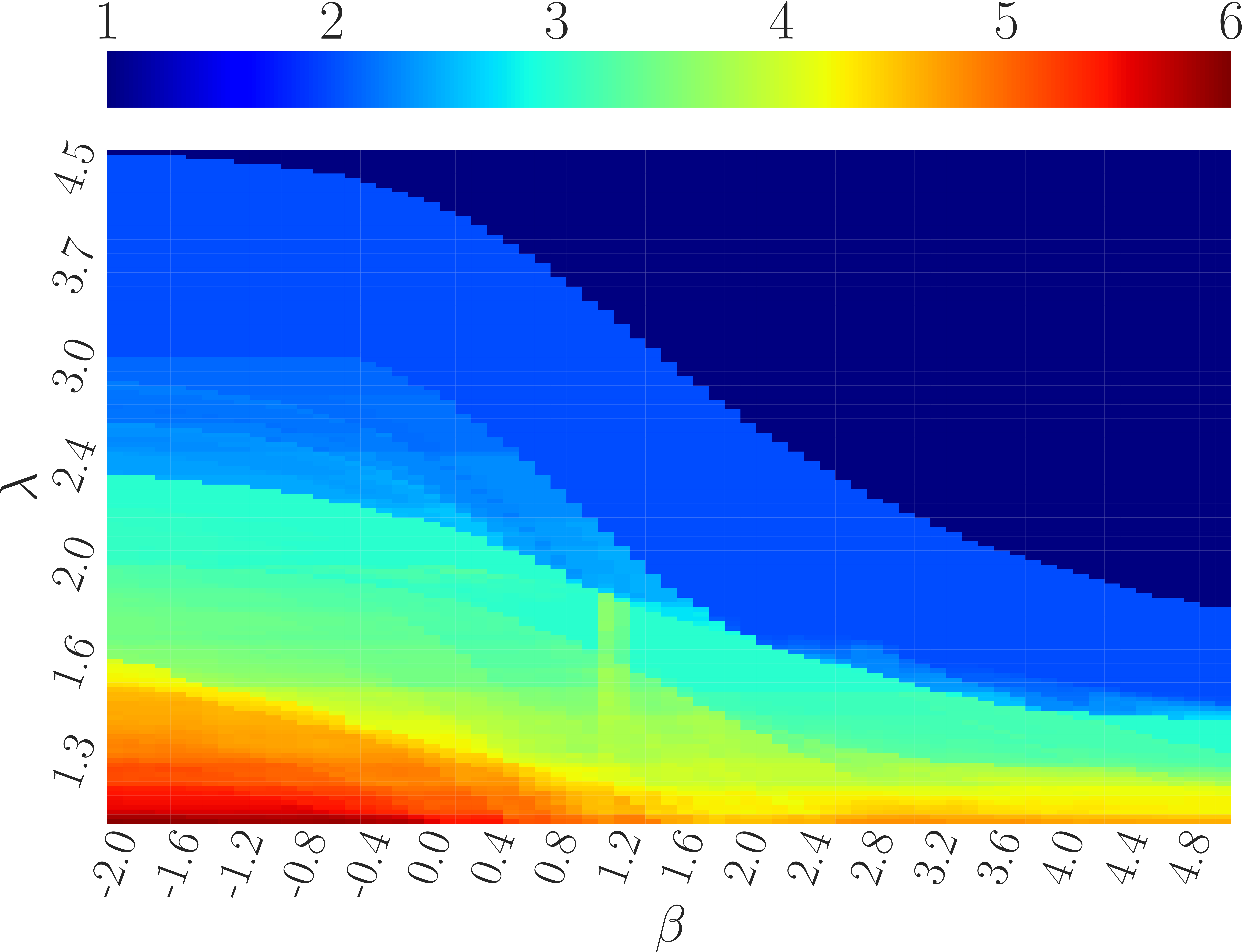}
		}
		\subfloat[NMI \label{fig:under_lim_min_max}]{
			 \includegraphics[width=47mm]{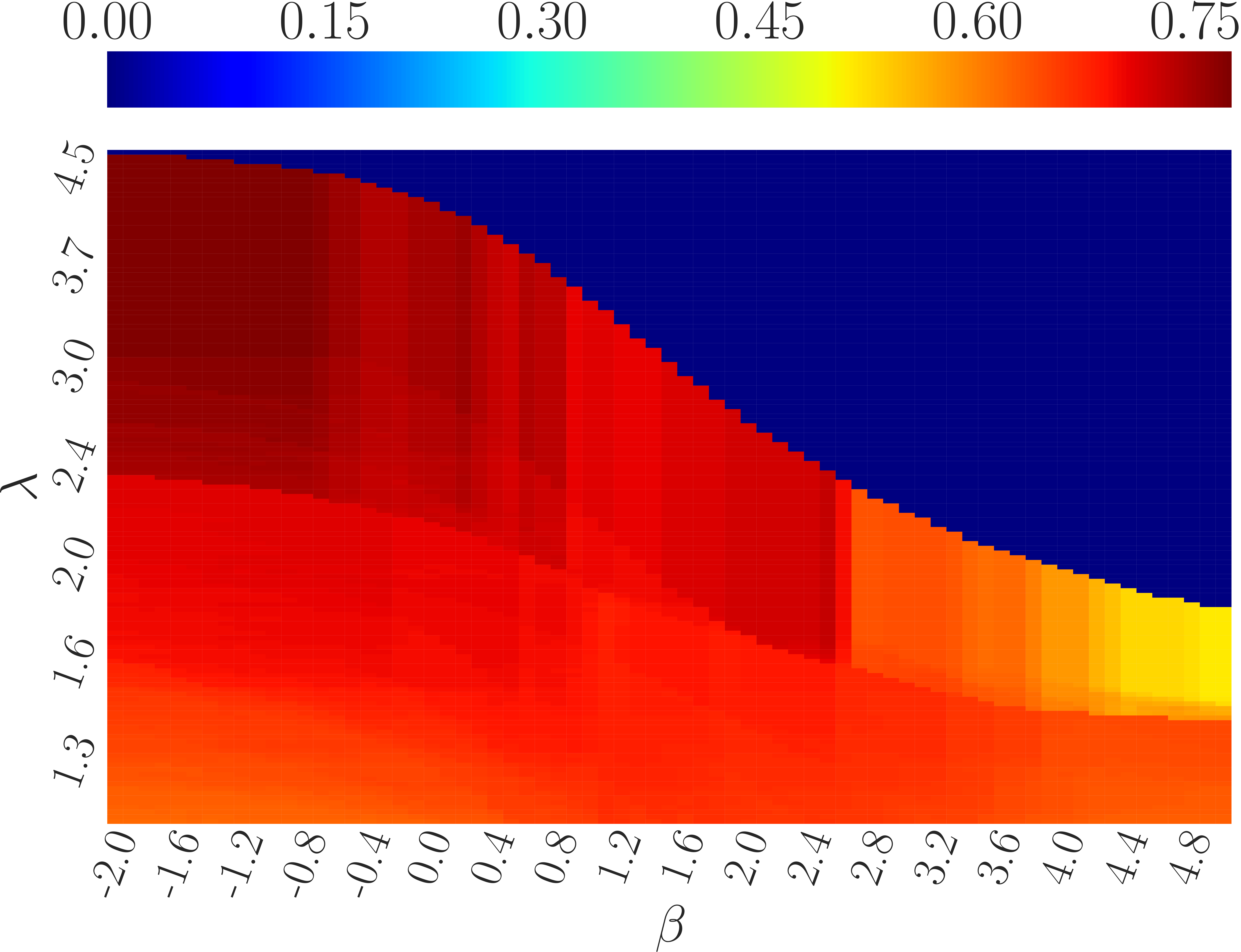}
		}
             \subfloat[Maximum distortion \label{fig:under_lim_min_max}]{
			 \includegraphics[width=47mm]{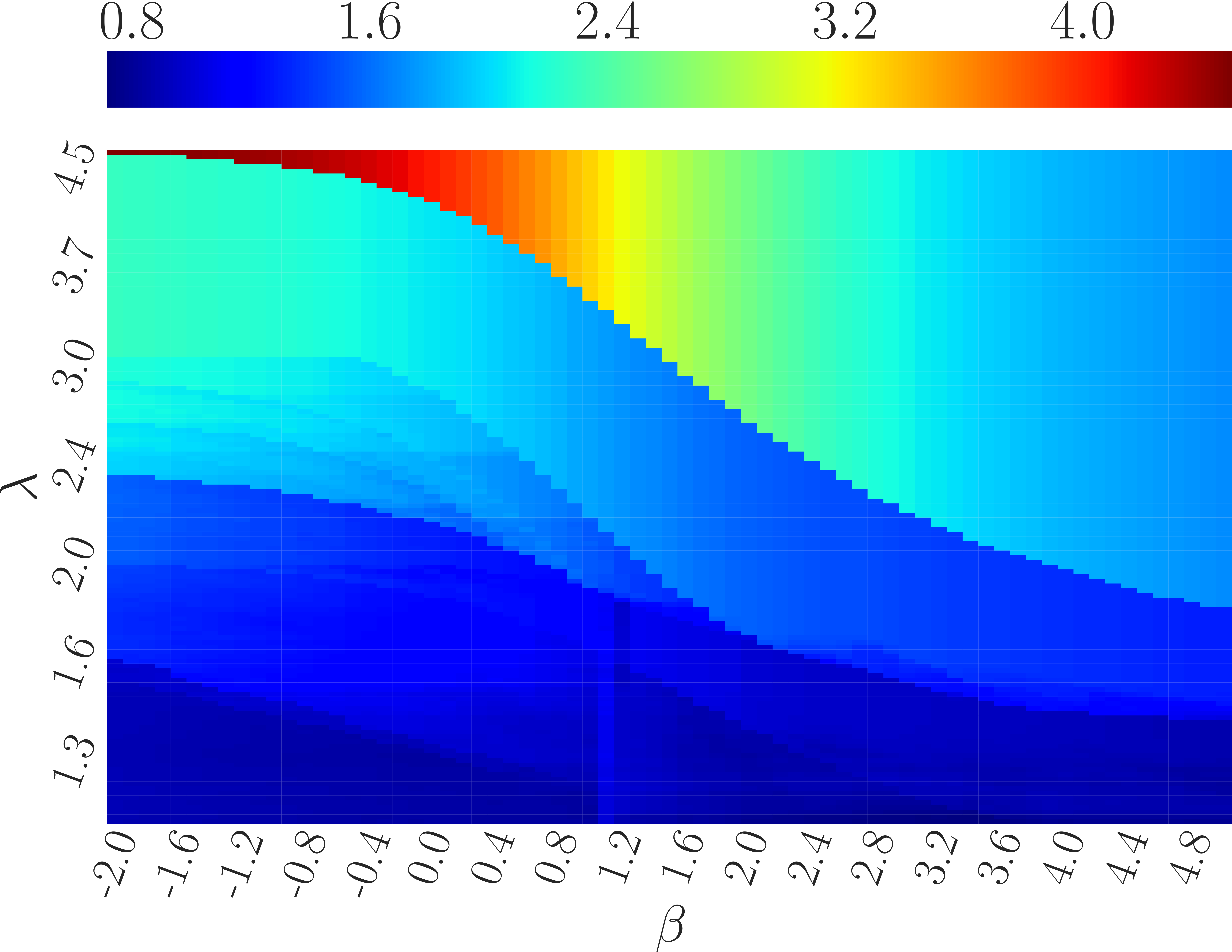}
		}
\vspace{-10pt}
		\caption{Breast Cancer Wisconsin, log-sum-exp \label{fig:bcw_exp} }
\vspace{-10pt}
\end{figure*}

\begin{figure*}[htbp]
\centering
		\subfloat[Number of clusters \label{fig:under_lim_min_ave}]{
 			 \includegraphics[width=47mm]{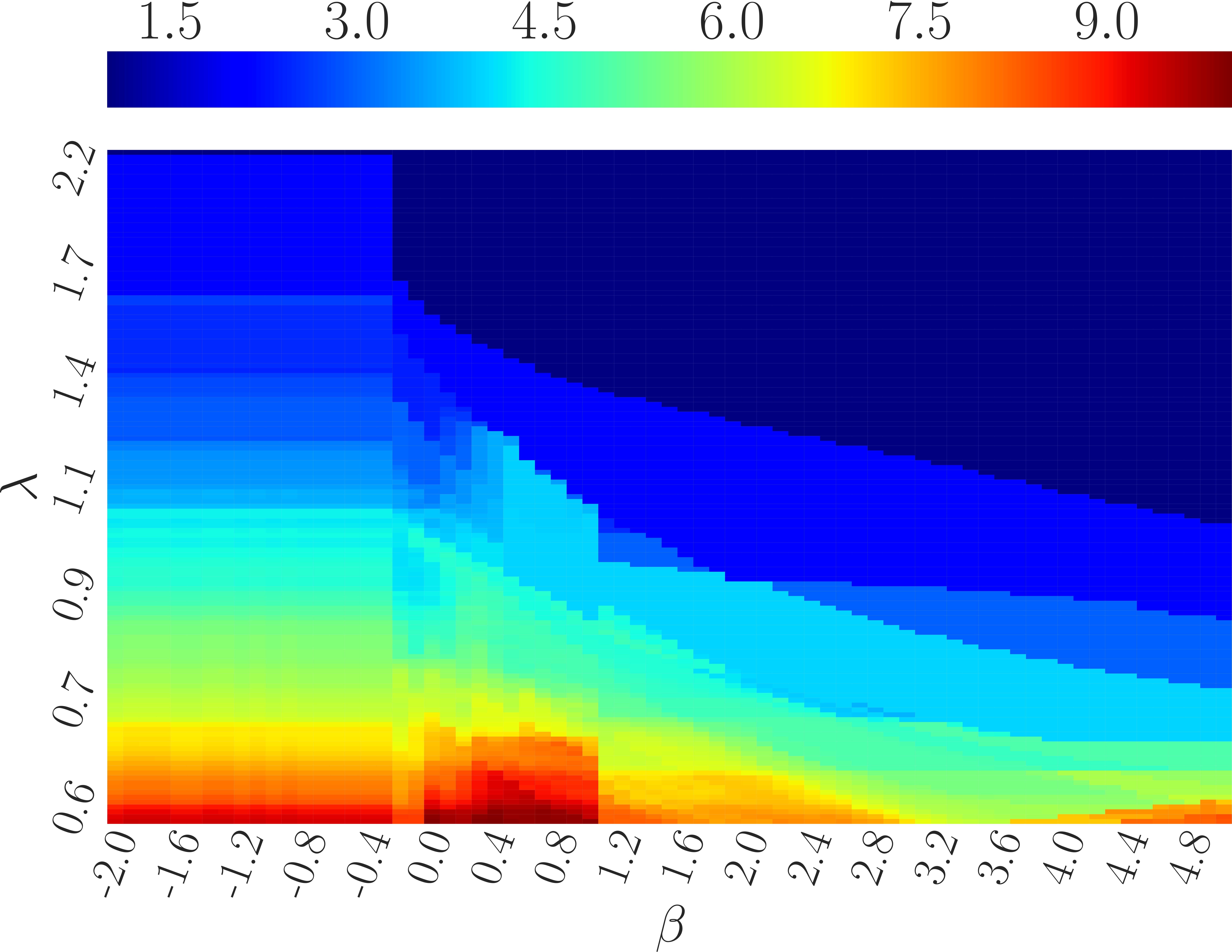}
		}
		\subfloat[NMI \label{fig:under_lim_min_max}]{
			 \includegraphics[width=47mm]{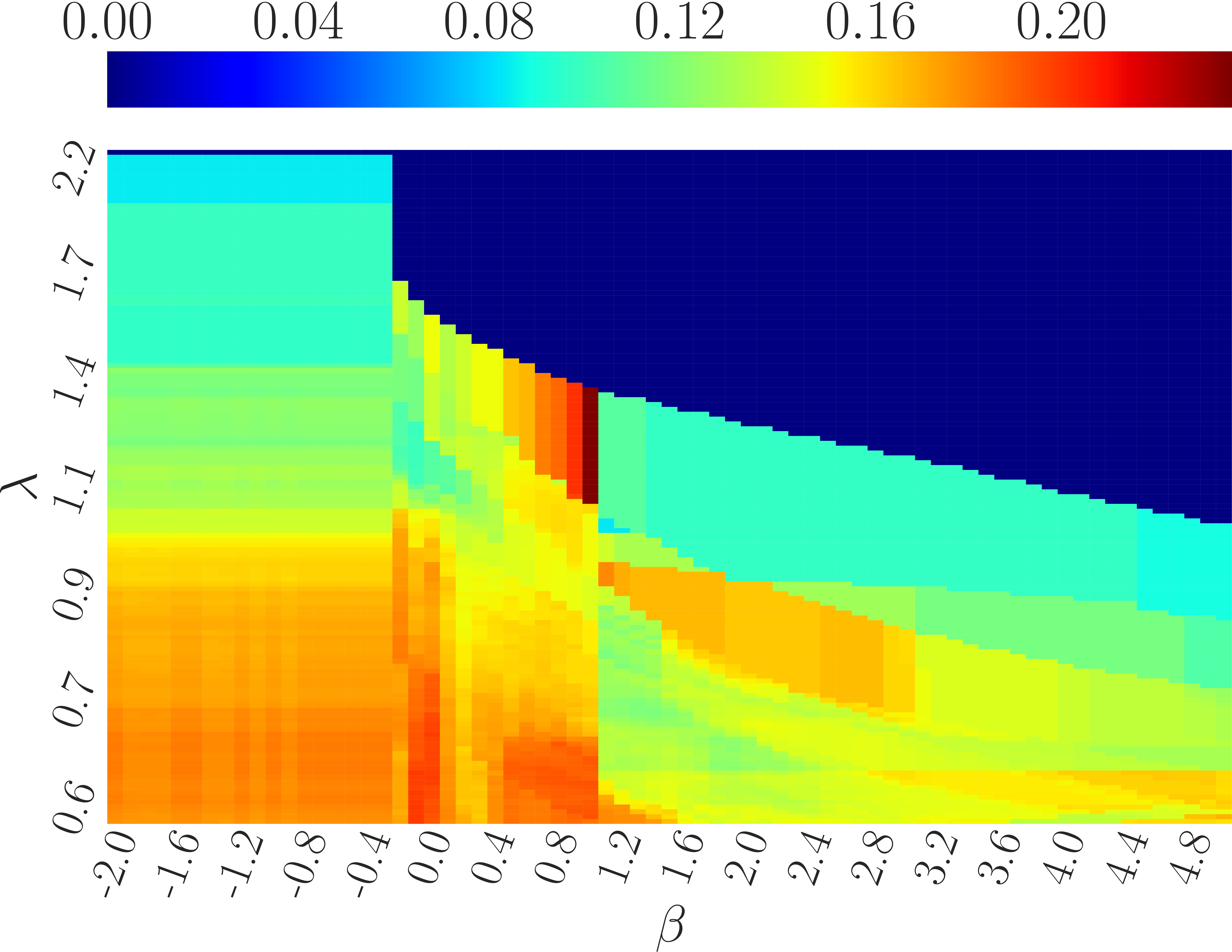}
		}
             \subfloat[Maximum distortion \label{fig:under_lim_min_max}]{
			 \includegraphics[width=47mm]{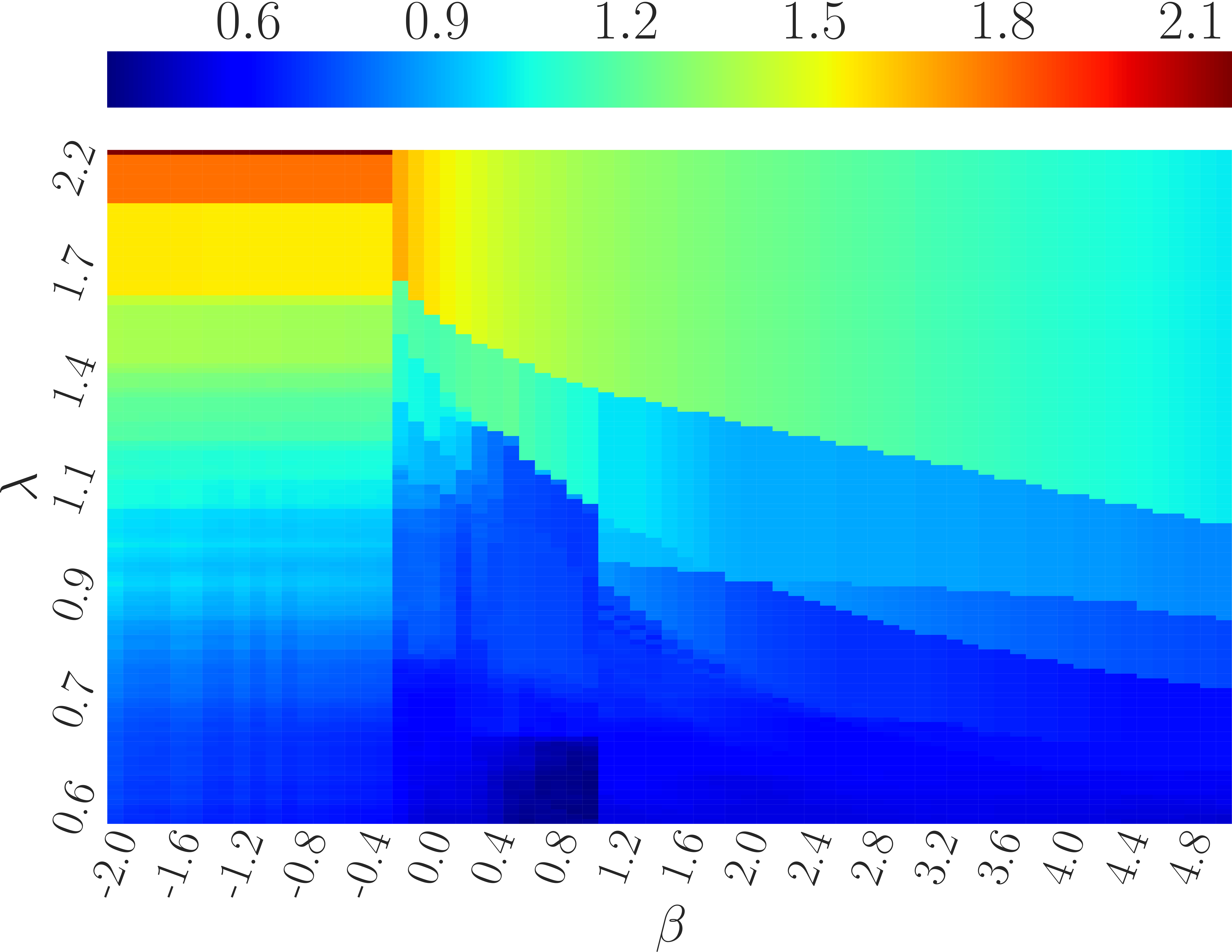}
		}
\vspace{-10pt}
		\caption{Heart, pow mean \label{fig:heart_pow} }
\vspace{-10pt}

\centering
		\subfloat[Number of clusters \label{fig:under_lim_min_ave}]{
 			 \includegraphics[width=47mm]{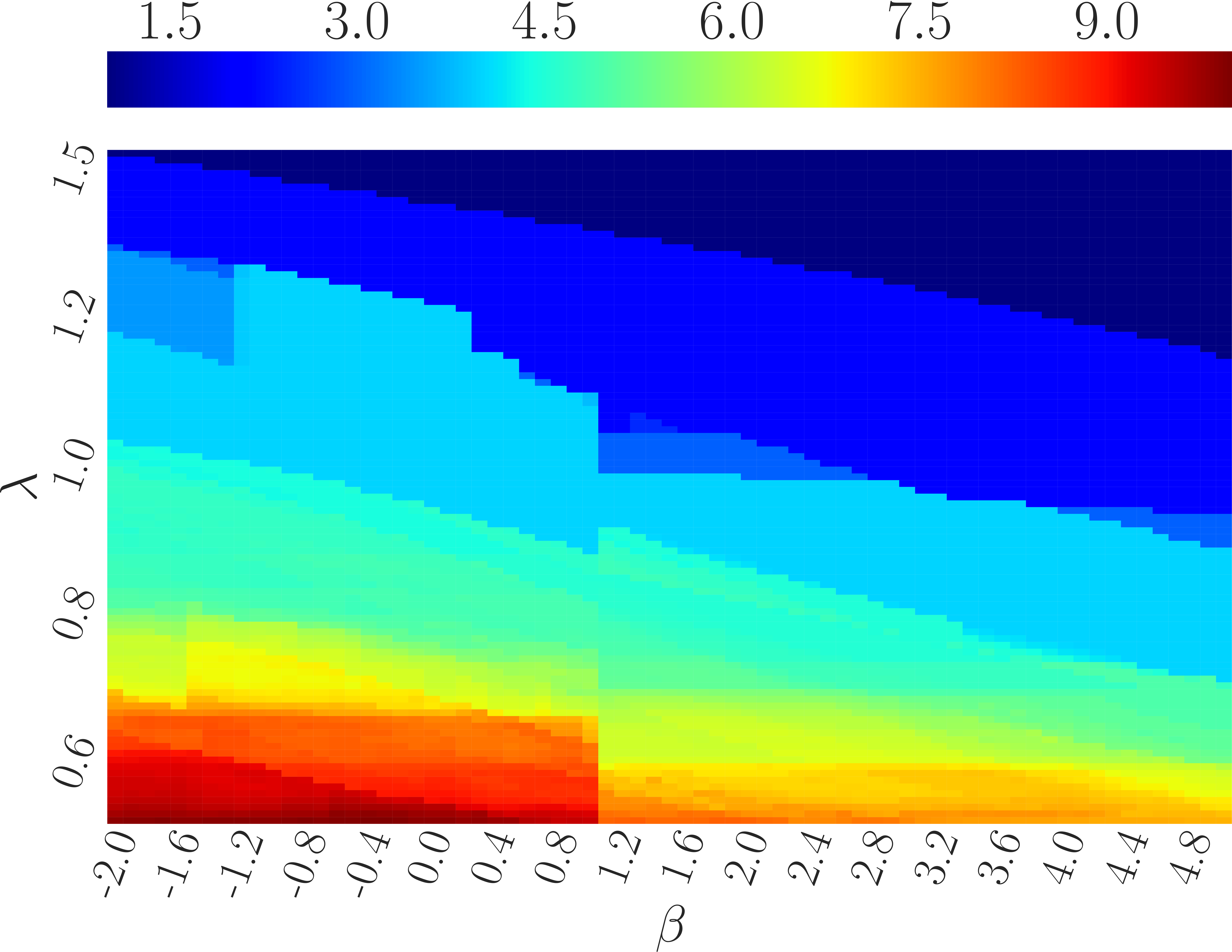}
		}
		\subfloat[NMI \label{fig:under_lim_min_max}]{
			 \includegraphics[width=47mm]{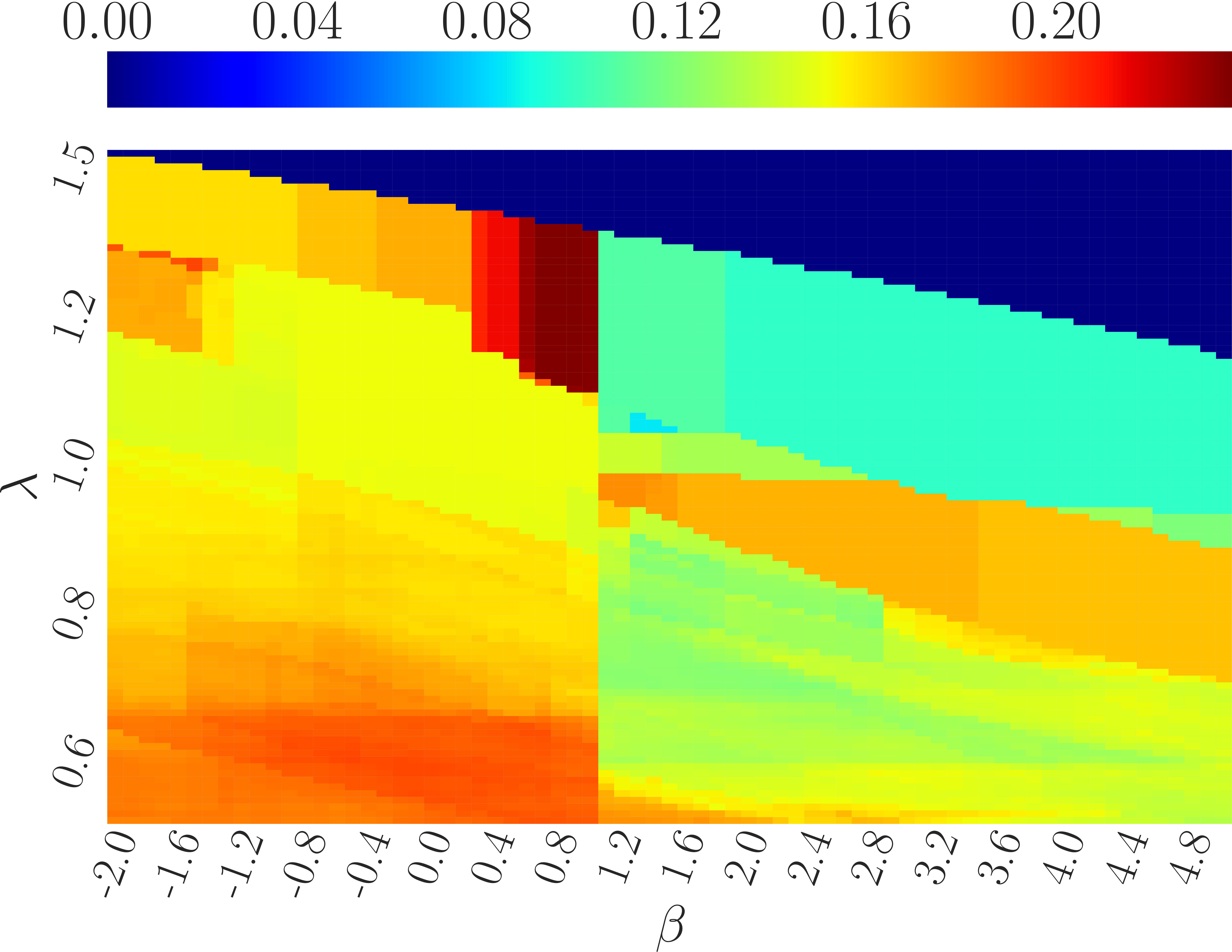}
		}
             \subfloat[Maximum distortion \label{fig:under_lim_min_max}]{
			 \includegraphics[width=47mm]{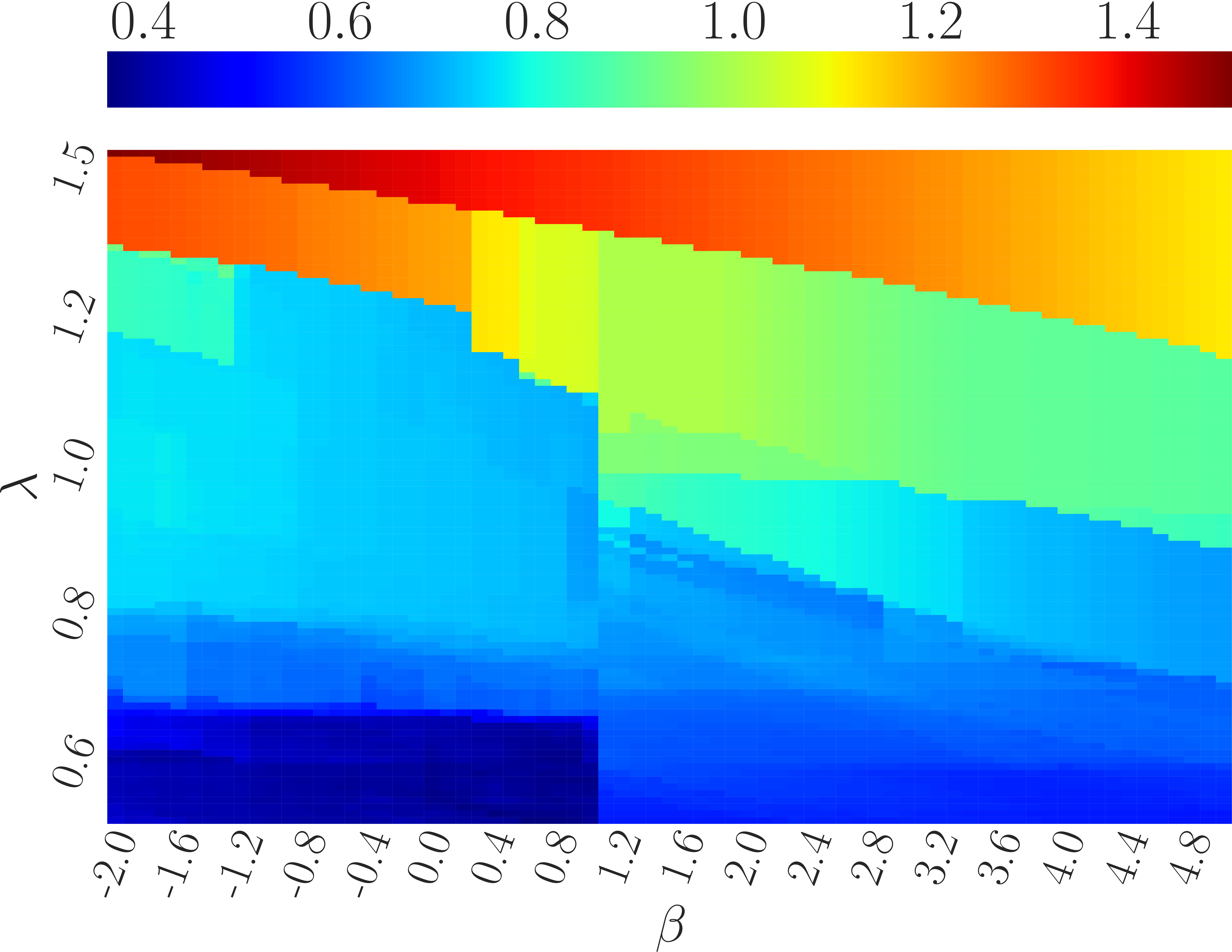}
		}
\vspace{-10pt}
		\caption{Heart, log-sum-exp \label{fig:Heart_exp} }
\vspace{-10pt}
\end{figure*}

\begin{figure*}[htbp]
\centering
		\subfloat[Number of clusters \label{fig:under_lim_min_ave}]{
 			 \includegraphics[width=47mm]{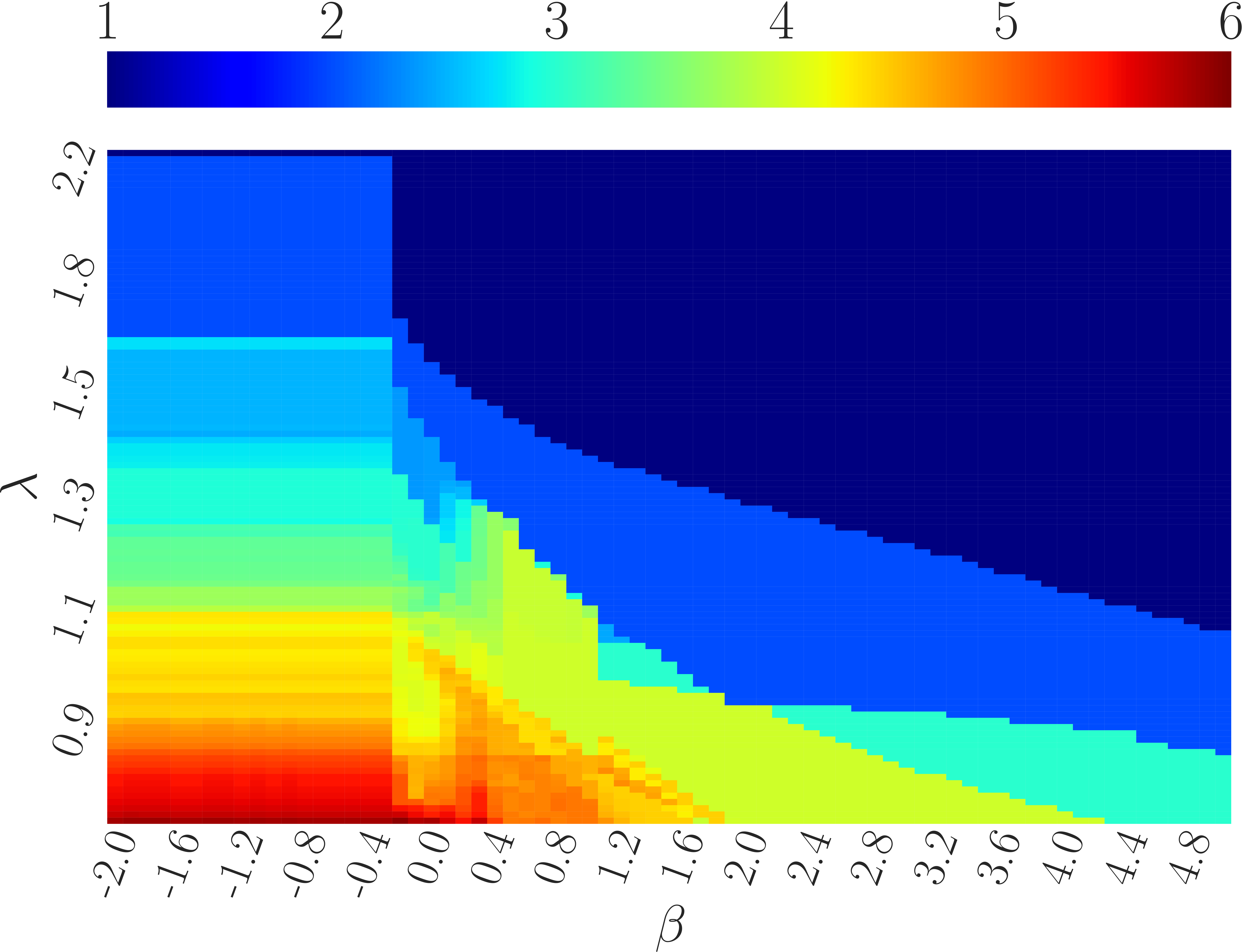}
		}
		\subfloat[NMI \label{fig:under_lim_min_max}]{
			 \includegraphics[width=47mm]{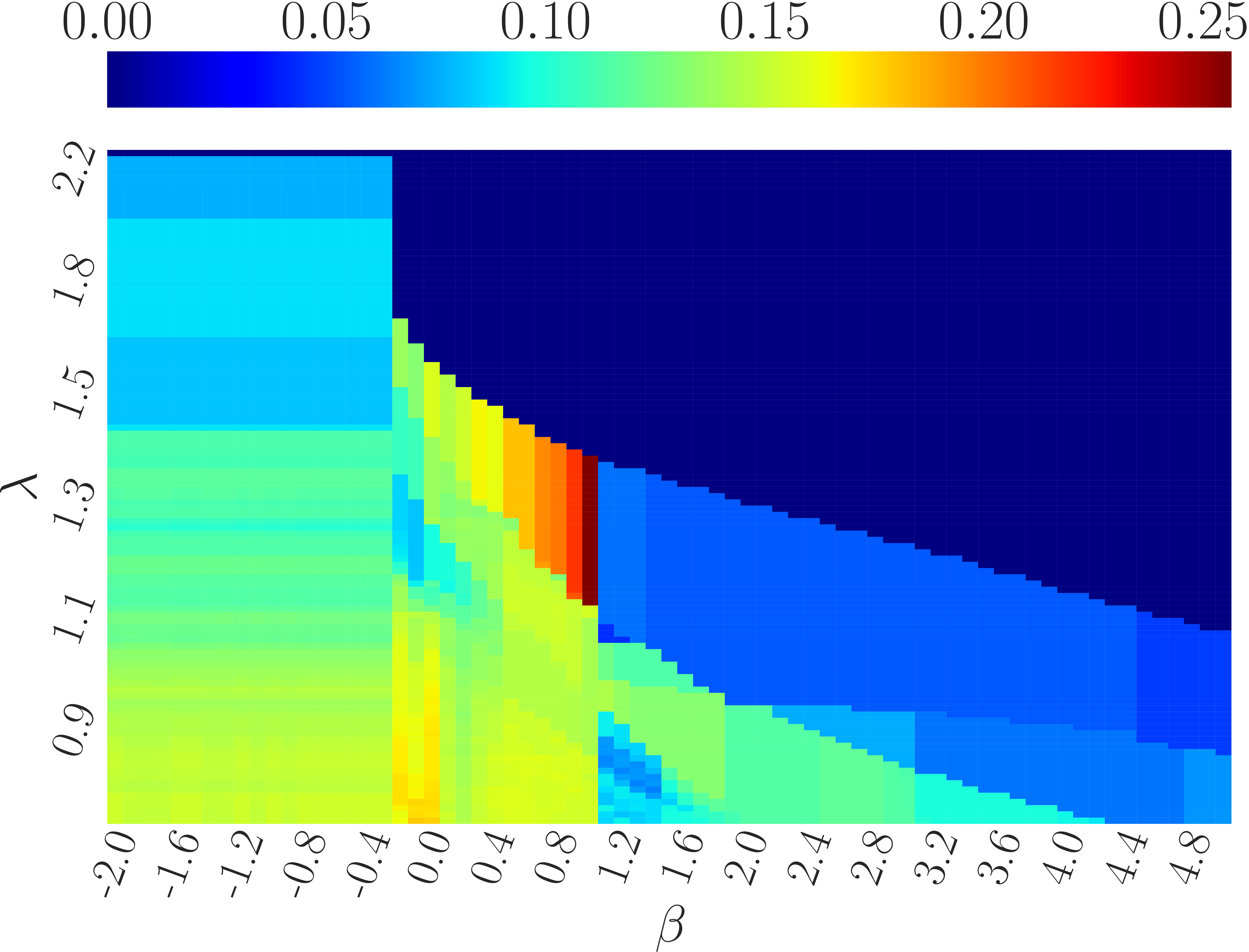}
		}
             \subfloat[Maximum distortion \label{fig:under_lim_min_max}]{
			 \includegraphics[width=47mm]{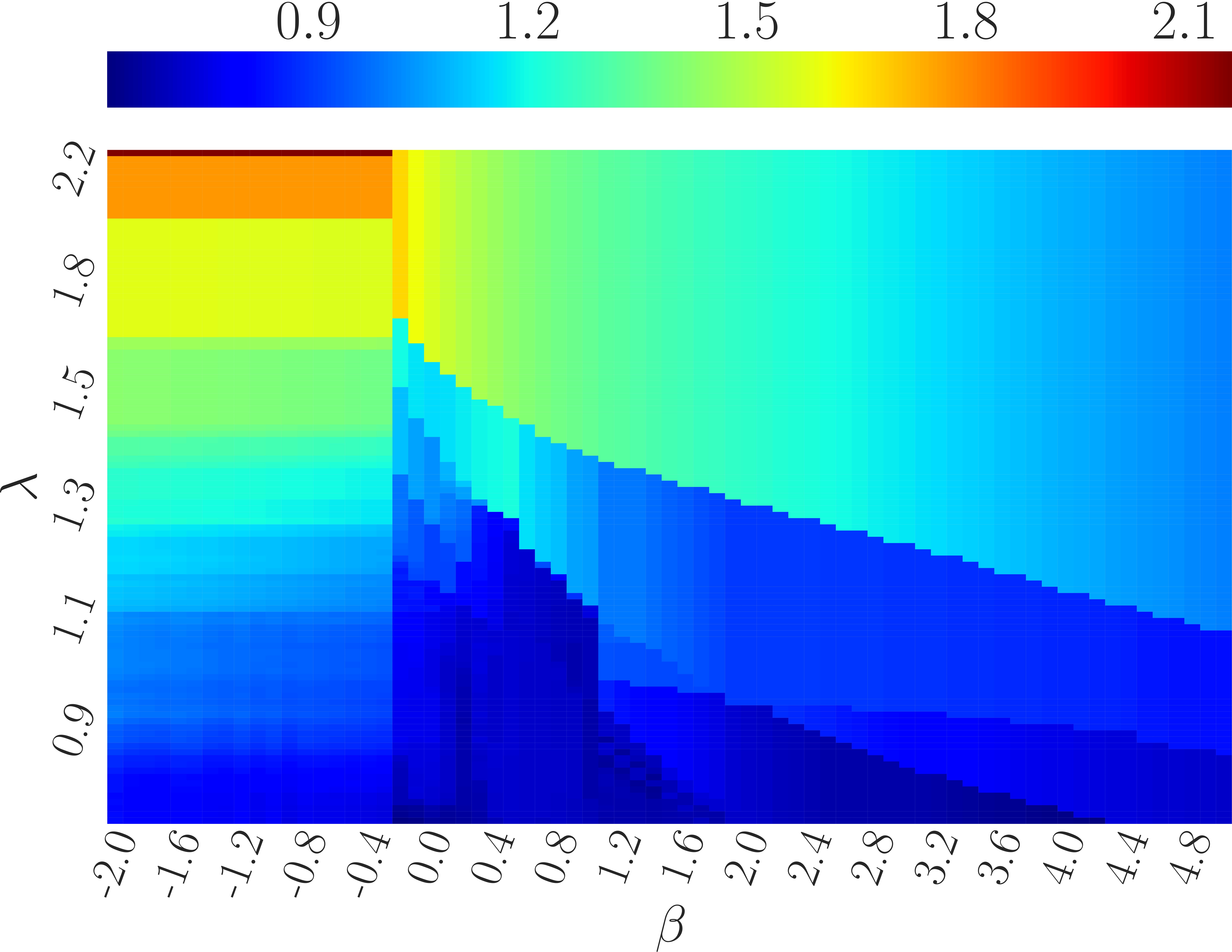}
		}
\vspace{-10pt}
		\caption{HeartK2, pow mean \label{fig:heartK2_pow} }
\vspace{-10pt}

\centering
		\subfloat[Number of clusters \label{fig:under_lim_min_ave}]{
 			 \includegraphics[width=47mm]{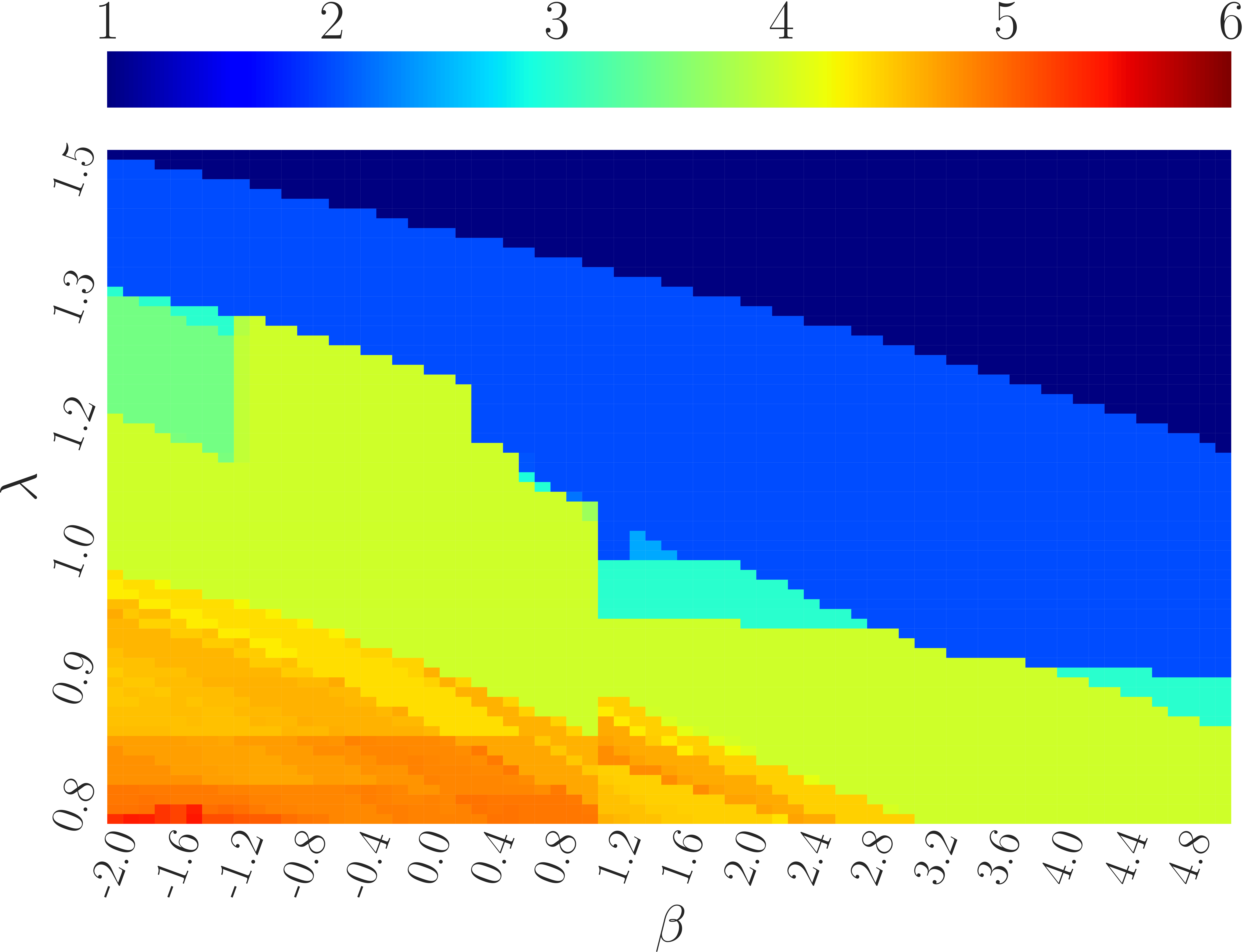}
		}
		\subfloat[NMI \label{fig:under_lim_min_max}]{
			 \includegraphics[width=47mm]{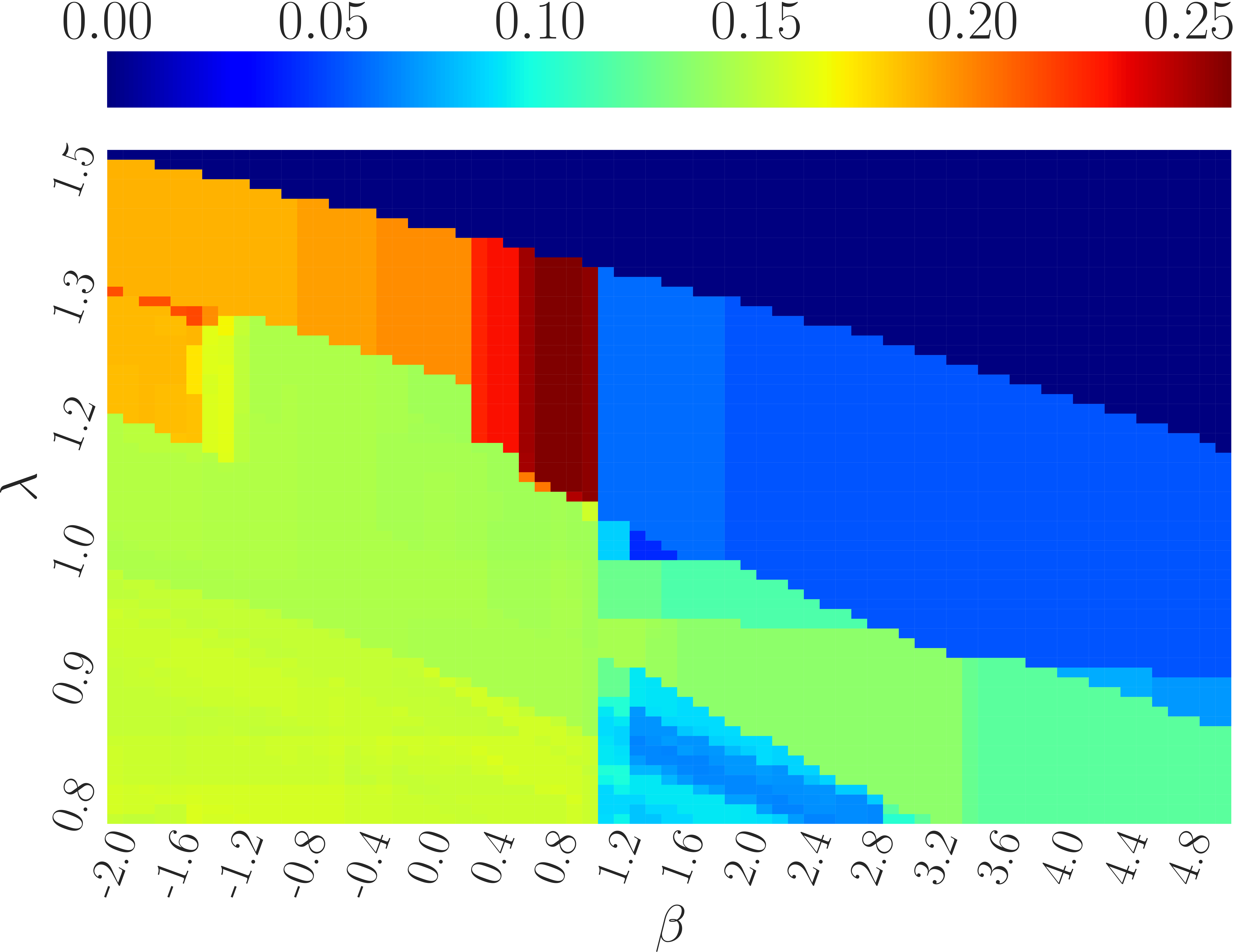}
		}
             \subfloat[Maximum distortion \label{fig:under_lim_min_max}]{
			 \includegraphics[width=47mm]{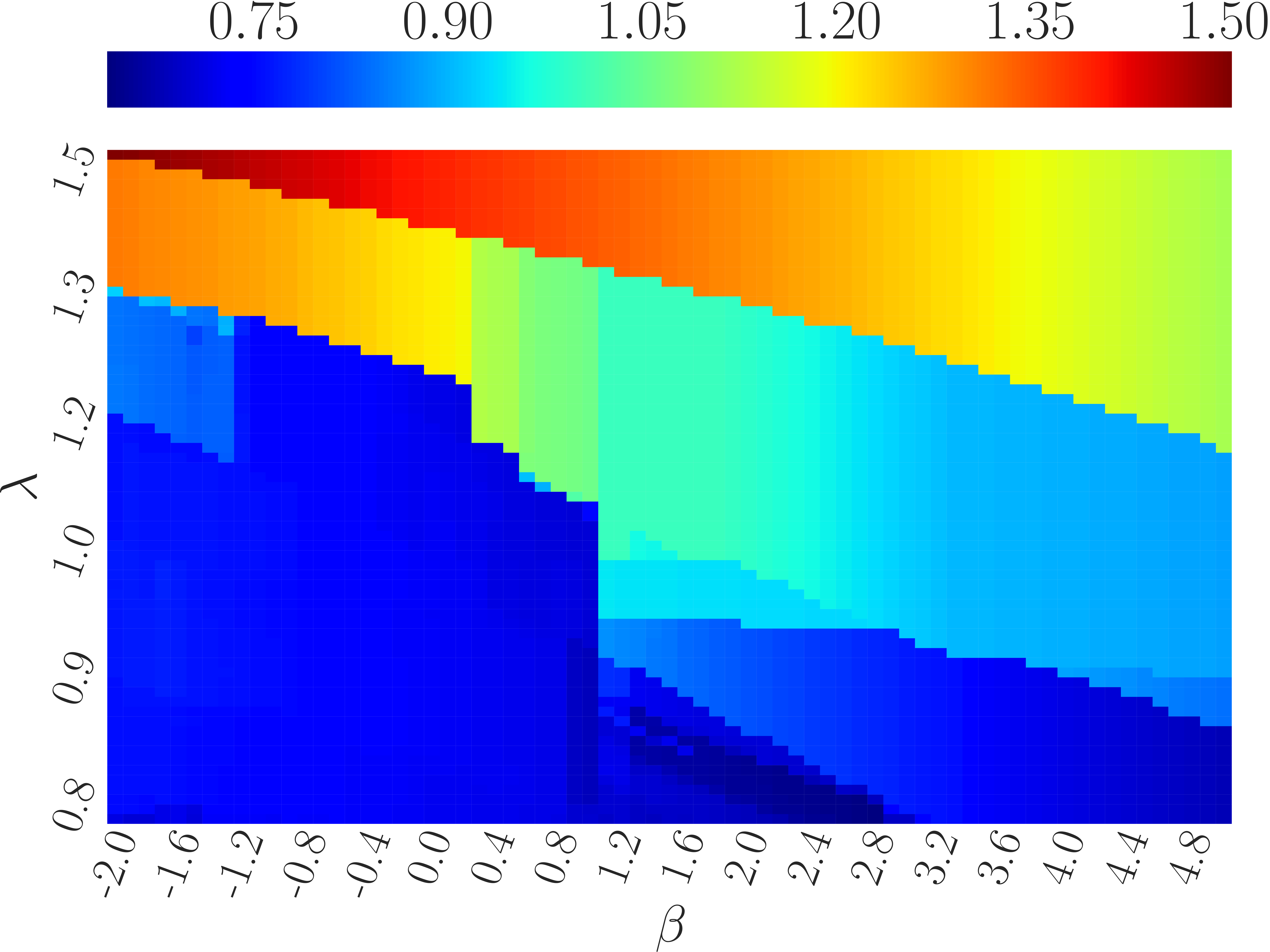}
		}
\vspace{-10pt}
		\caption{HeartK2, log-sum-exp \label{fig:HeartK2_exp} }
\vspace{-10pt}
\end{figure*}

\begin{figure*}[htbp]
\centering
		\subfloat[Number of clusters \label{fig:under_lim_min_ave}]{
 			 \includegraphics[width=47mm]{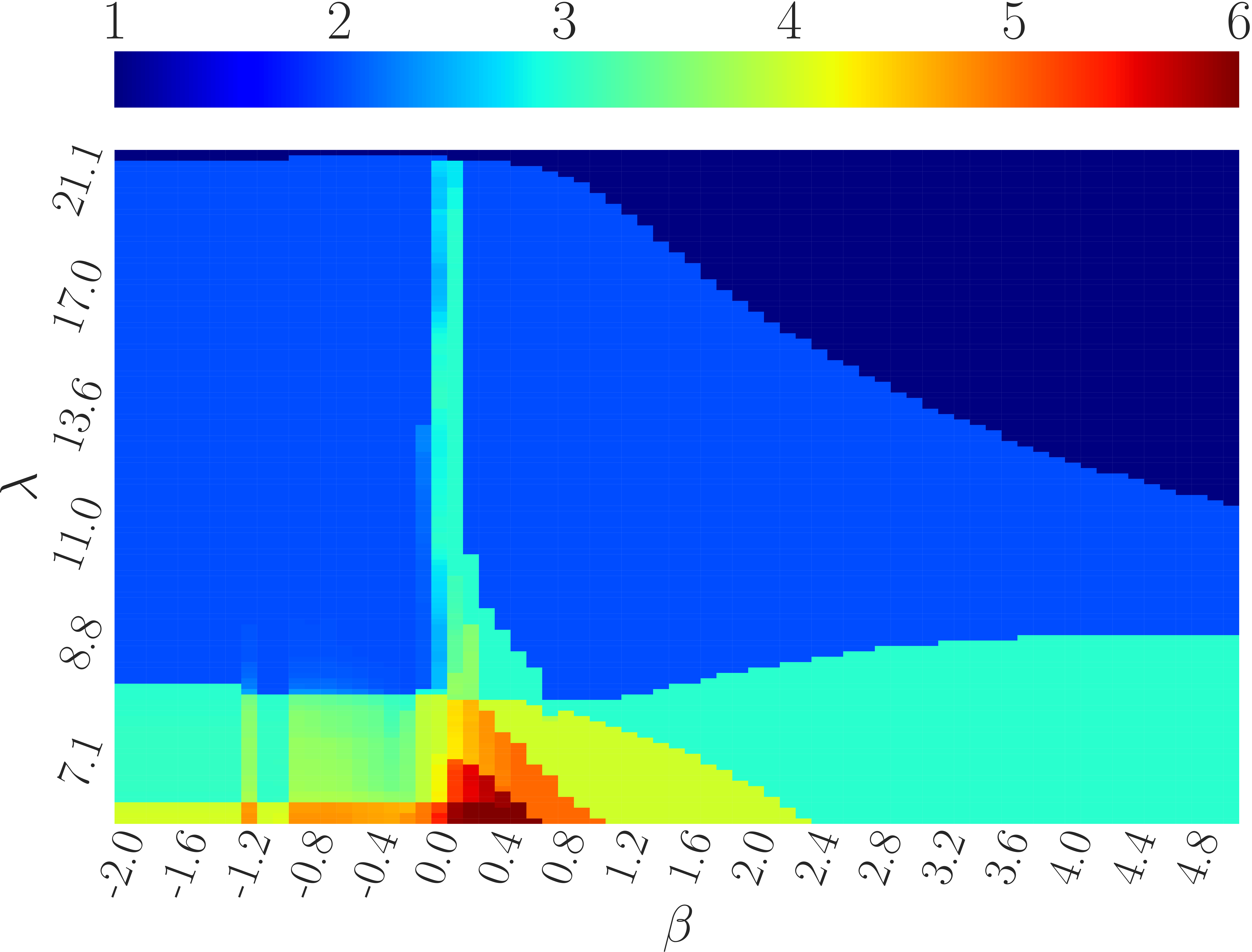}
		}
		\subfloat[NMI \label{fig:under_lim_min_max}]{
			 \includegraphics[width=47mm]{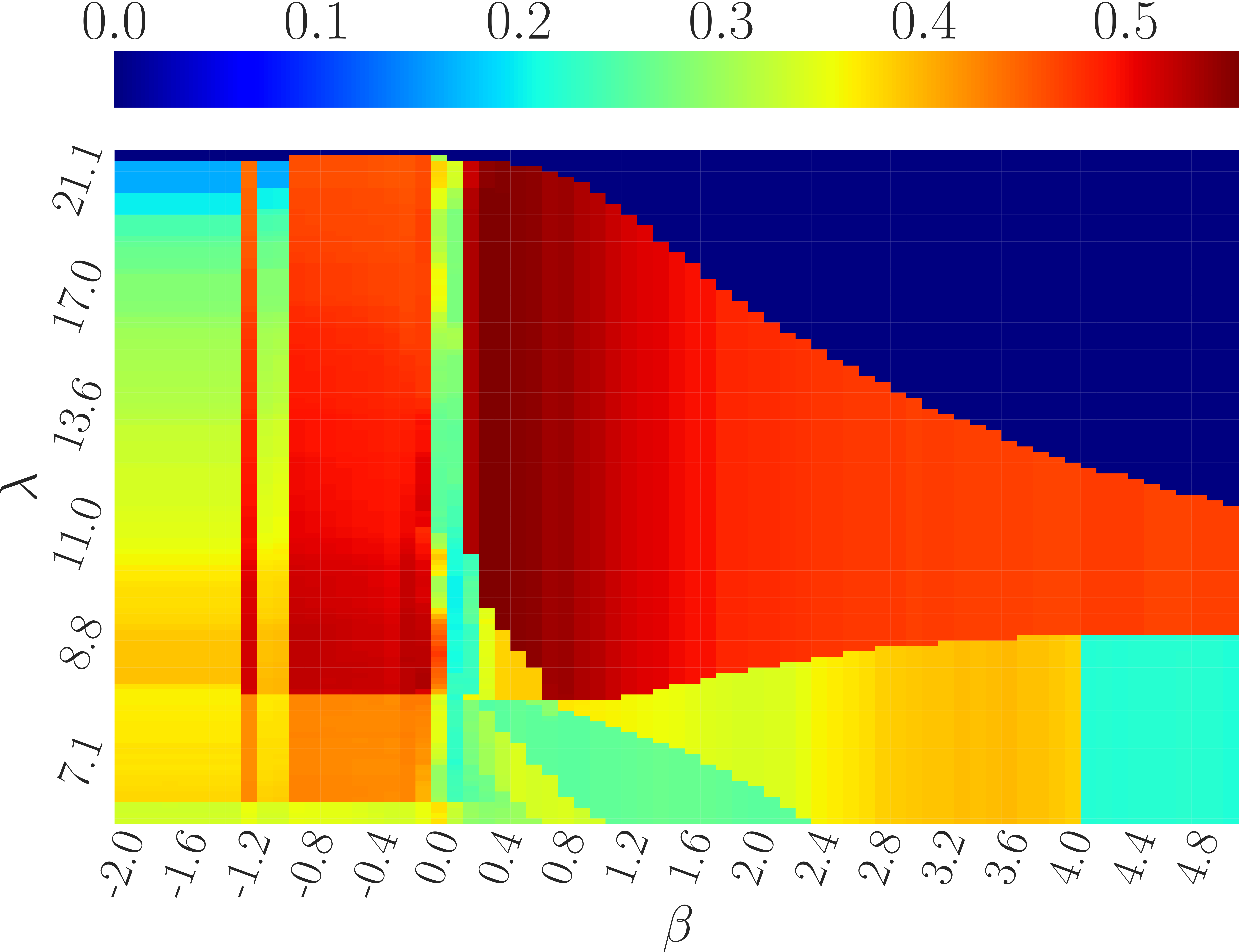}
		}
             \subfloat[Maximum distortion \label{fig:under_lim_min_max}]{
			 \includegraphics[width=47mm]{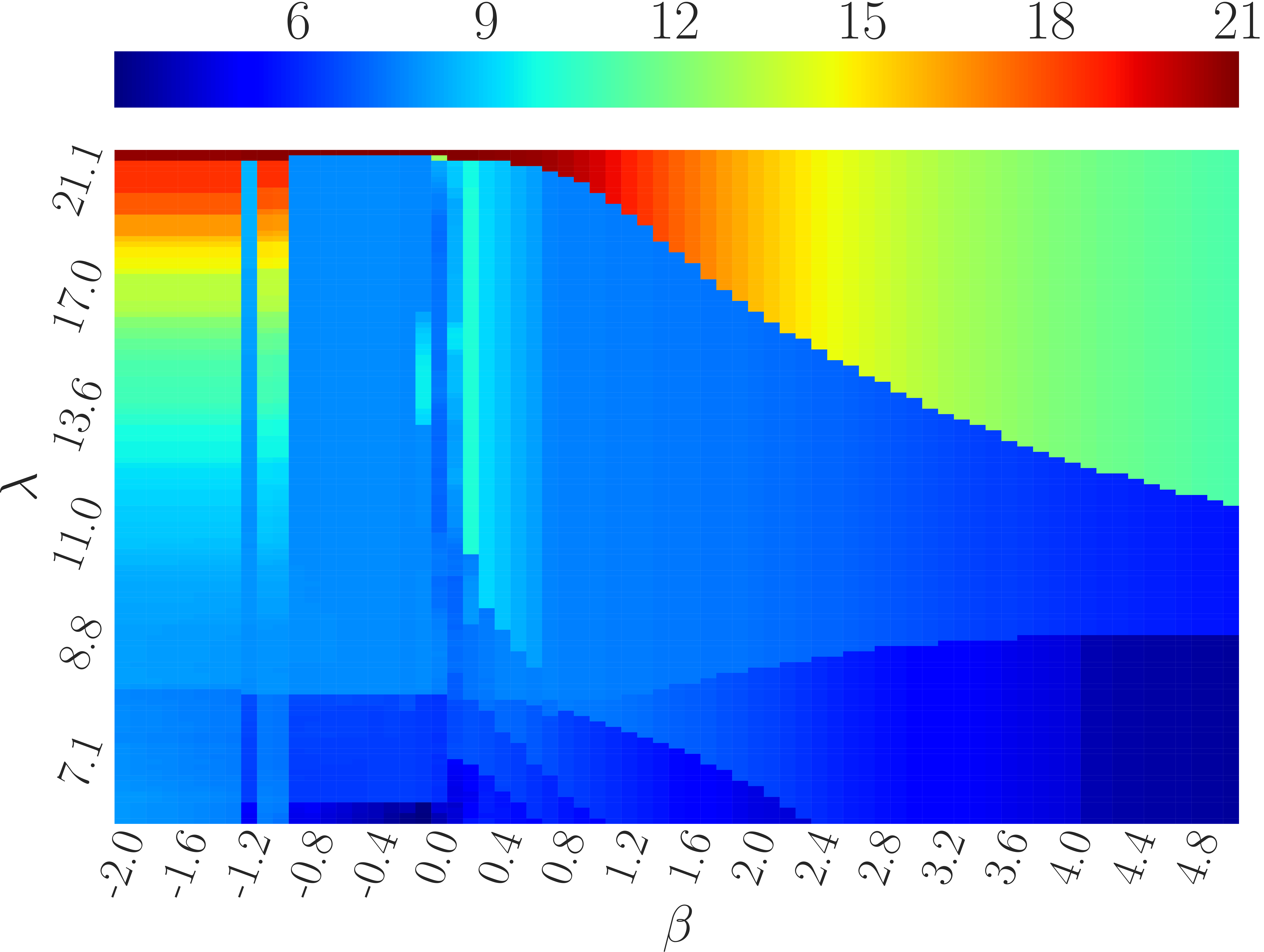}
		}
\vspace{-10pt}
		\caption{HTRU2, pow mean \label{fig:htru2_pow} }
\vspace{-10pt}

\centering
		\subfloat[Number of clusters \label{fig:under_lim_min_ave}]{
 			 \includegraphics[width=47mm]{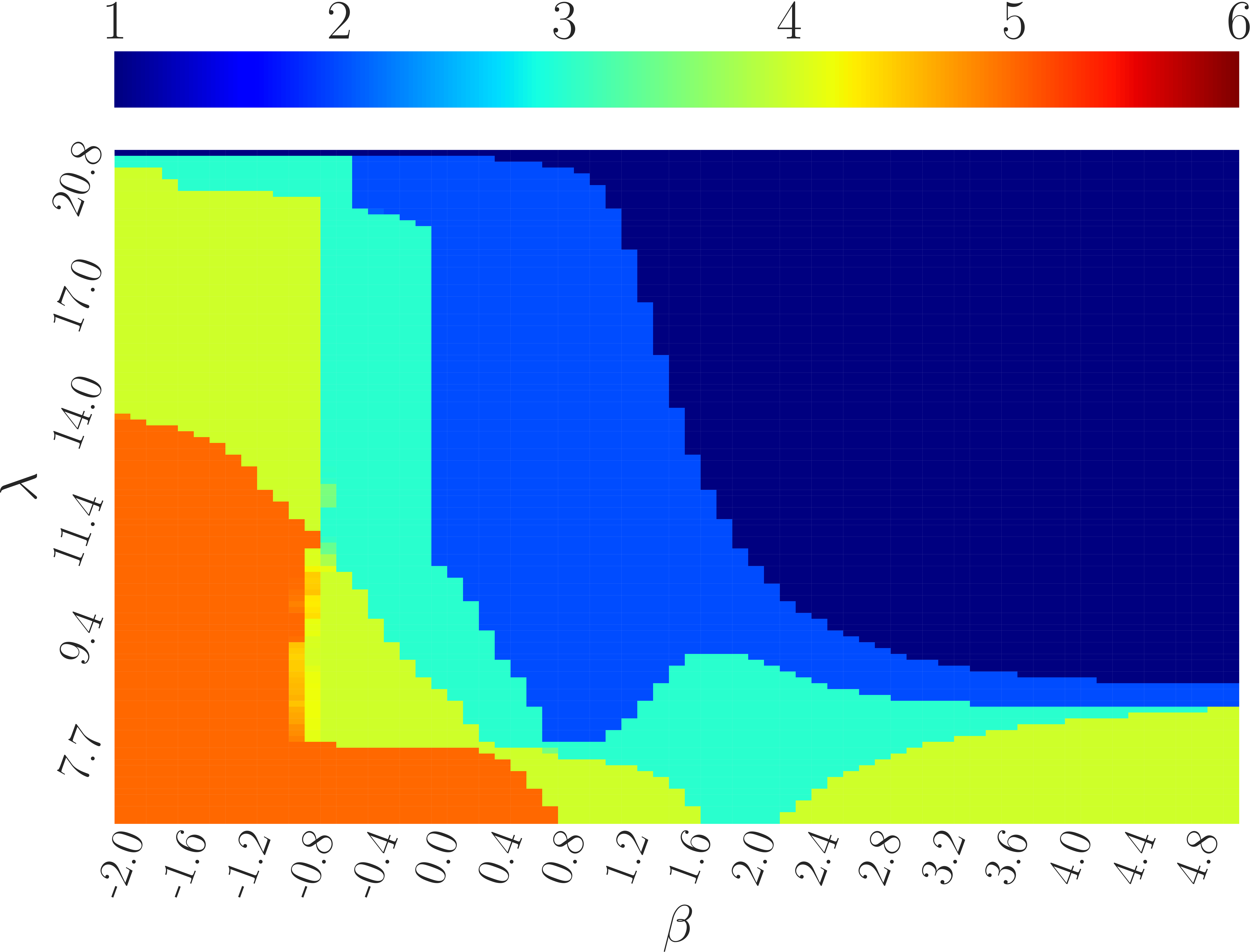}
		}
		\subfloat[NMI \label{fig:under_lim_min_max}]{
			 \includegraphics[width=47mm]{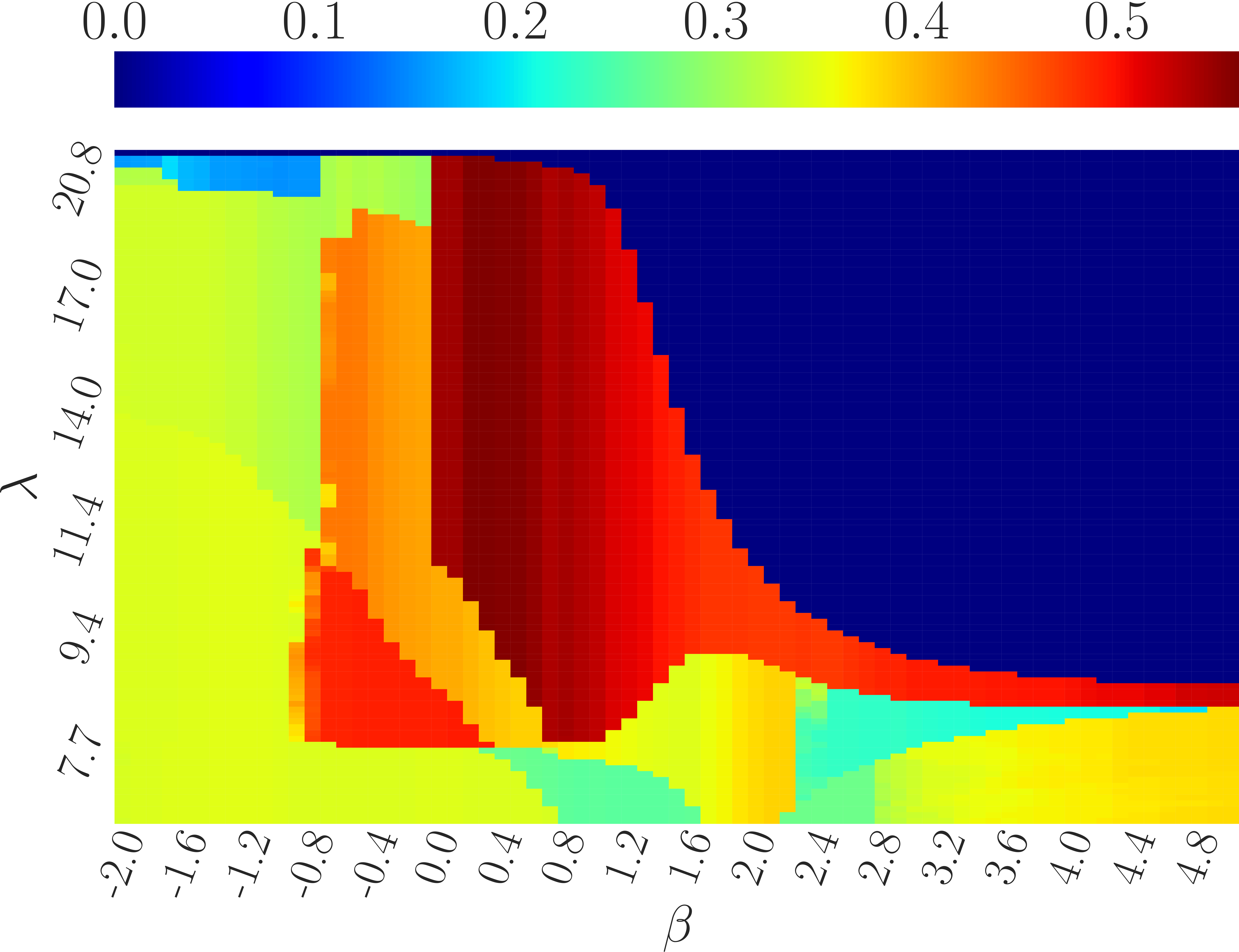}
		}
             \subfloat[Maximum distortion \label{fig:under_lim_min_max}]{
			 \includegraphics[width=47mm]{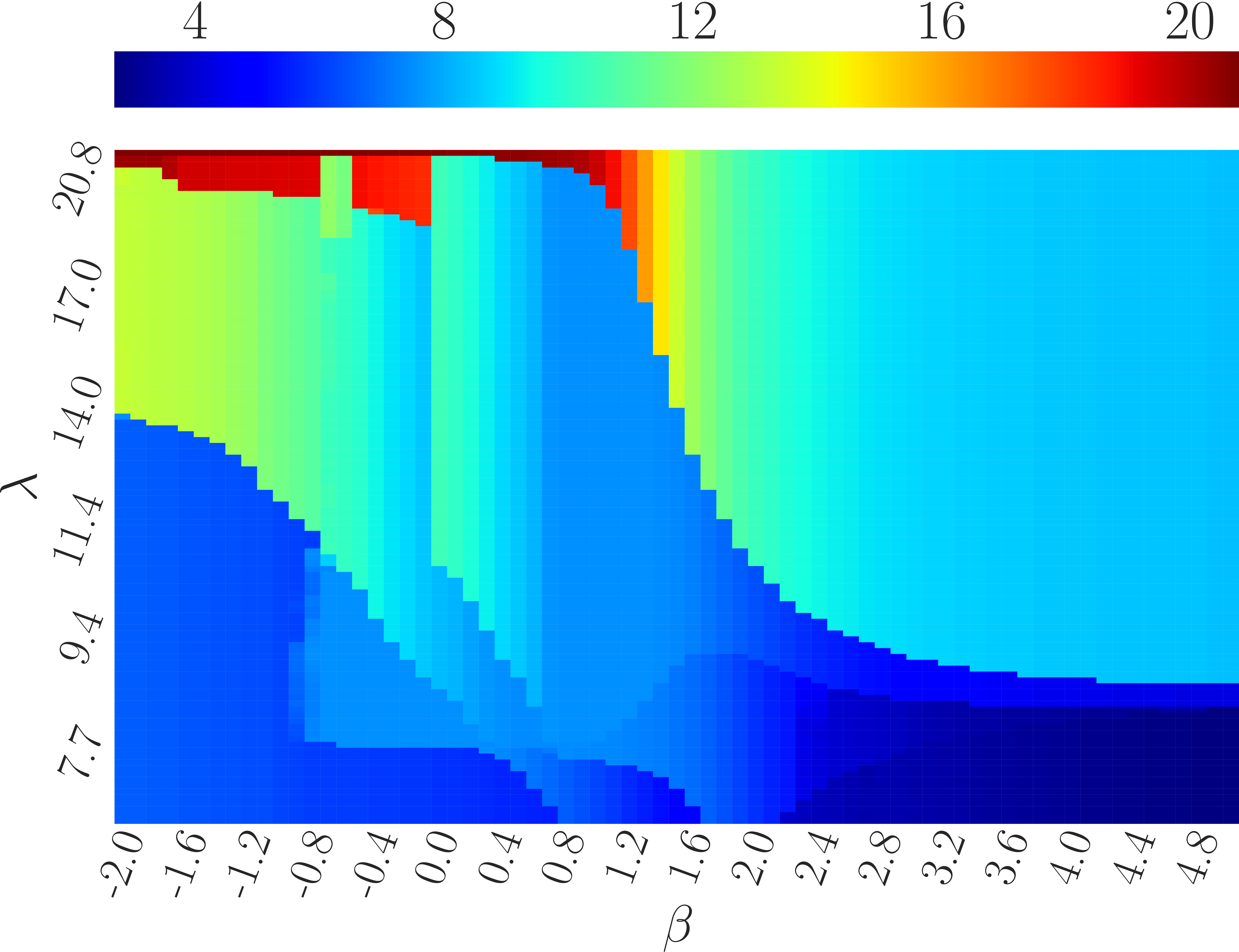}
		}
\vspace{-10pt}
		\caption{HTRU2, log-sum-exp \label{fig:htru2_exp} }
\vspace{-10pt}
\end{figure*}

\begin{figure*}[htbp]
\centering
		\subfloat[Number of clusters \label{fig:under_lim_min_ave}]{
 			 \includegraphics[width=47mm]{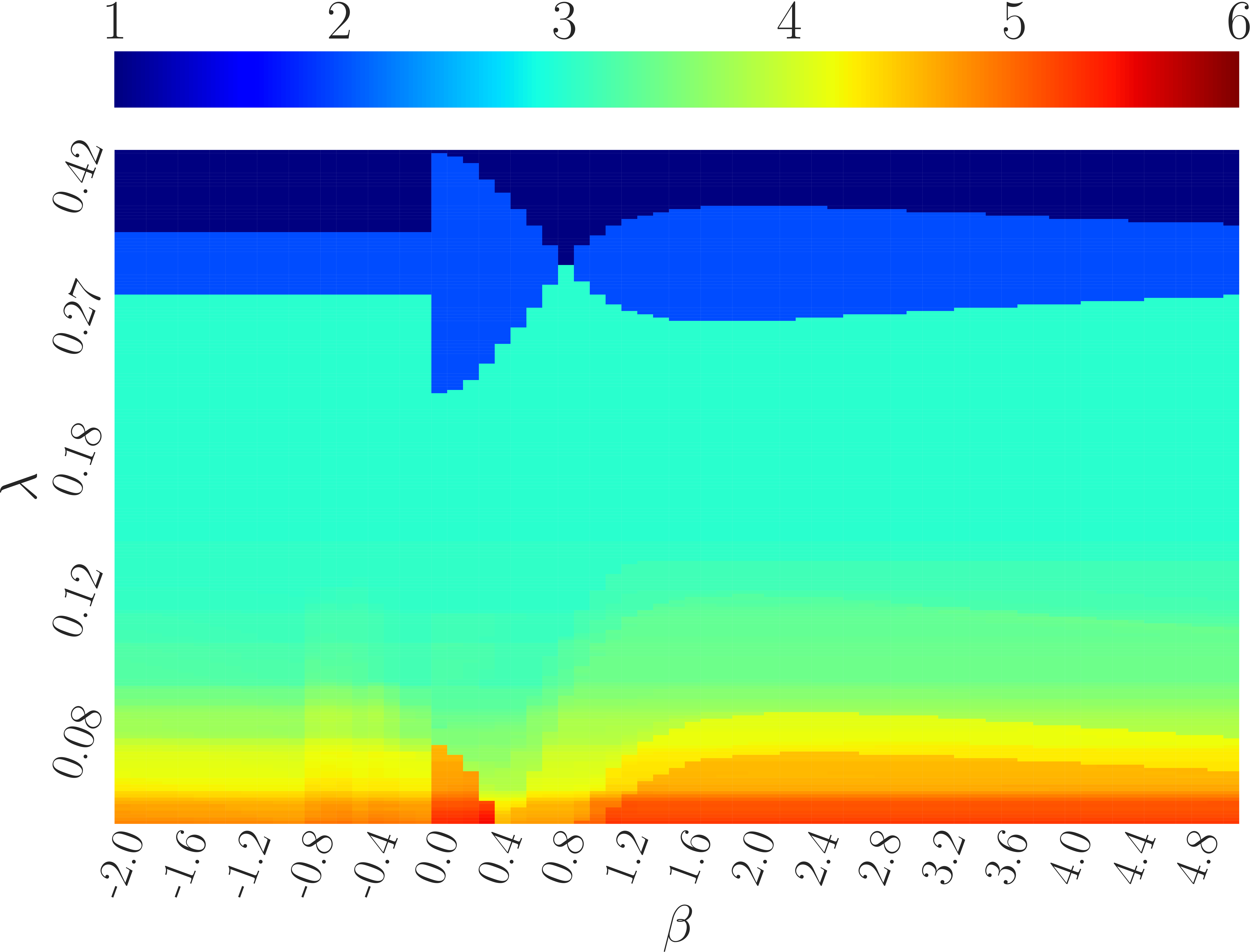}
		}
		\subfloat[NMI \label{fig:under_lim_min_max}]{
			 \includegraphics[width=47mm]{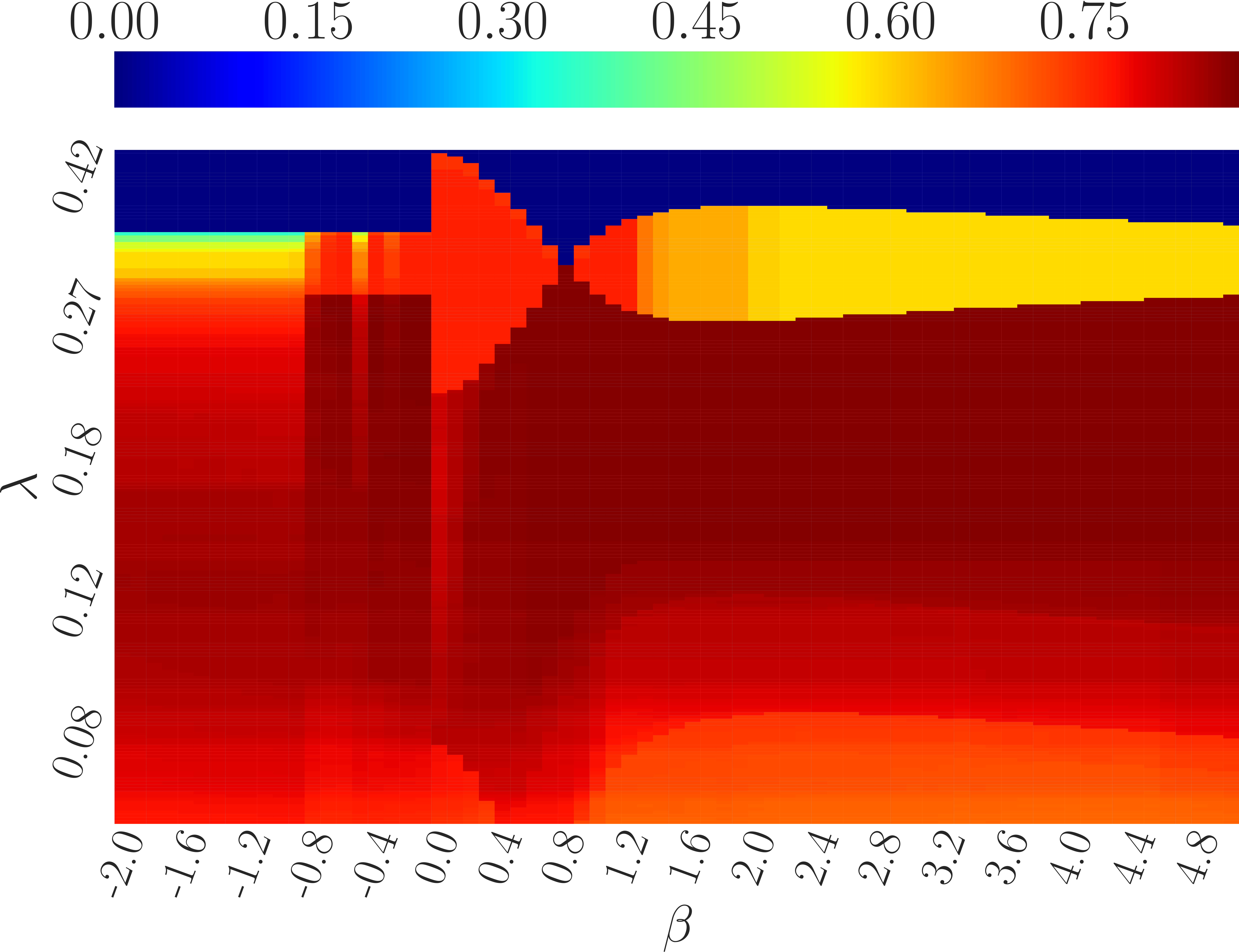}
		}
             \subfloat[Maximum distortion \label{fig:under_lim_min_max}]{
			 \includegraphics[width=47mm]{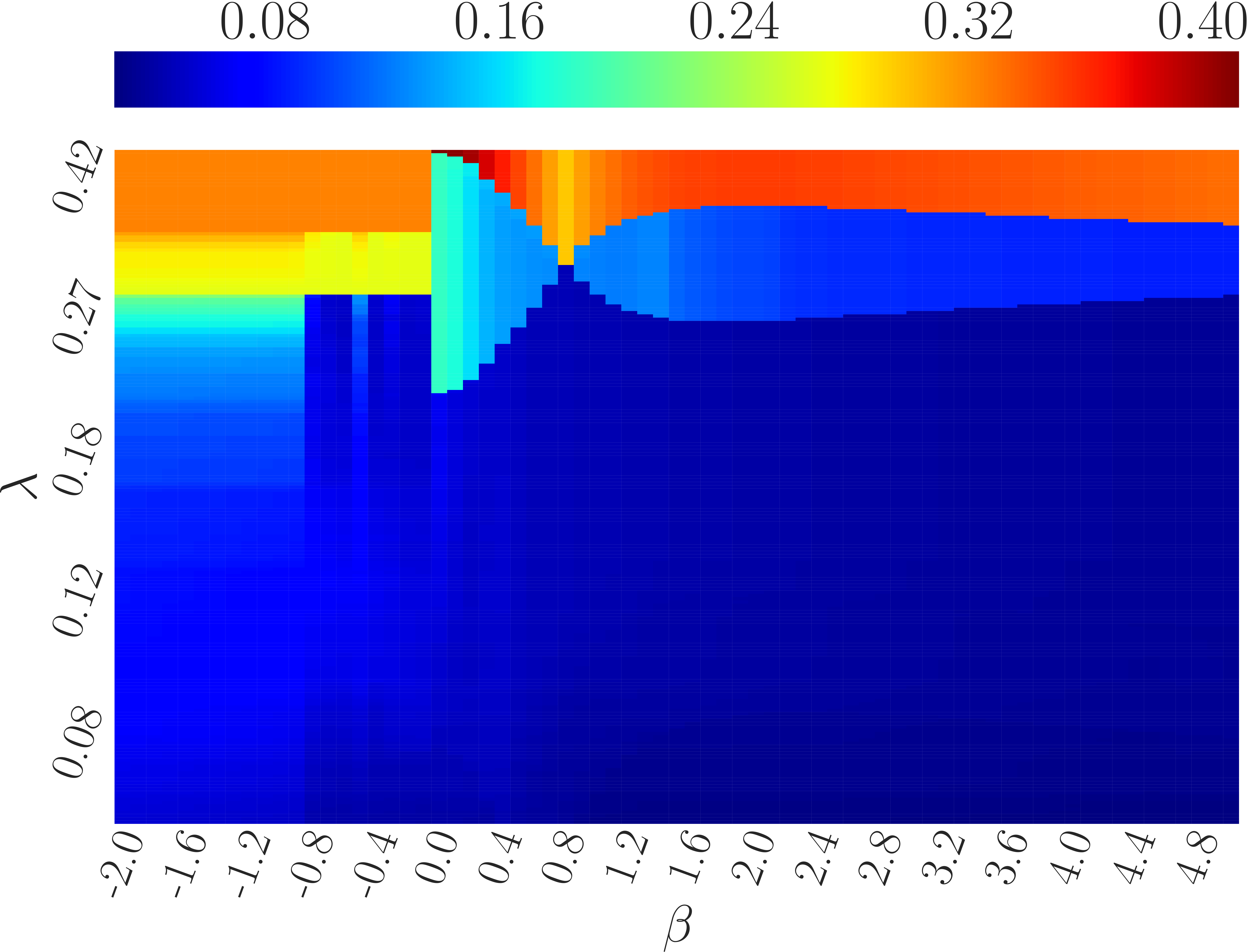}
		}
\vspace{-10pt}
		\caption{Iris, pow mean \label{fig:iris_pow} }
\vspace{-10pt}

\centering
		\subfloat[Number of clusters \label{fig:under_lim_min_ave}]{
 			 \includegraphics[width=47mm]{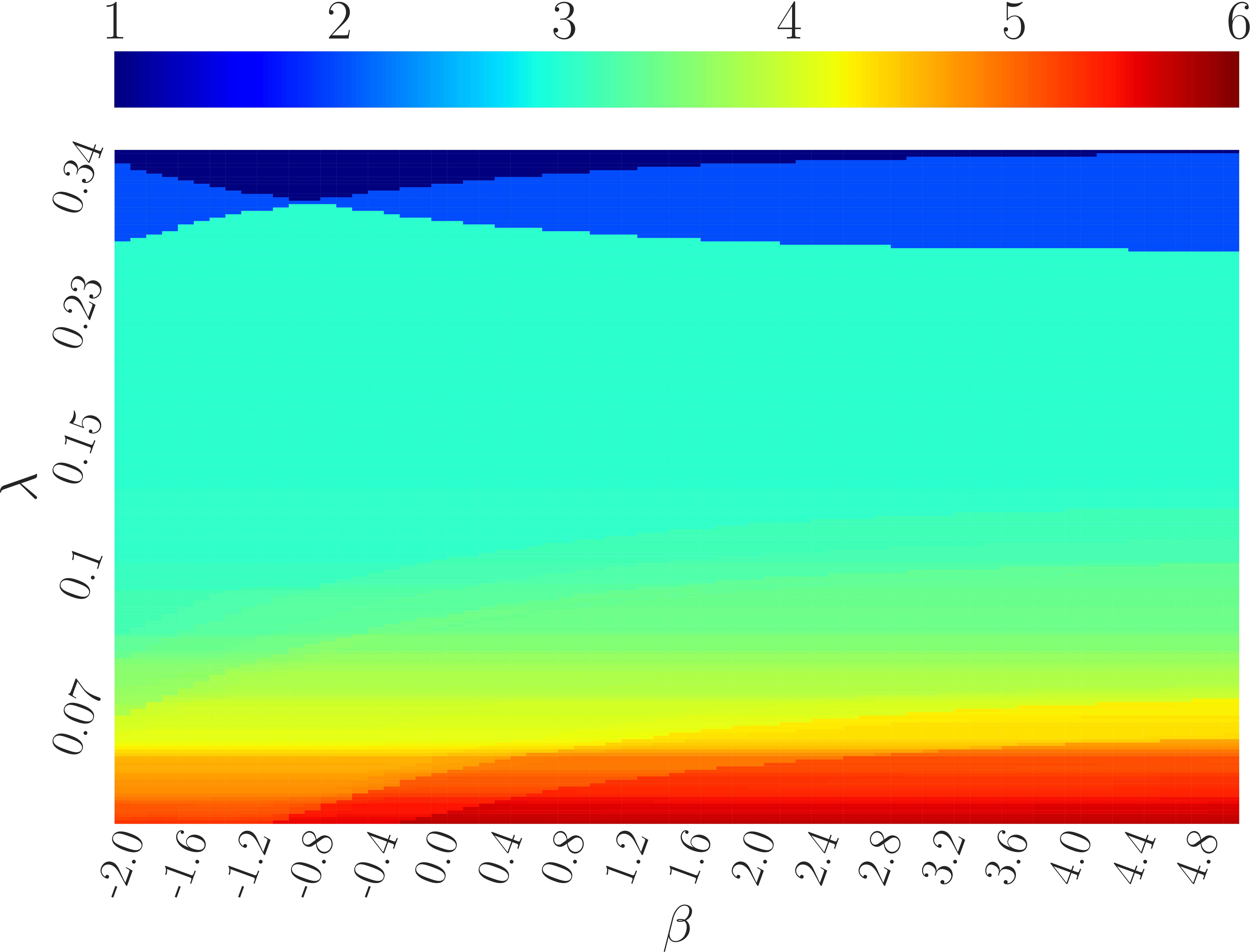}
		}
		\subfloat[NMI \label{fig:under_lim_min_max}]{
			 \includegraphics[width=47mm]{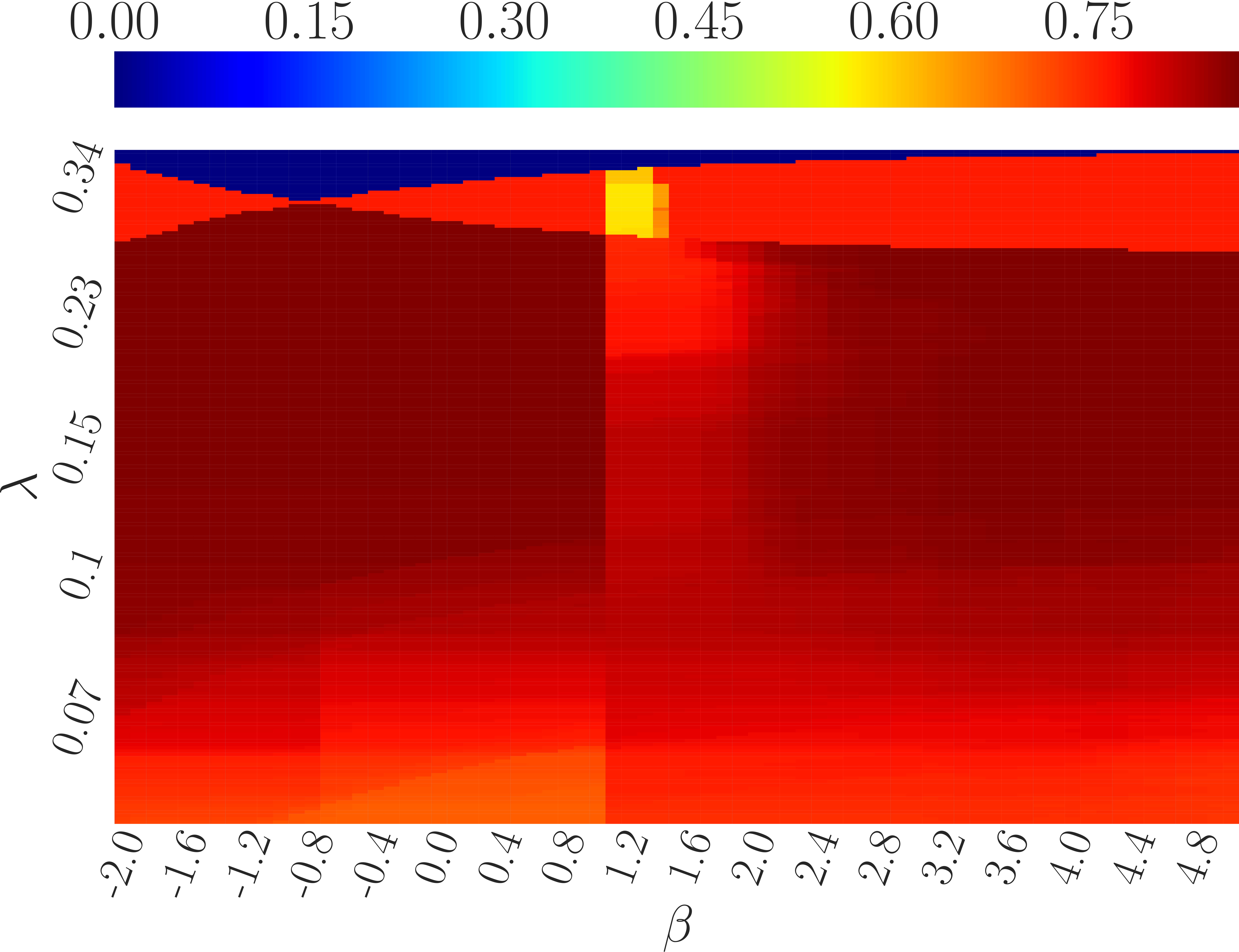}
		}
             \subfloat[Maximum distortion \label{fig:under_lim_min_max}]{
			 \includegraphics[width=47mm]{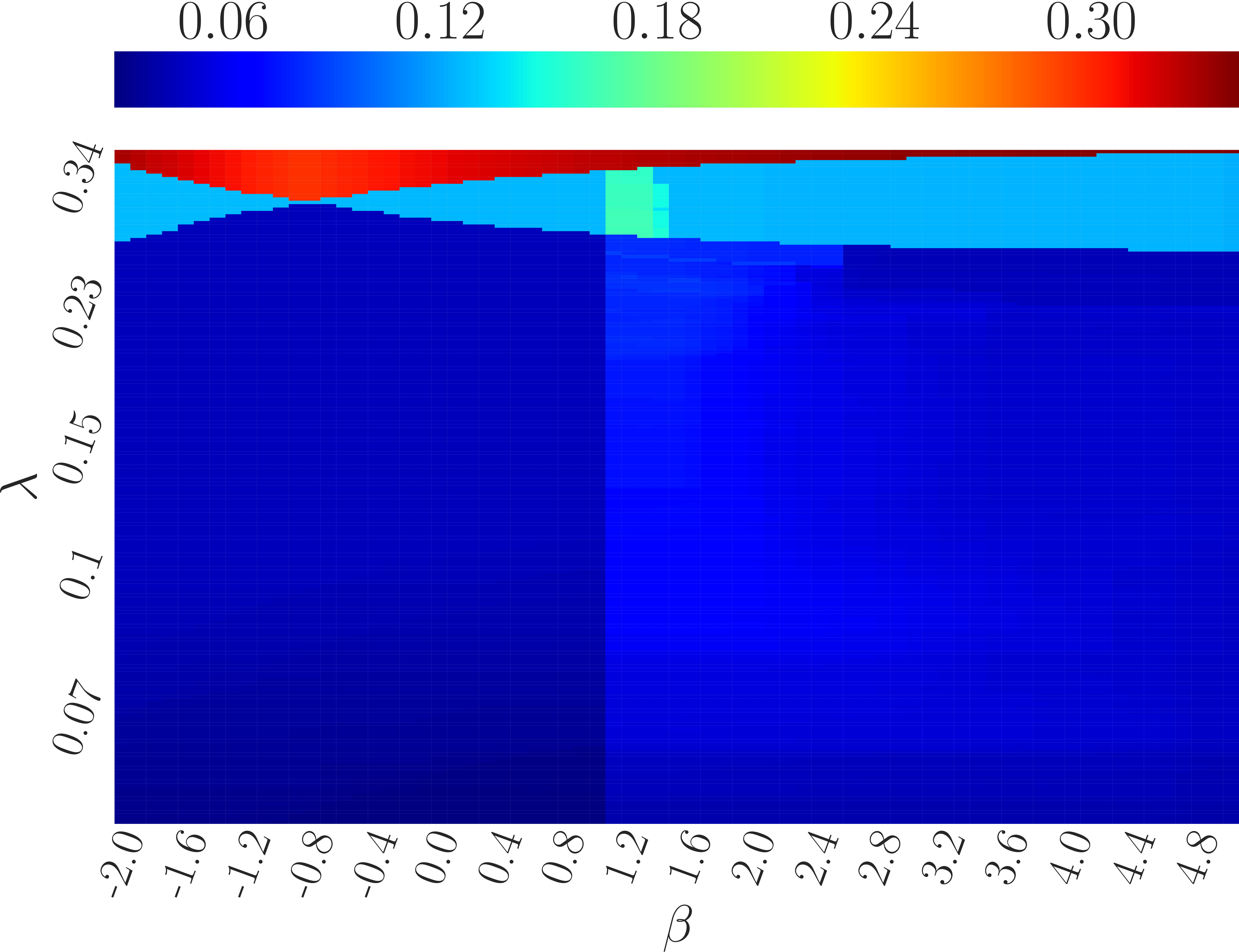}
		}
\vspace{-10pt}
		\caption{Iris, log-sum-exp \label{fig:iris_exp} }
\vspace{-10pt}
\end{figure*}

\begin{figure*}[htbp]
\centering
		\subfloat[Number of clusters \label{fig:under_lim_min_ave}]{
 			 \includegraphics[width=47mm]{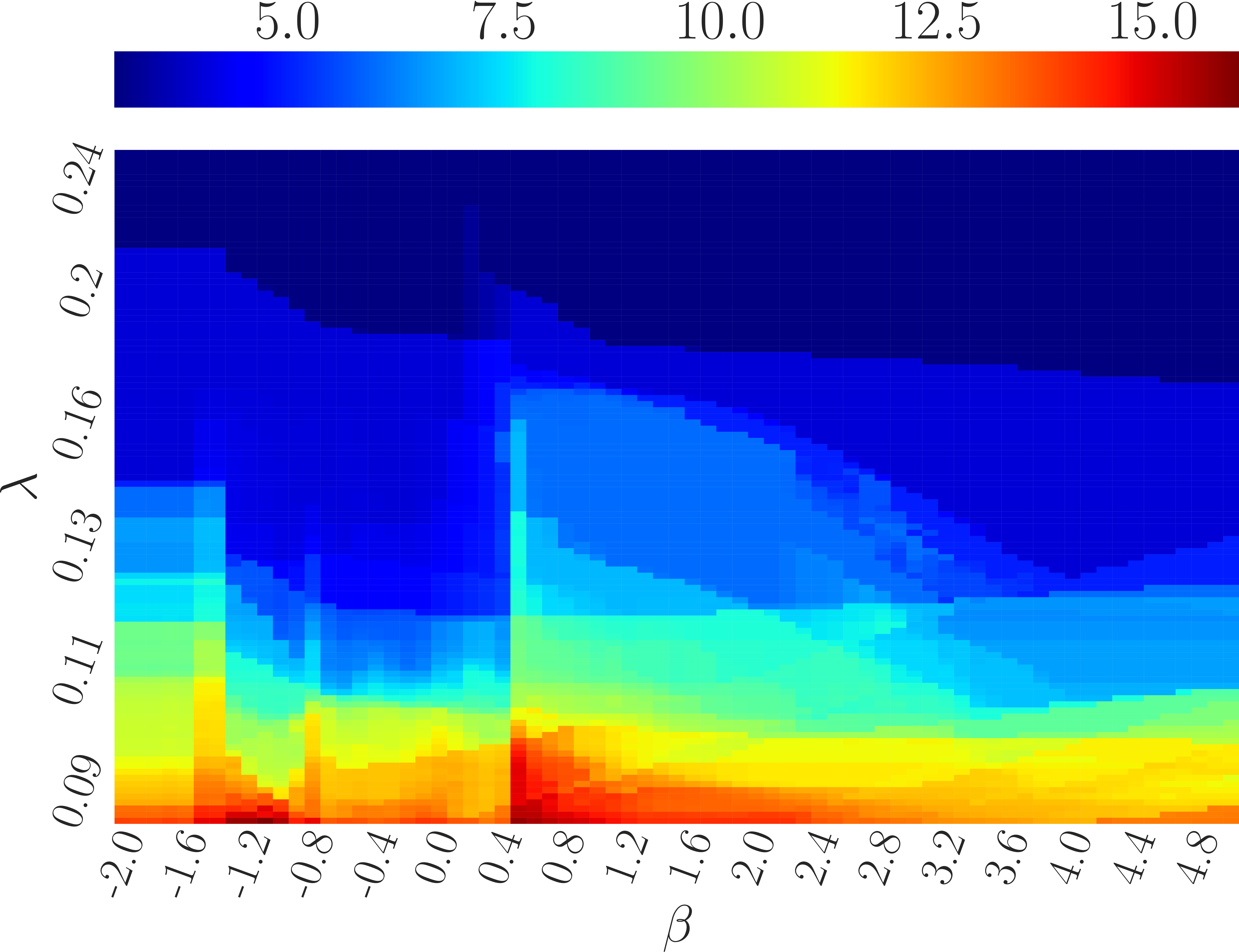}
		}
		\subfloat[NMI \label{fig:under_lim_min_max}]{
			 \includegraphics[width=47mm]{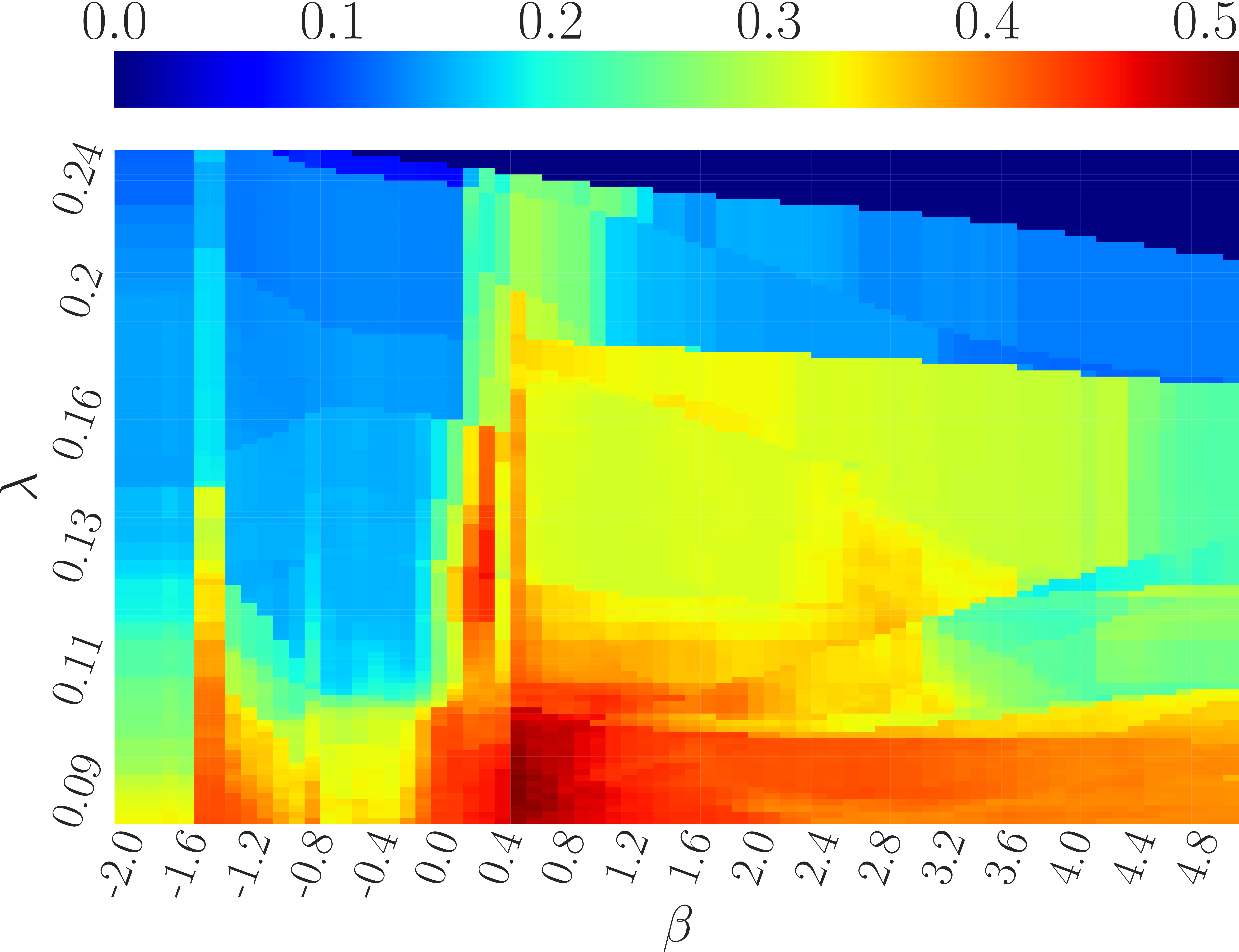}
		}
             \subfloat[Maximum distortion \label{fig:under_lim_min_max}]{
			 \includegraphics[width=48mm]{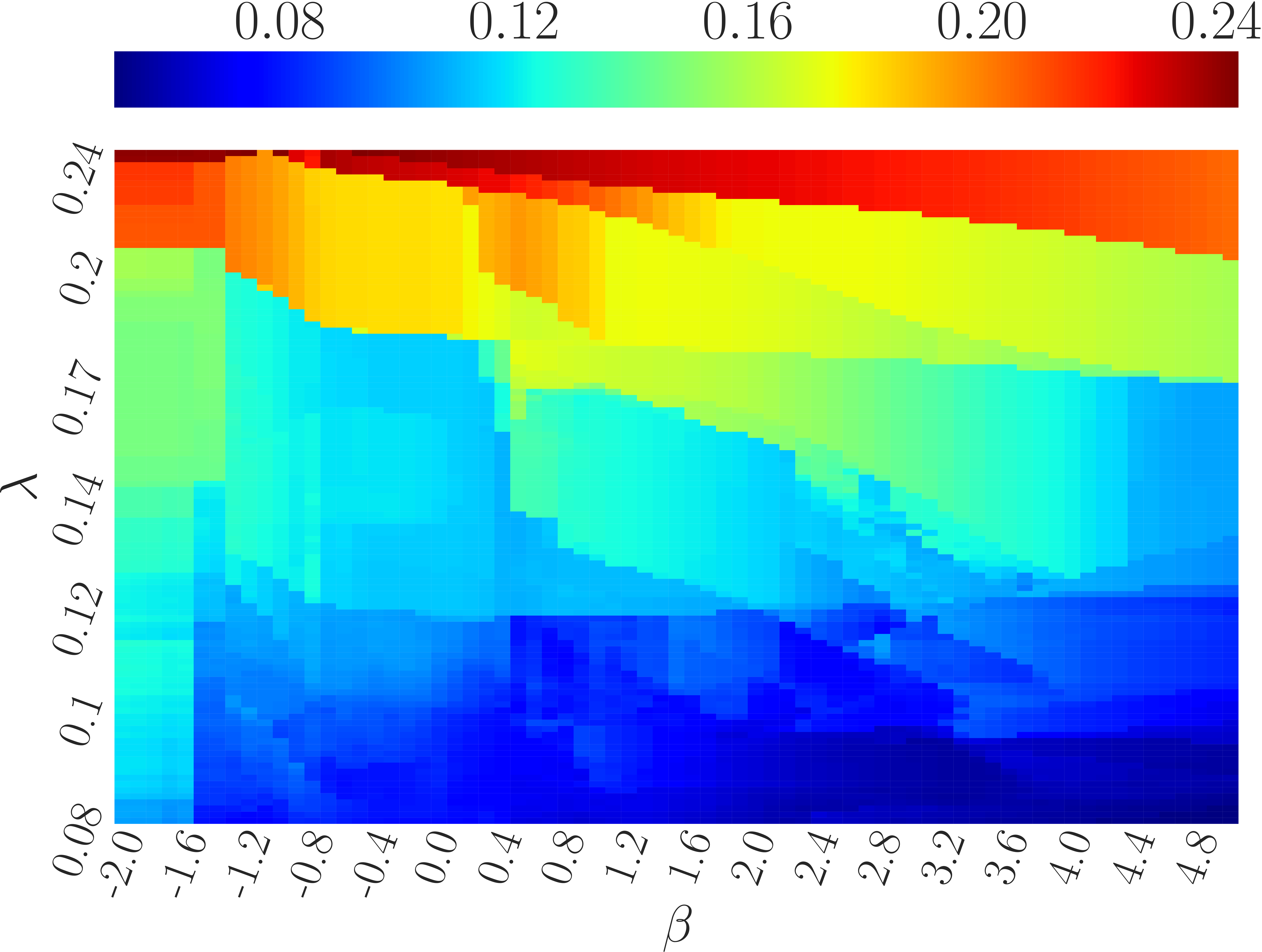}
		}
\vspace{-10pt}
		\caption{Mice Protein Expression, pow mean \label{fig:mpe_pow} }
\vspace{-10pt}

\centering
		\subfloat[Number of clusters \label{fig:under_lim_min_ave}]{
 			 \includegraphics[width=47mm]{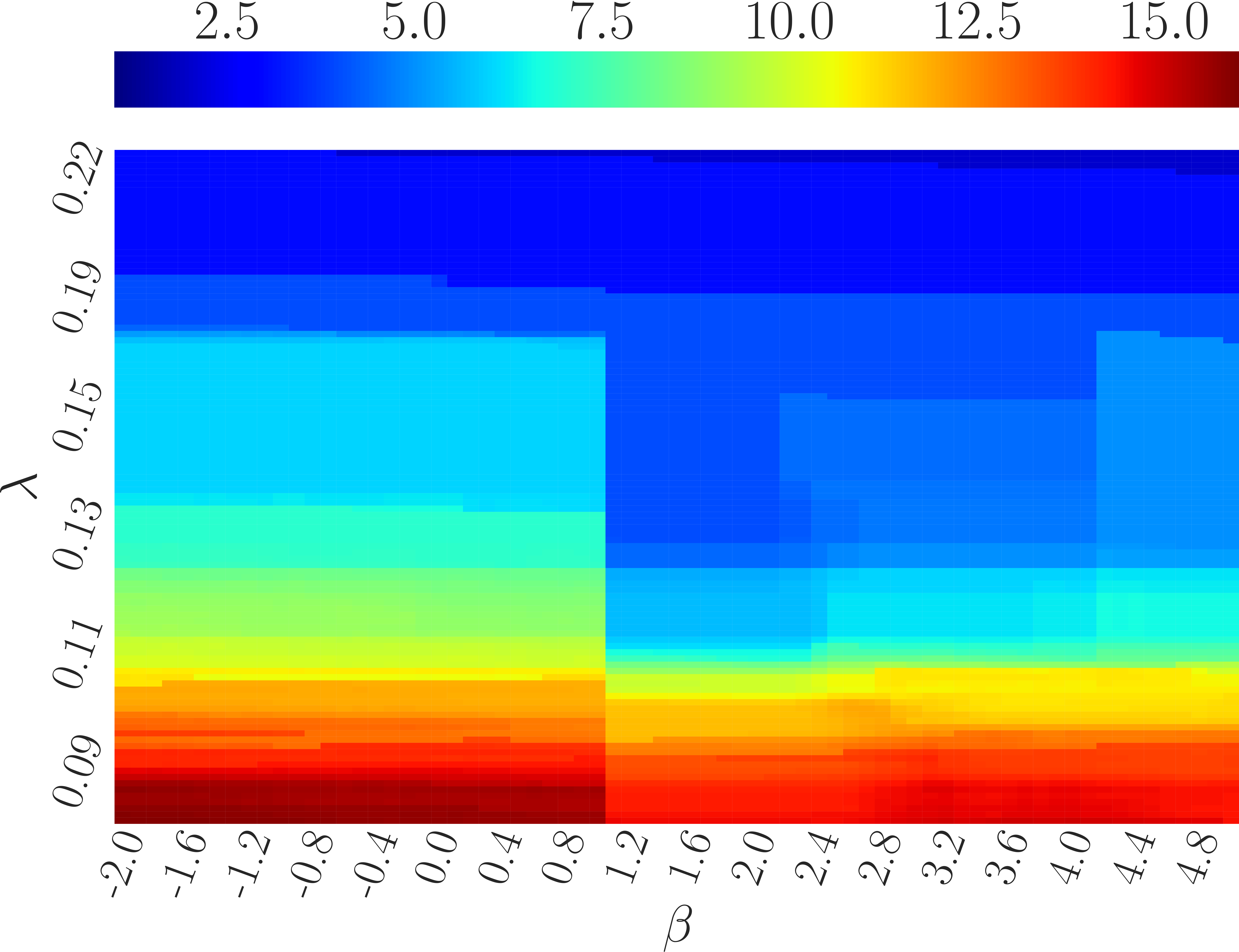}
		}
		\subfloat[NMI \label{fig:under_lim_min_max}]{
			 \includegraphics[width=48mm]{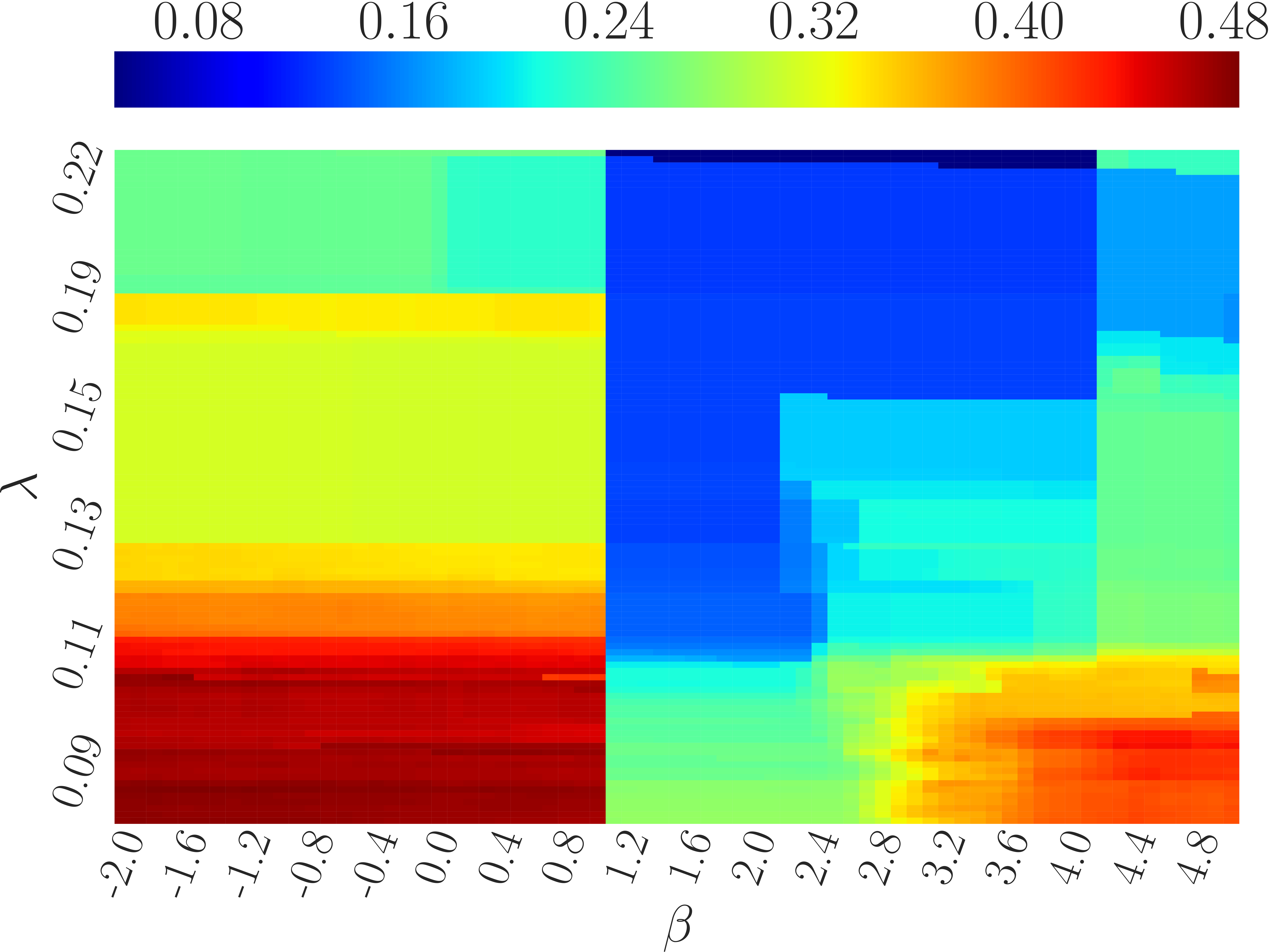}
		}
             \subfloat[Maximum distortion \label{fig:under_lim_min_max}]{
			 \includegraphics[width=47mm]{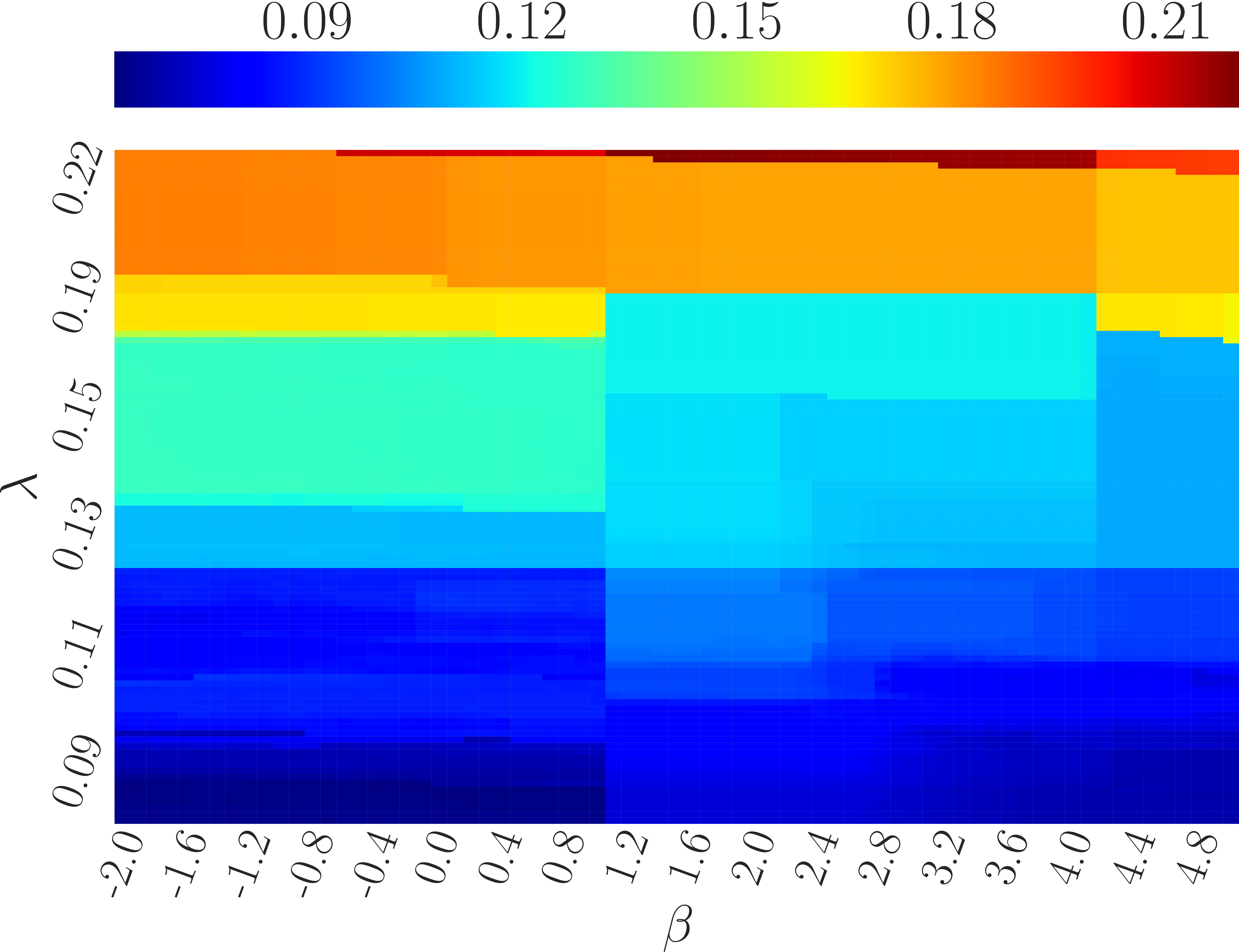}
		}
\vspace{-10pt}
		\caption{Mice Protein Expression, log-sum-exp \label{fig:mpe_exp} }
\vspace{-10pt}
\end{figure*}

\begin{figure*}[htbp]
\centering
		\subfloat[Number of clusters \label{fig:under_lim_min_ave}]{
 			 \includegraphics[width=47mm]{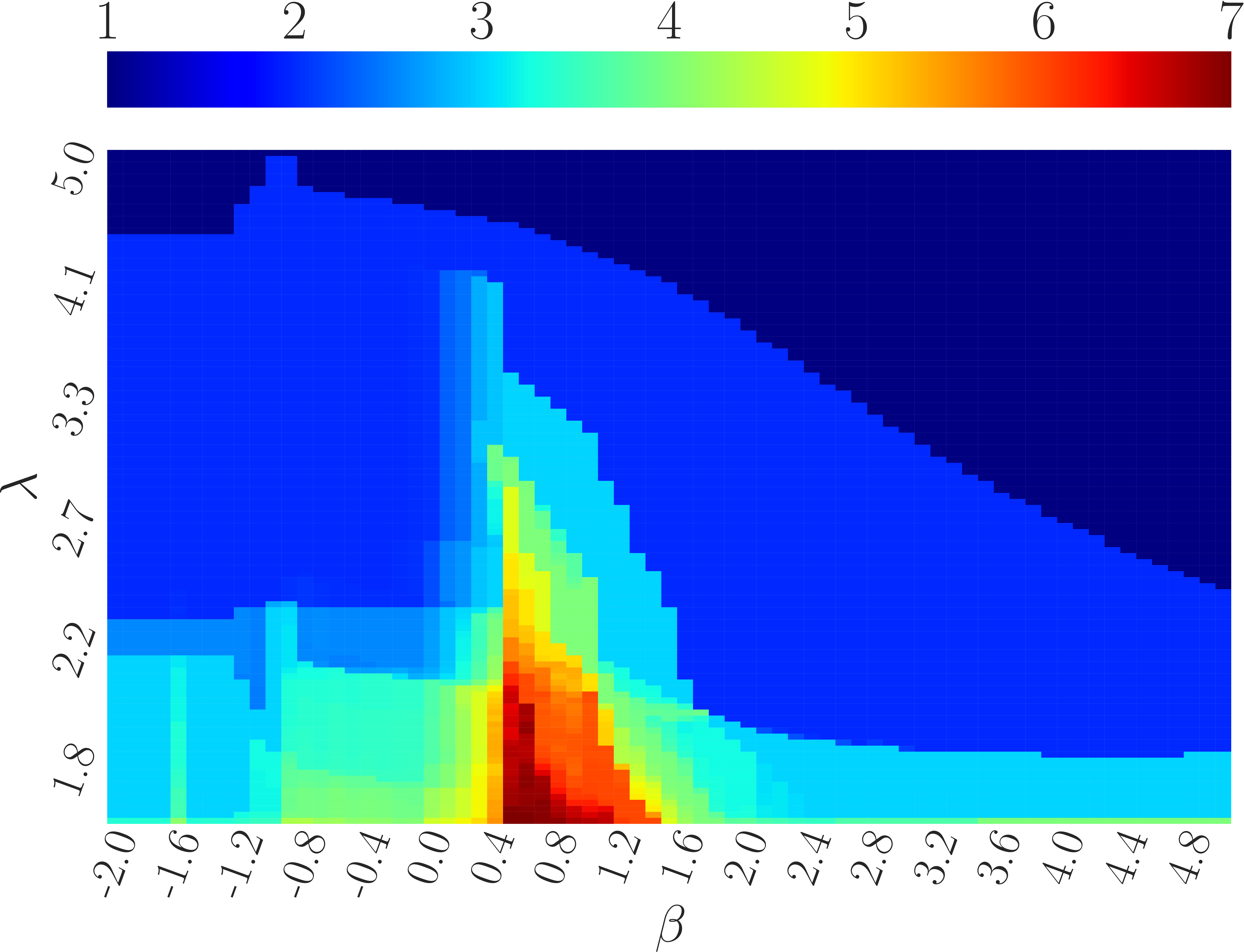}
		}
		\subfloat[NMI \label{fig:under_lim_min_max}]{
			 \includegraphics[width=47mm]{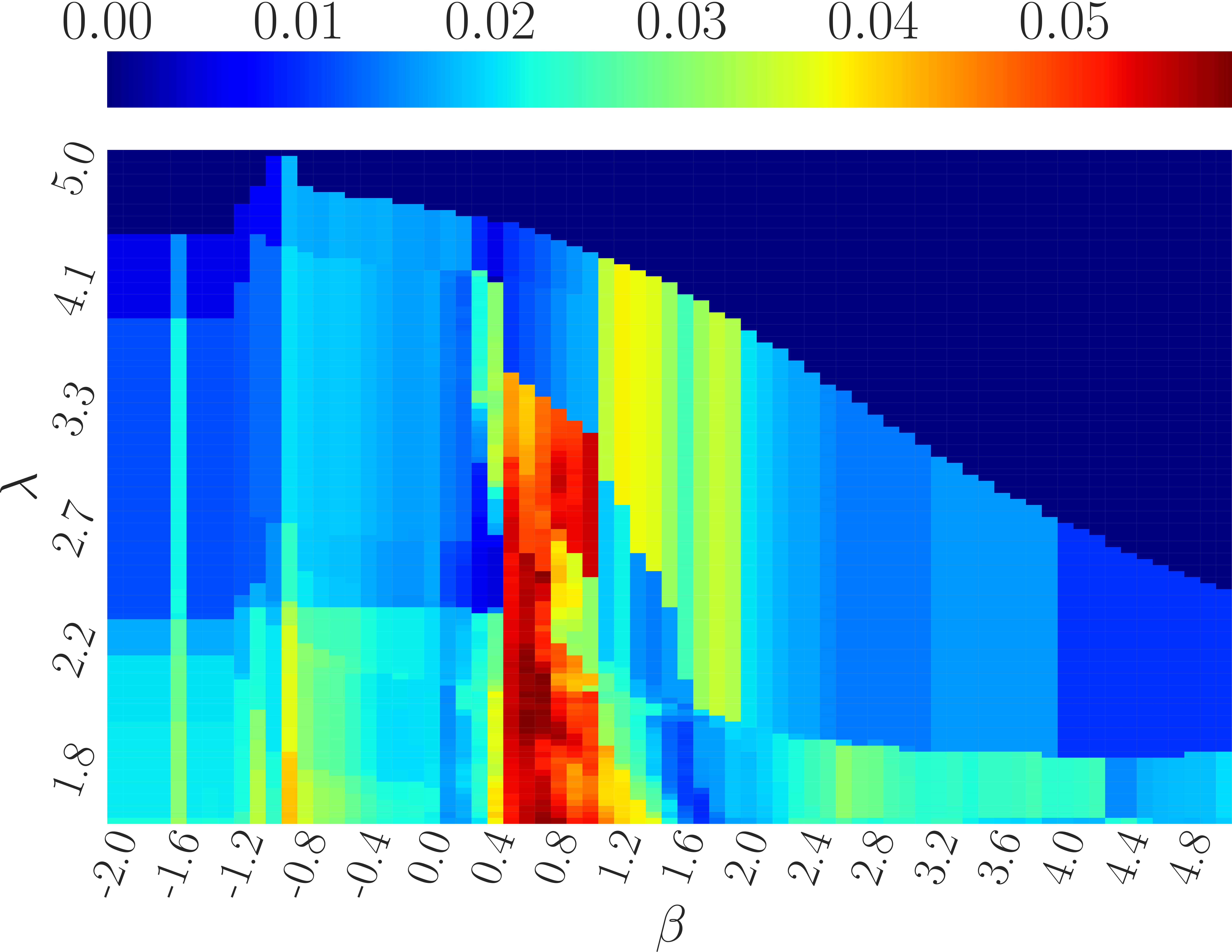}
		}
             \subfloat[Maximum distortion \label{fig:under_lim_min_max}]{
			 \includegraphics[width=47mm]{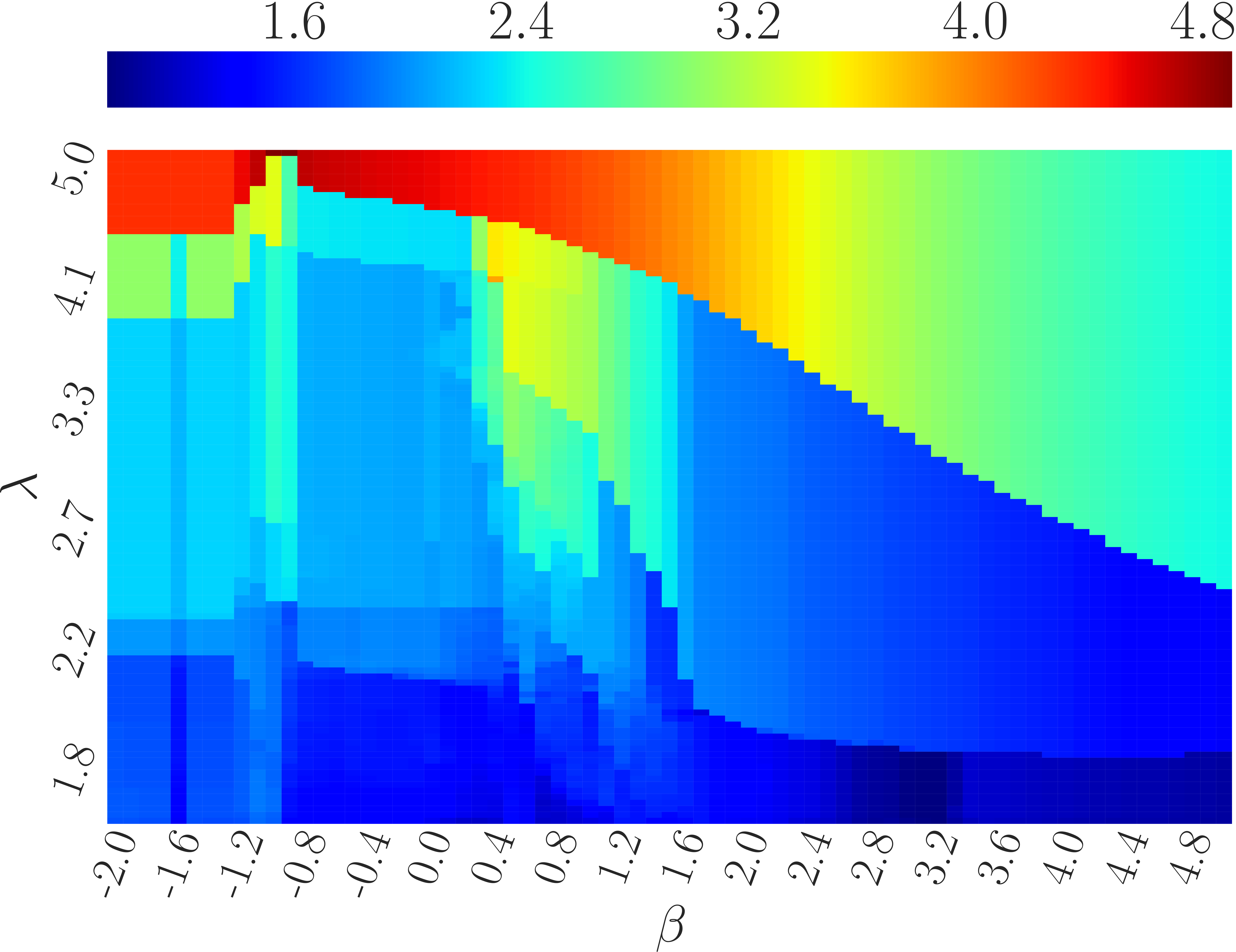}
		}
\vspace{-10pt}
		\caption{Pima, pow mean \label{fig:pima_pow} }
\vspace{-10pt}

\centering
		\subfloat[Number of clusters \label{fig:under_lim_min_ave}]{
 			 \includegraphics[width=47mm]{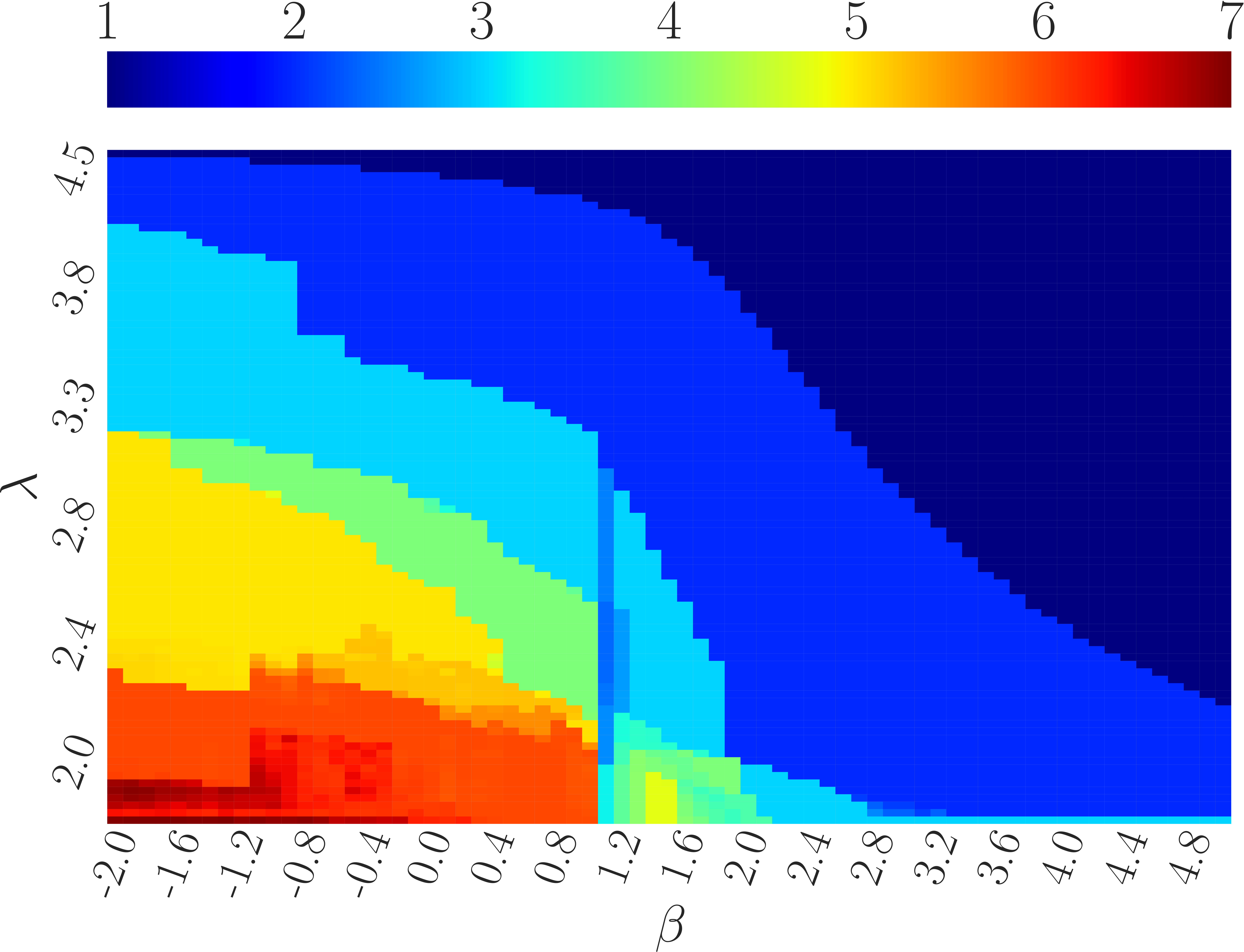}
		}
		\subfloat[NMI \label{fig:under_lim_min_max}]{
			 \includegraphics[width=48mm]{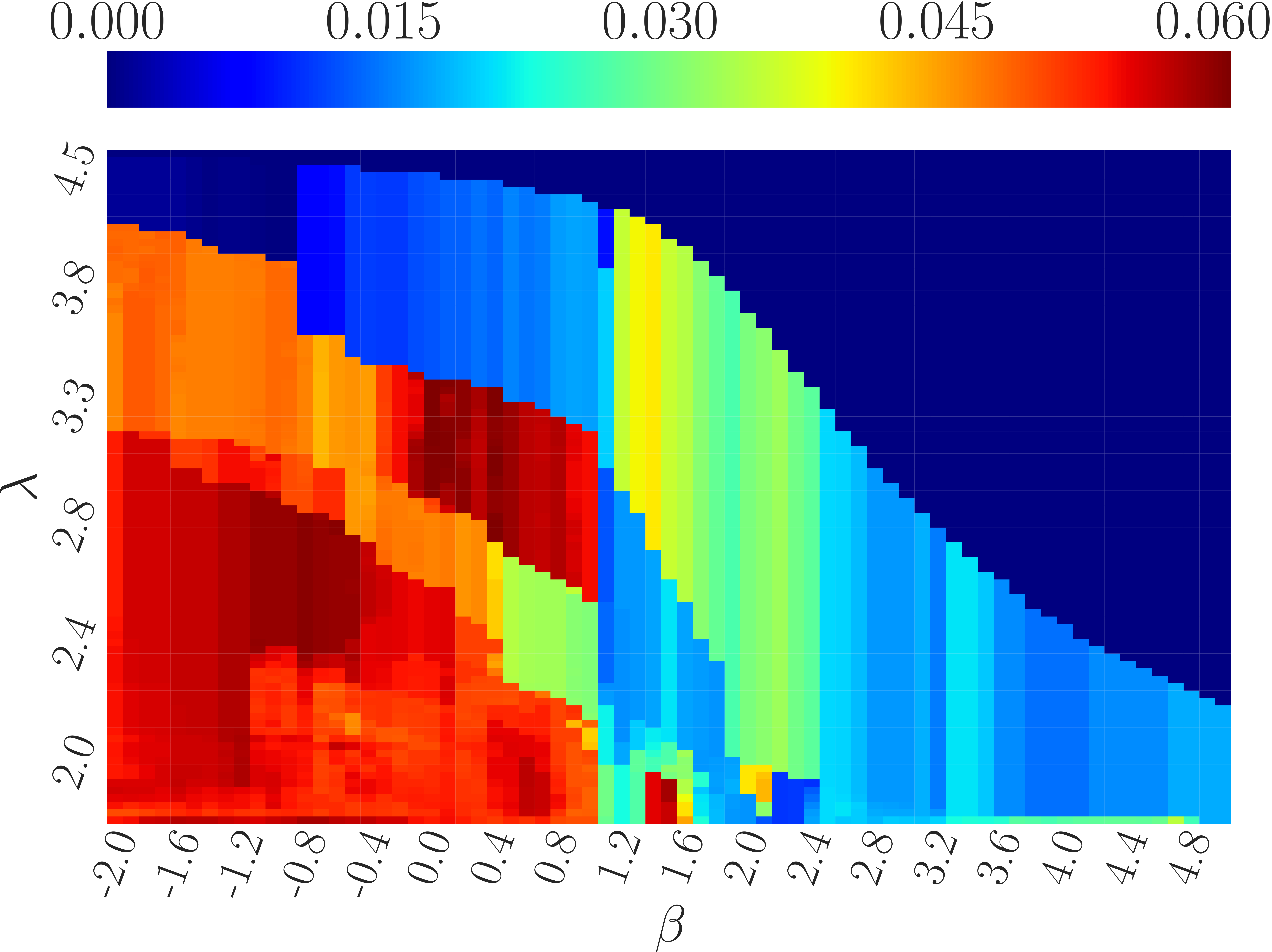}
		}
             \subfloat[Maximum distortion \label{fig:under_lim_min_max}]{
			 \includegraphics[width=47mm]{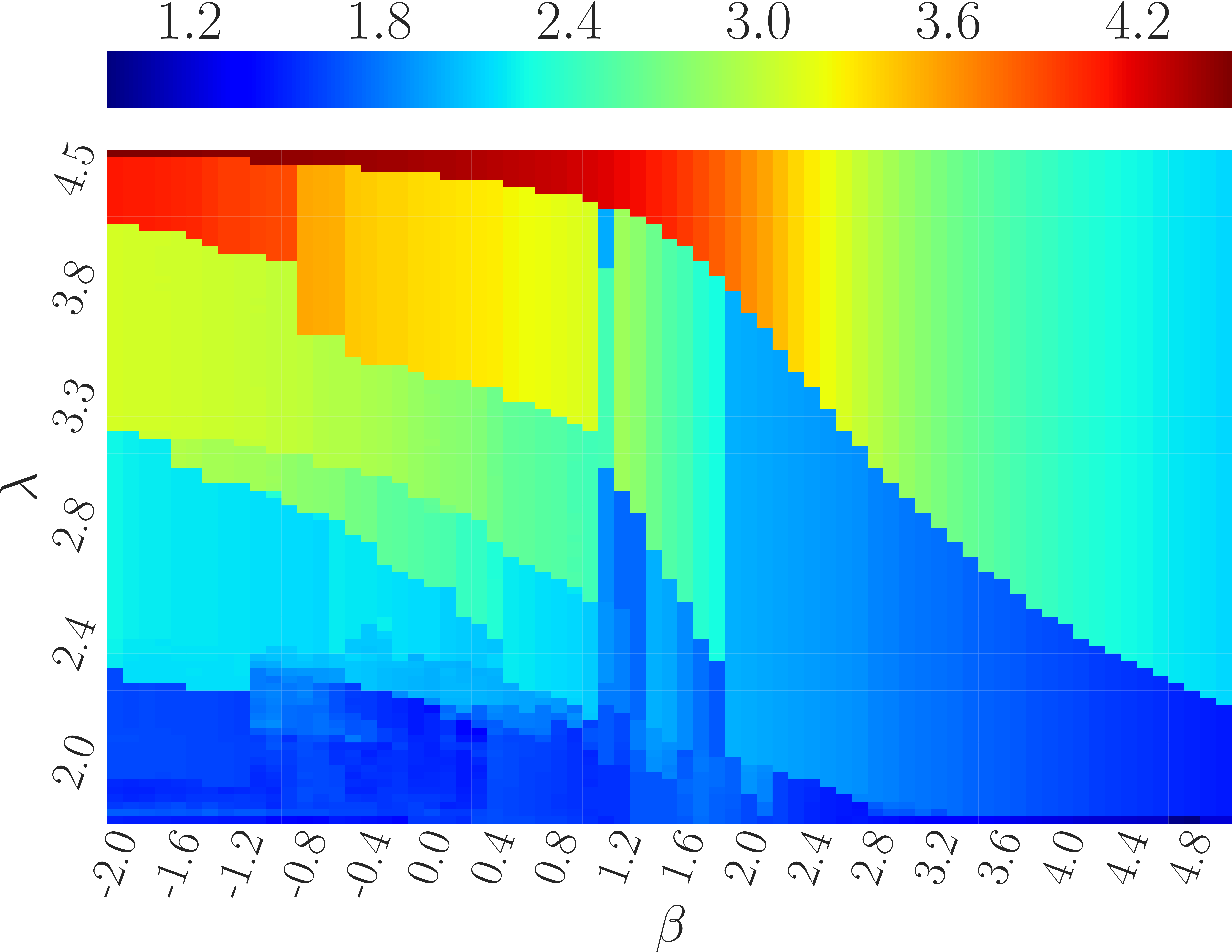}
		}
\vspace{-10pt}
		\caption{Pima, log-sum-exp \label{fig:pima_exp} }
\vspace{-10pt}
\end{figure*}

\begin{figure*}[htbp]
\centering
		\subfloat[Number of clusters \label{fig:under_lim_min_ave}]{
 			 \includegraphics[width=47mm]{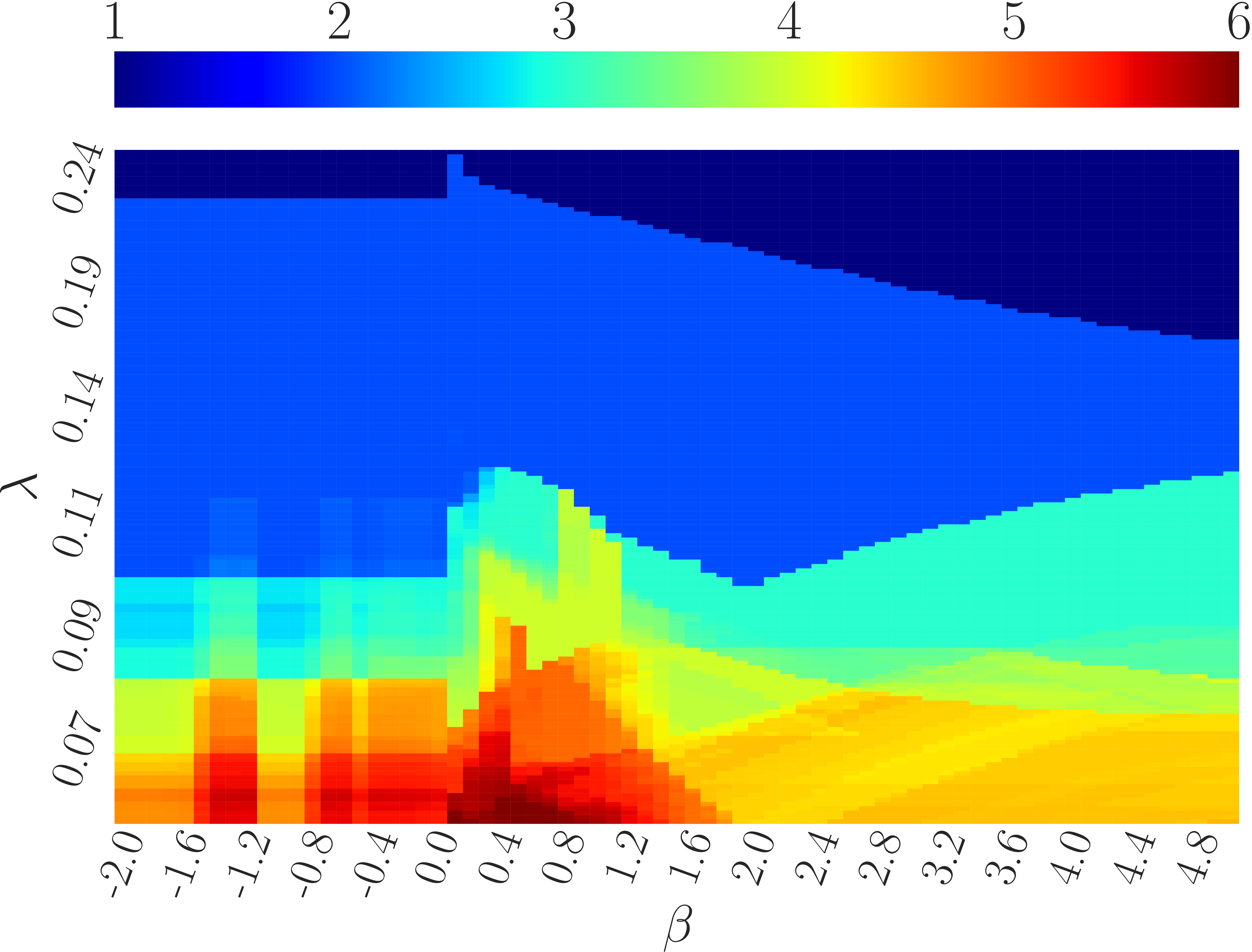}
		}
		\subfloat[NMI \label{fig:under_lim_min_max}]{
			 \includegraphics[width=47mm]{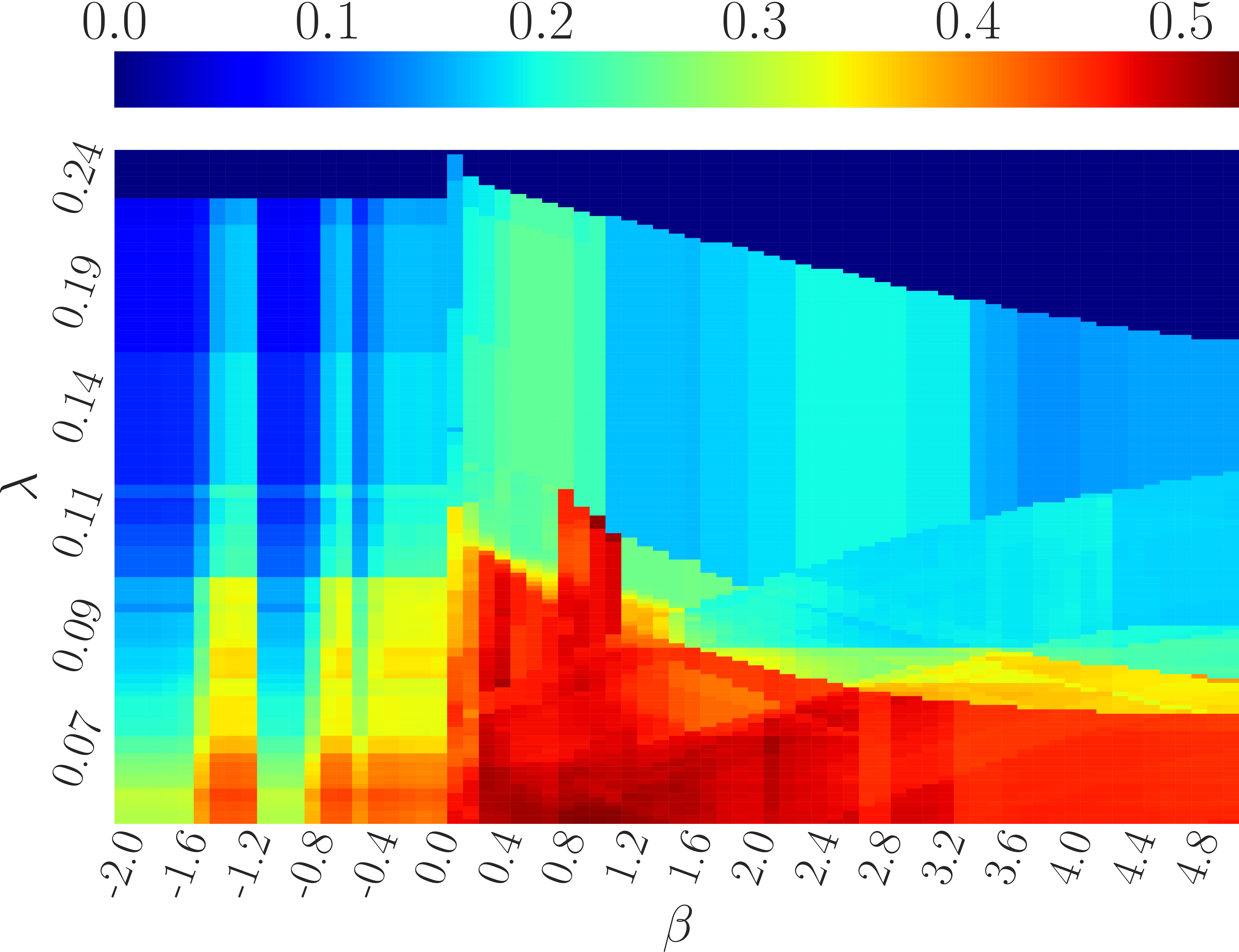}
		}
             \subfloat[Maximum distortion \label{fig:under_lim_min_max}]{
			 \includegraphics[width=47mm]{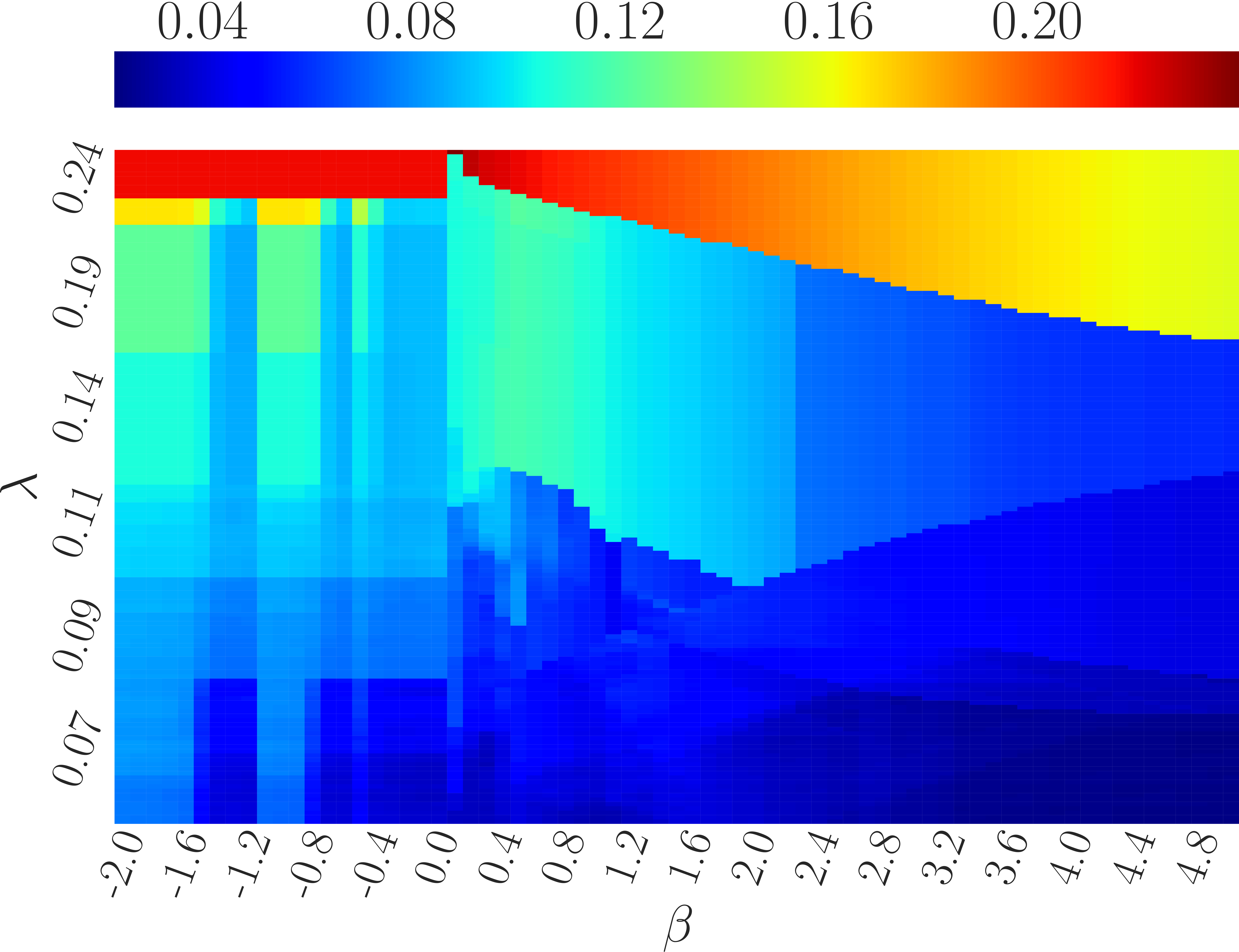}
		}
\vspace{-10pt}
		\caption{Seeds, pow mean \label{fig:seeds_pow} }
\vspace{-10pt}

\centering
		\subfloat[Number of clusters \label{fig:under_lim_min_ave}]{
 			 \includegraphics[width=47mm]{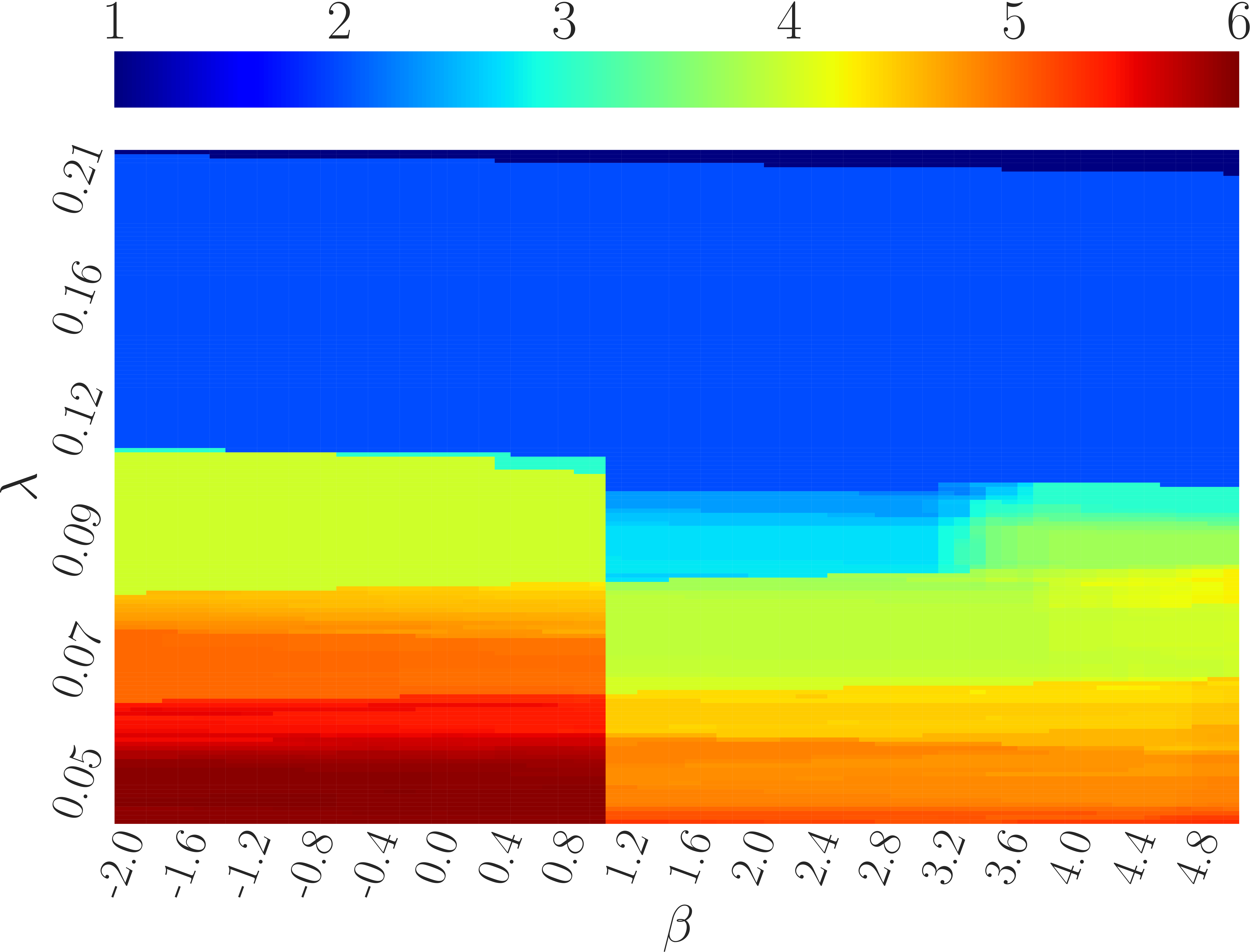}
		}
		\subfloat[NMI \label{fig:under_lim_min_max}]{
			 \includegraphics[width=47mm]{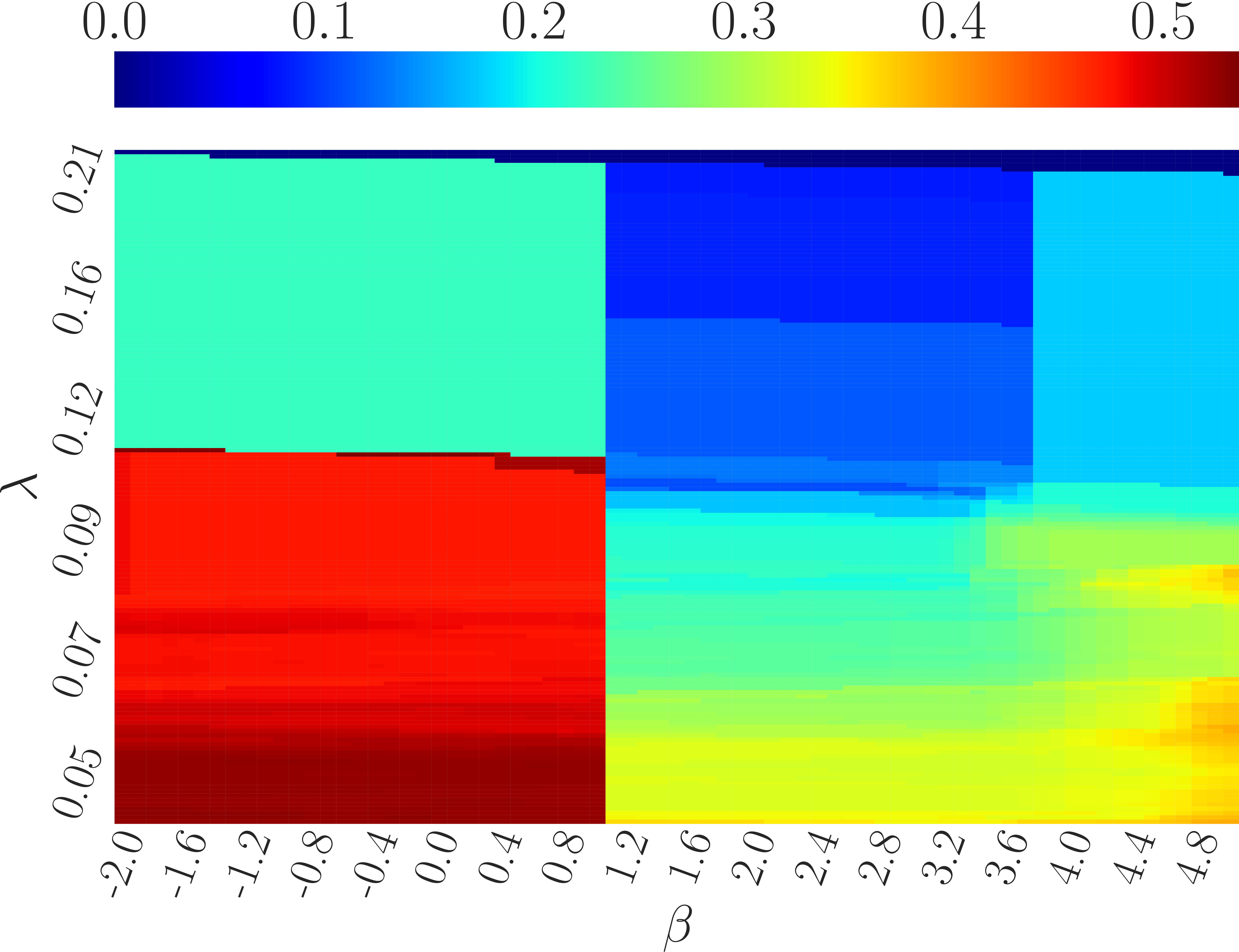}
		}
             \subfloat[Maximum distortion \label{fig:under_lim_min_max}]{
			 \includegraphics[width=47mm]{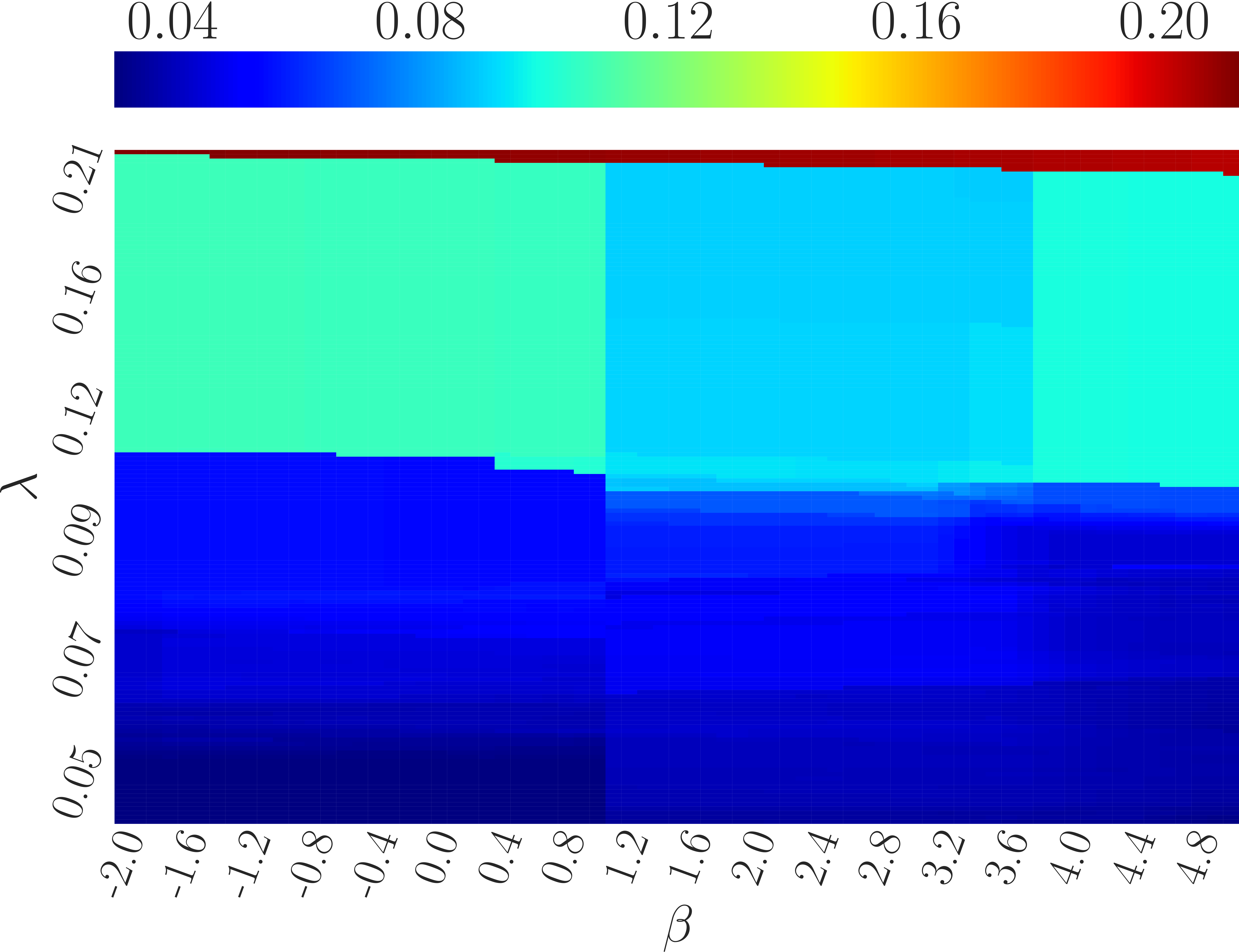}
		}
\vspace{-10pt}
		\caption{Seeds, log-sum-exp \label{fig:seeds_exp} }
\vspace{-10pt}
\end{figure*}

\begin{figure*}[htbp]
\centering
		\subfloat[Number of clusters \label{fig:under_lim_min_ave}]{
 			 \includegraphics[width=47mm]{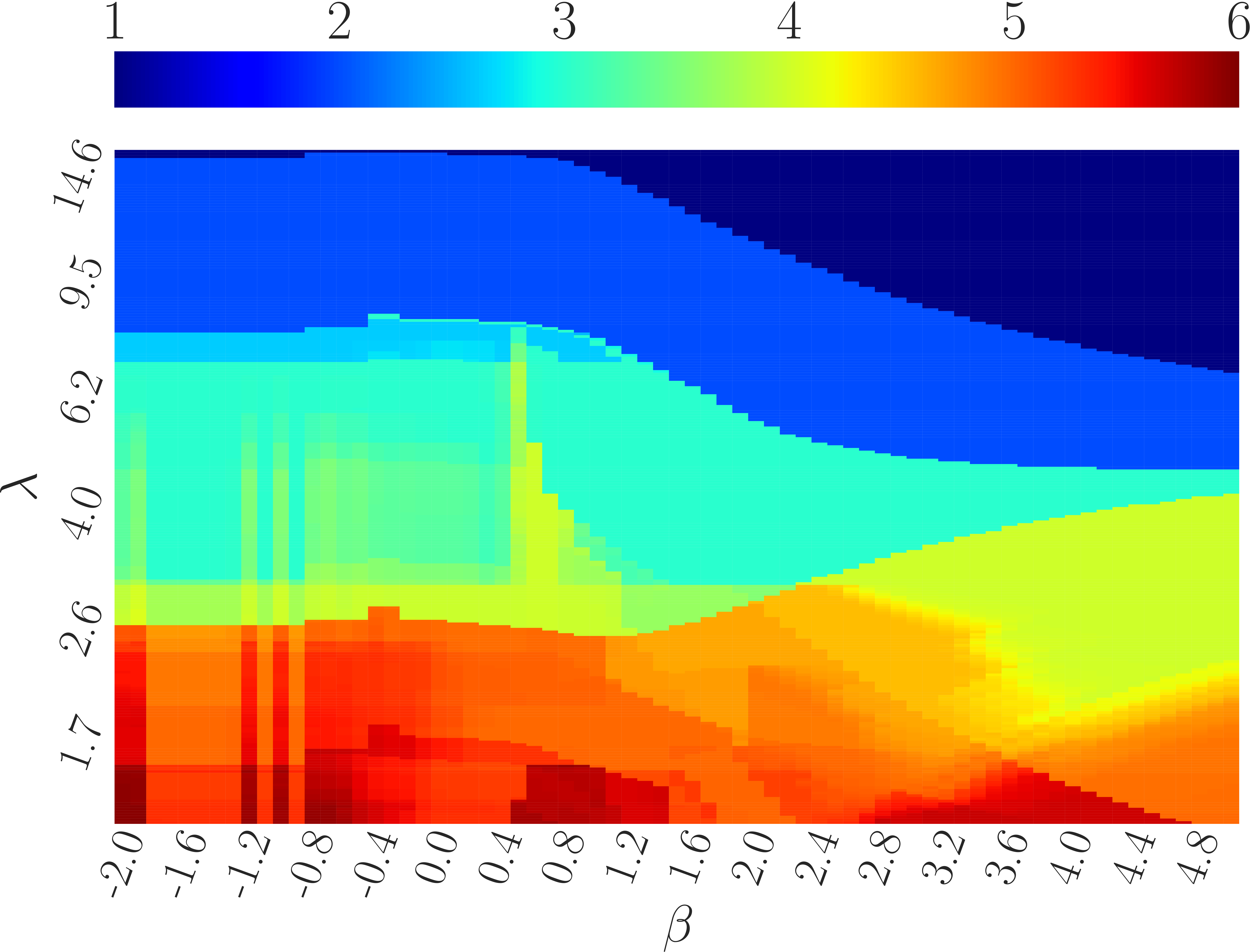}
		}
		\subfloat[NMI \label{fig:under_lim_min_max}]{
			 \includegraphics[width=48mm]{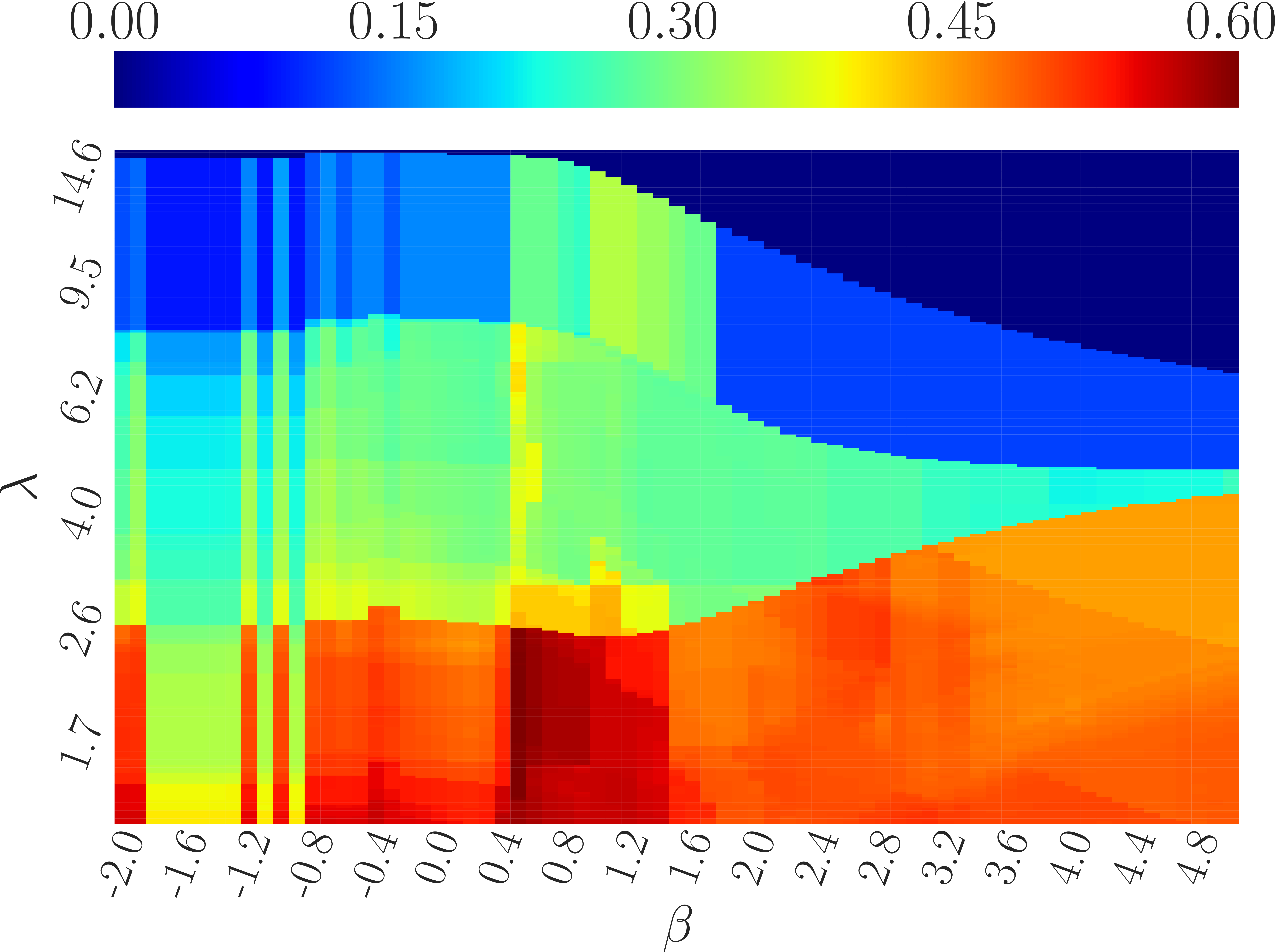}
		}
             \subfloat[Maximum distortion \label{fig:under_lim_min_max}]{
			 \includegraphics[width=47mm]{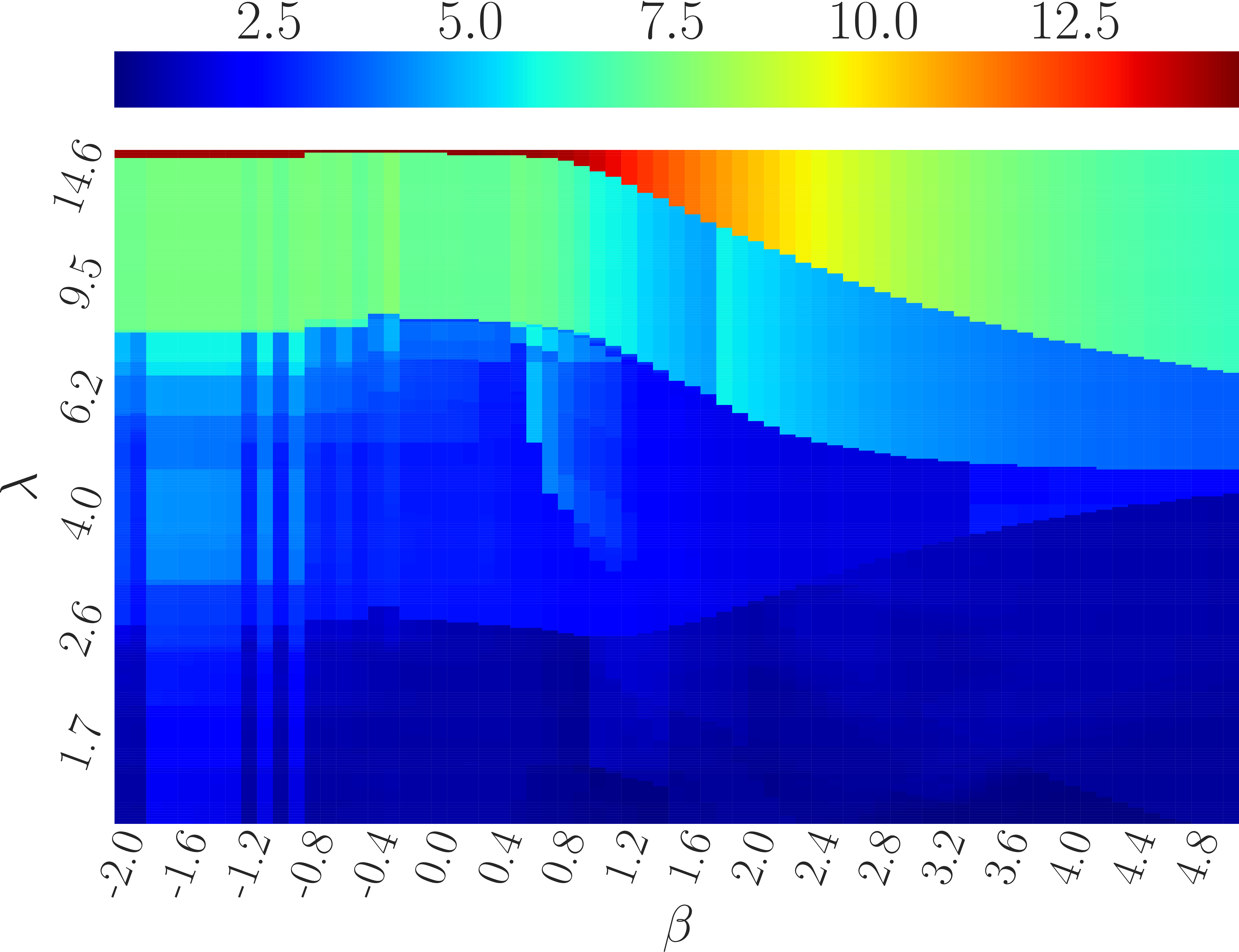}
		}
\vspace{-10pt}
		\caption{Thyroid, pow mean \label{fig:thyroid_pow} }
\vspace{-10pt}

\centering
		\subfloat[Number of clusters \label{fig:under_lim_min_ave}]{
 			 \includegraphics[width=47mm]{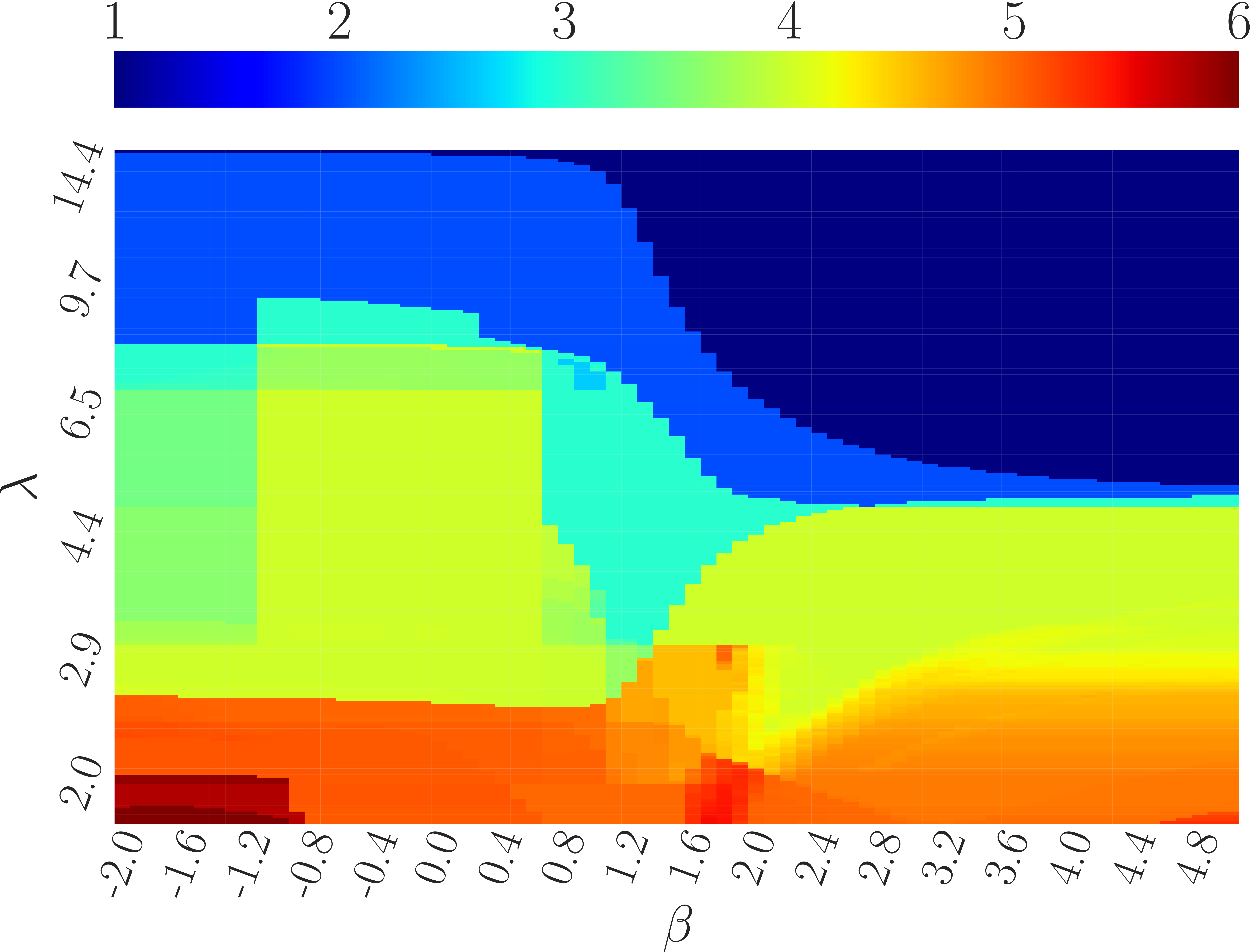}
		}
		\subfloat[NMI \label{fig:under_lim_min_max}]{
			 \includegraphics[width=48mm]{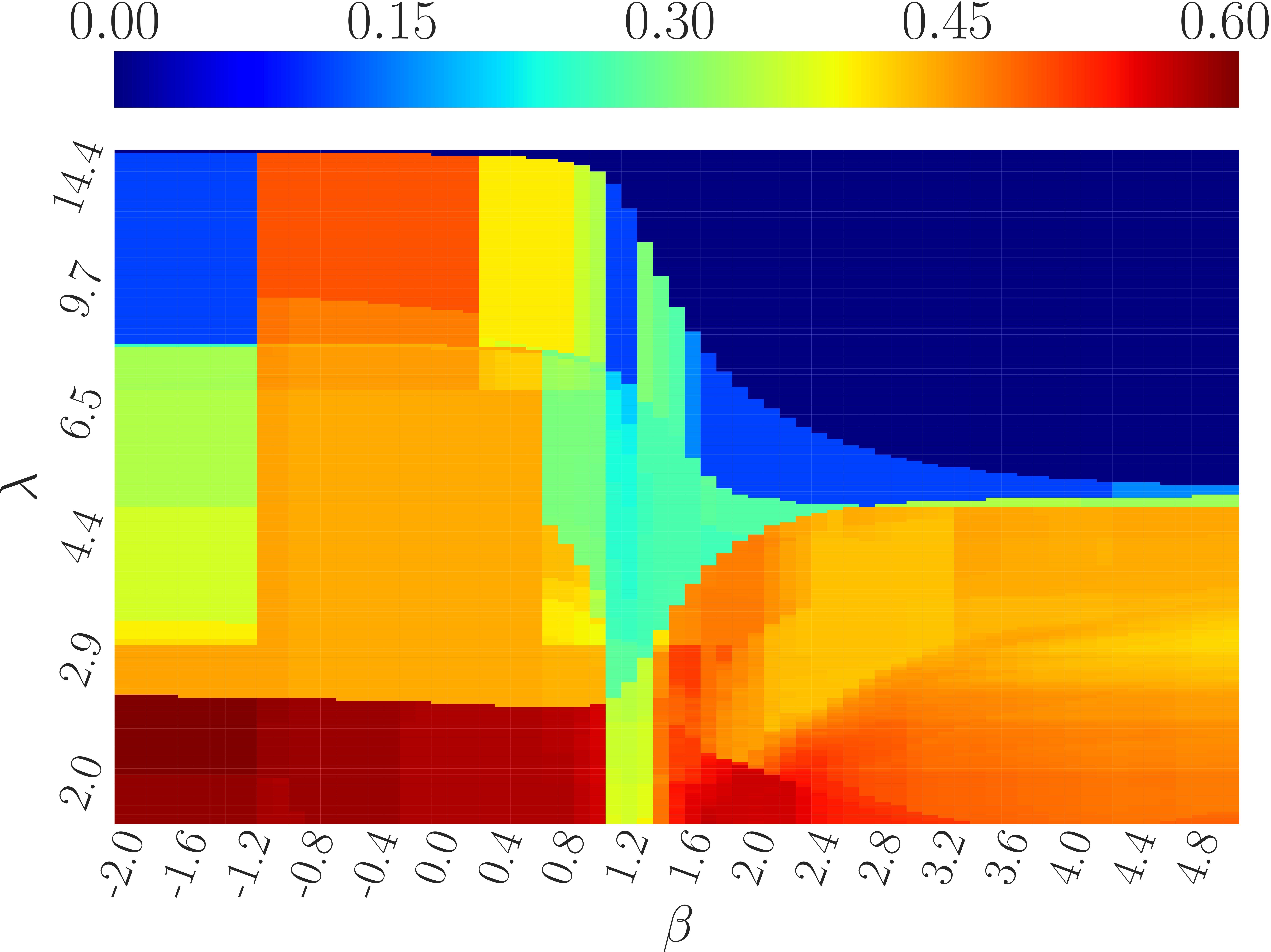}
		}
             \subfloat[Maximum distortion \label{fig:under_lim_min_max}]{
			 \includegraphics[width=47mm]{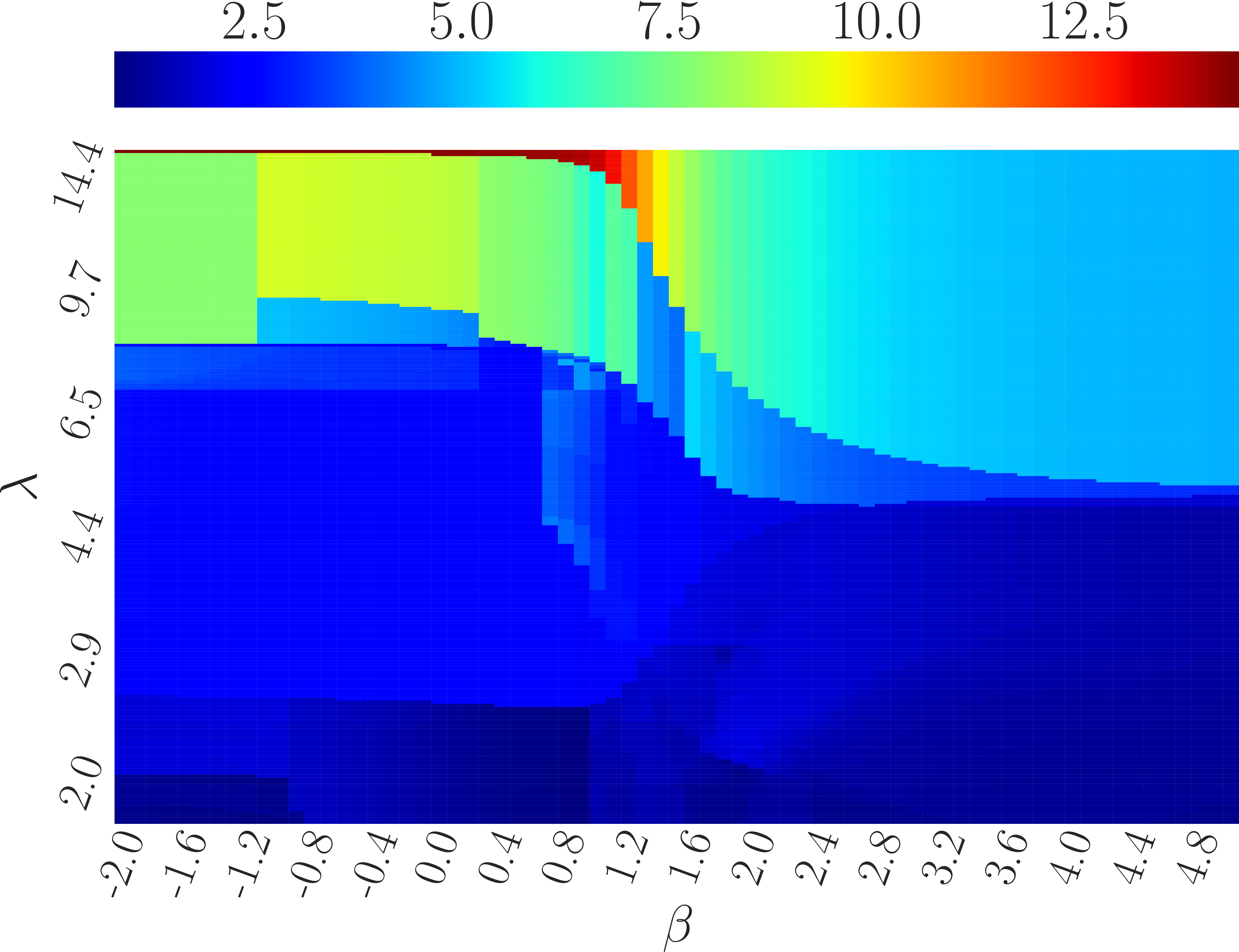}
		}
\vspace{-10pt}
		\caption{Thyroid, log-sum-exp \label{fig:thyroid_exp} }
\vspace{-10pt}
\end{figure*}

\begin{figure*}[htbp]
\centering
		\subfloat[Number of clusters \label{fig:under_lim_min_ave}]{
 			 \includegraphics[width=47mm]{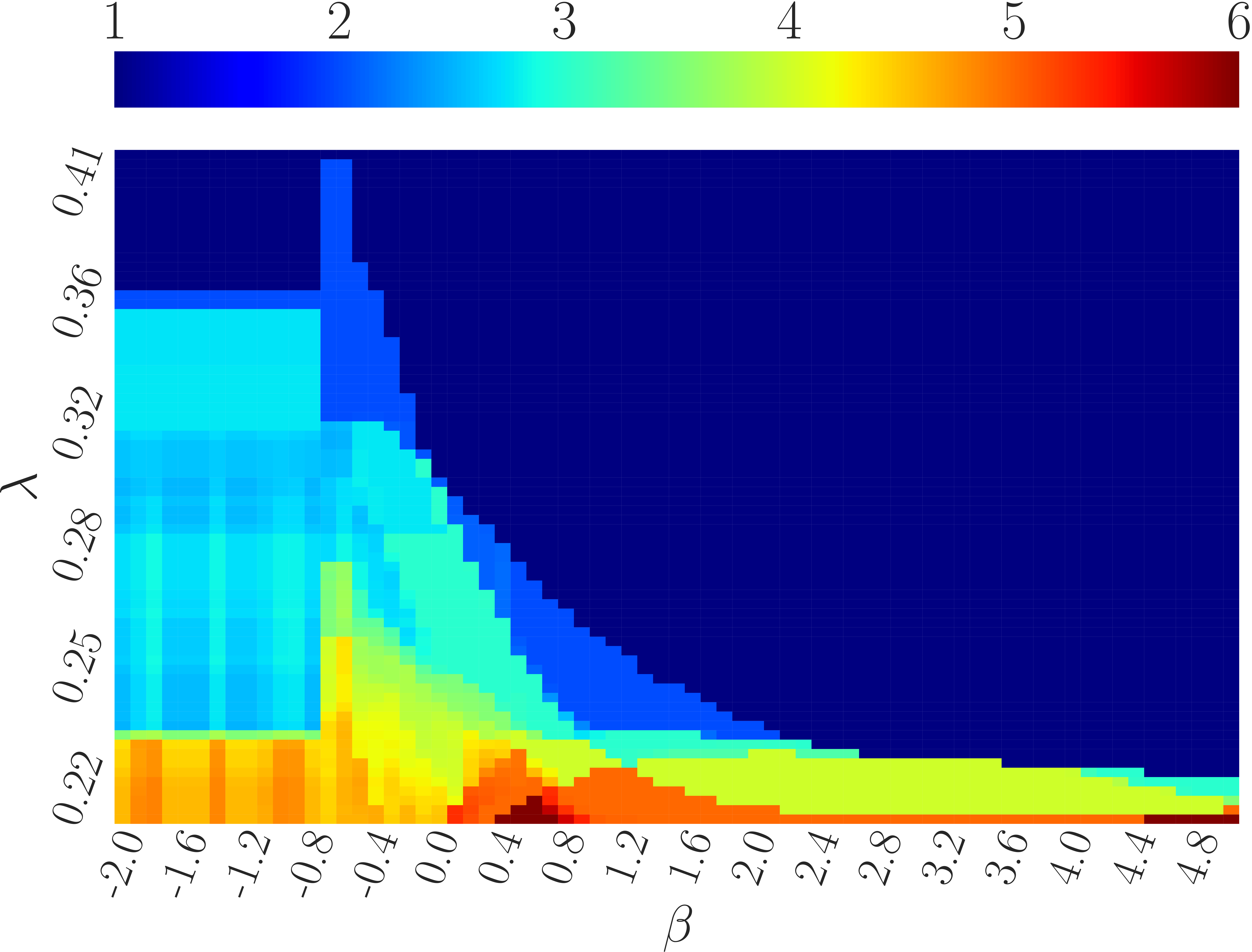}
		}
		\subfloat[NMI \label{fig:under_lim_min_max}]{
			 \includegraphics[width=47mm]{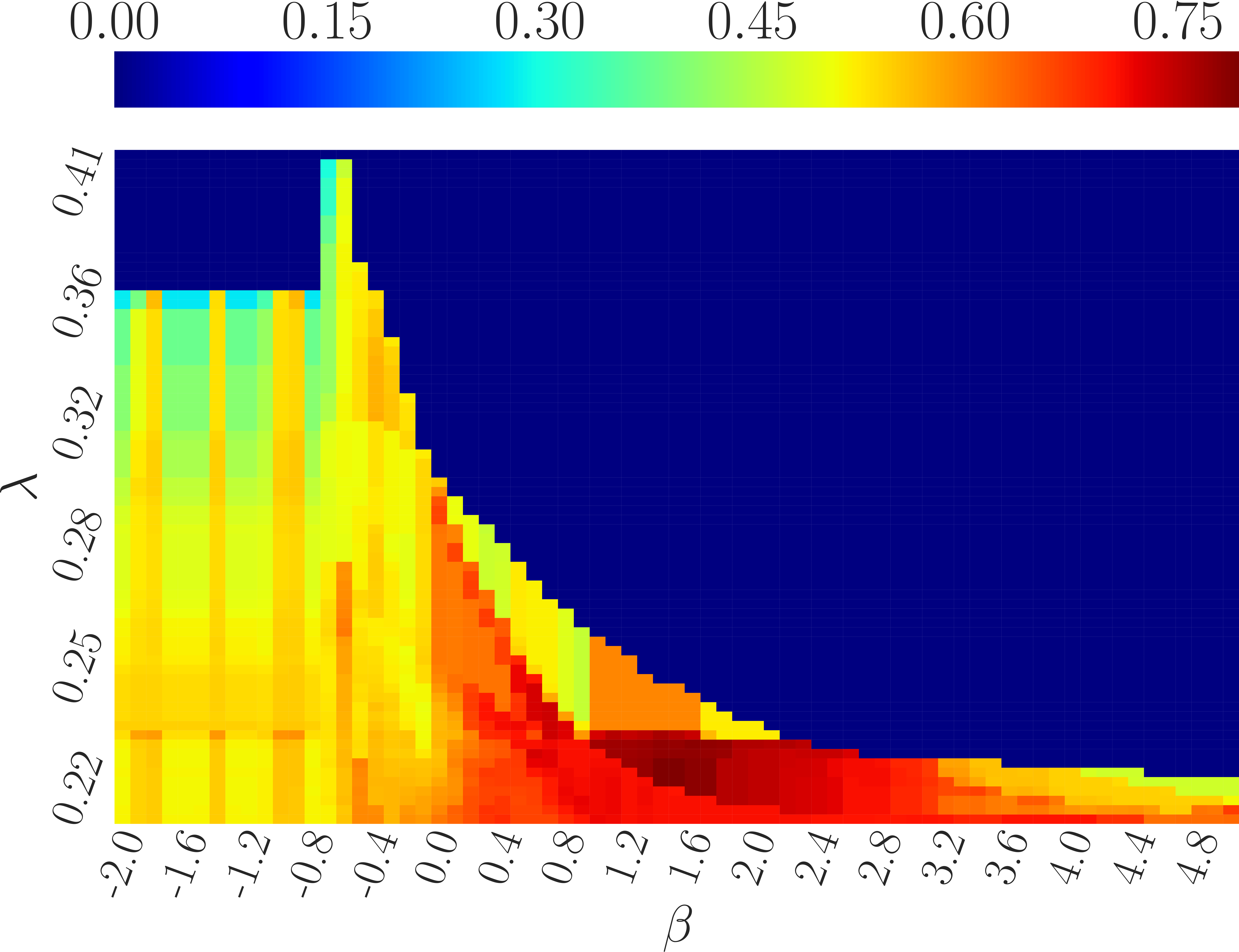}
		}
             \subfloat[Maximum distortion \label{fig:under_lim_min_max}]{
			 \includegraphics[width=47mm]{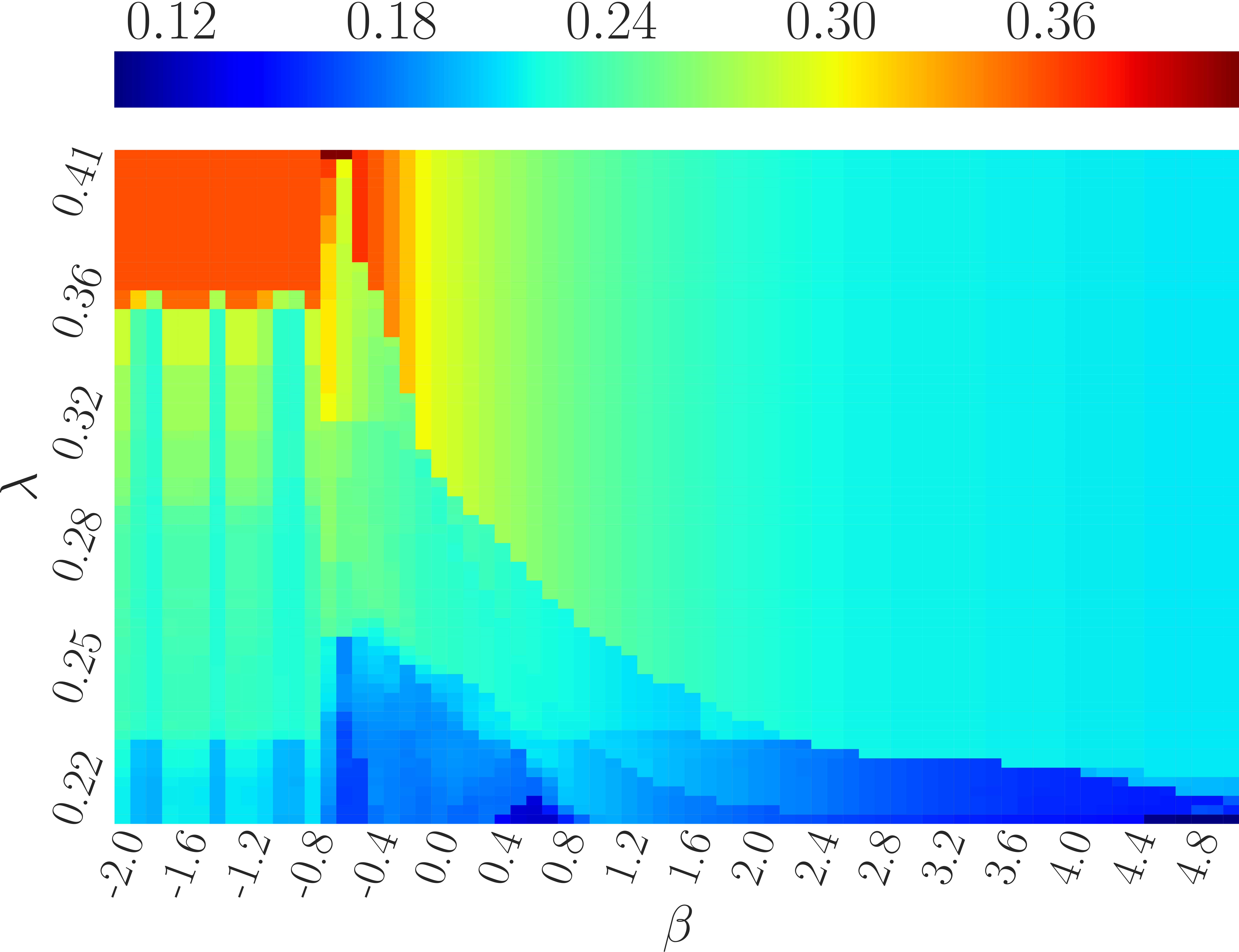}
		}
\vspace{-10pt}
		\caption{Wine, pow mean \label{fig:wine_pow} }
\vspace{-10pt}

\centering
		\subfloat[Number of clusters \label{fig:under_lim_min_ave}]{
 			 \includegraphics[width=47mm]{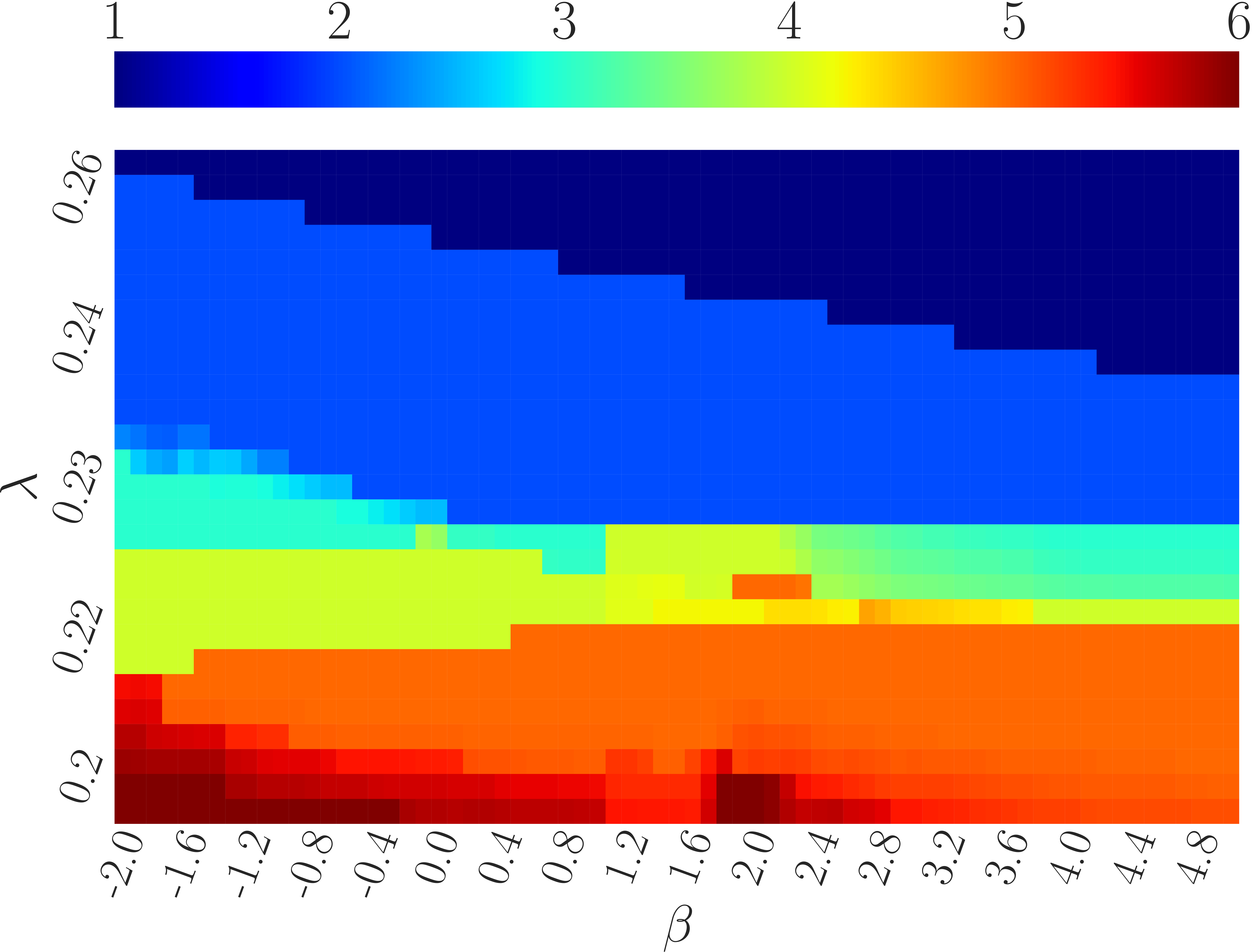}
		}
		\subfloat[NMI \label{fig:under_lim_min_max}]{
			 \includegraphics[width=47mm]{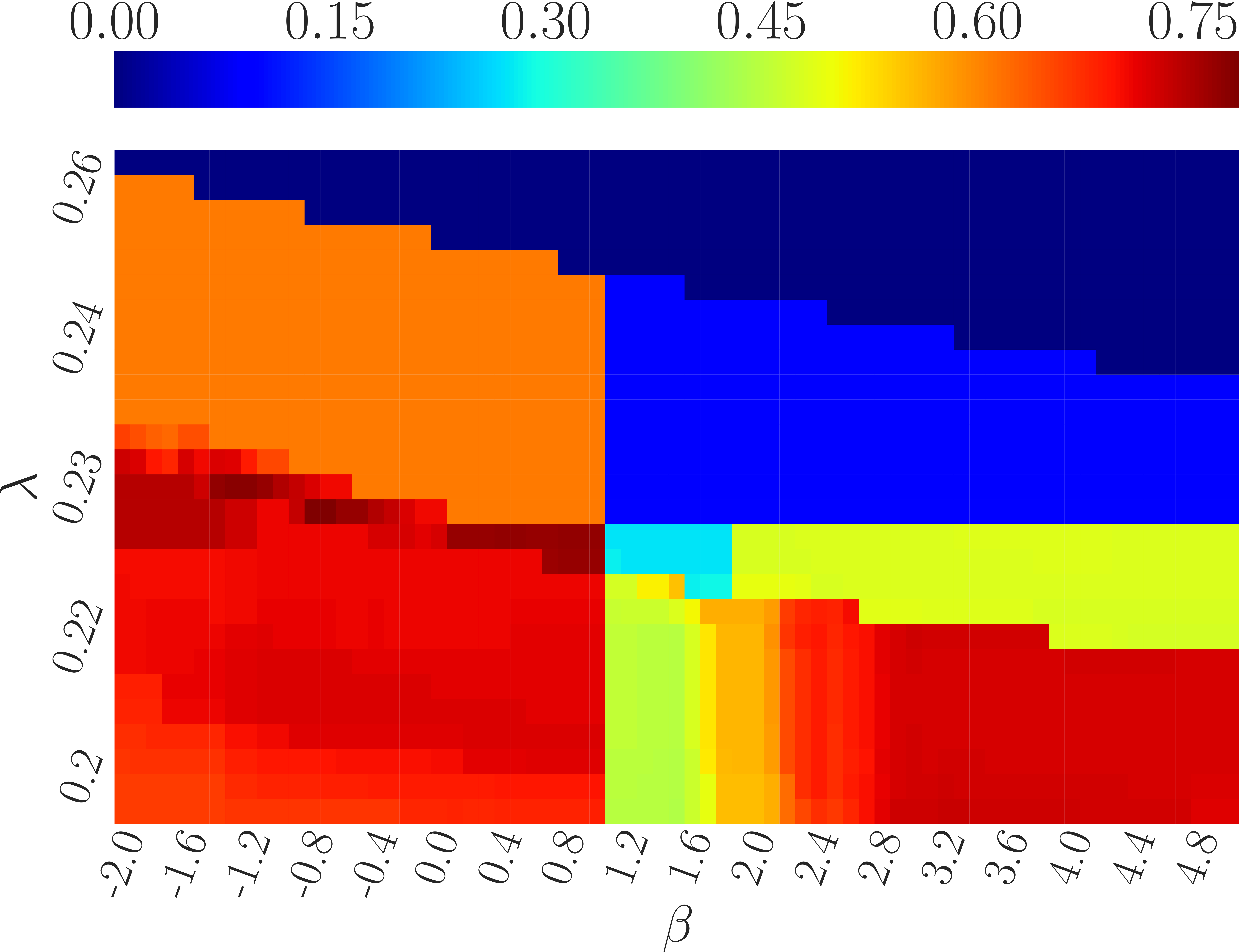}
		}
             \subfloat[Maximum distortion \label{fig:under_lim_min_max}]{
			 \includegraphics[width=47mm]{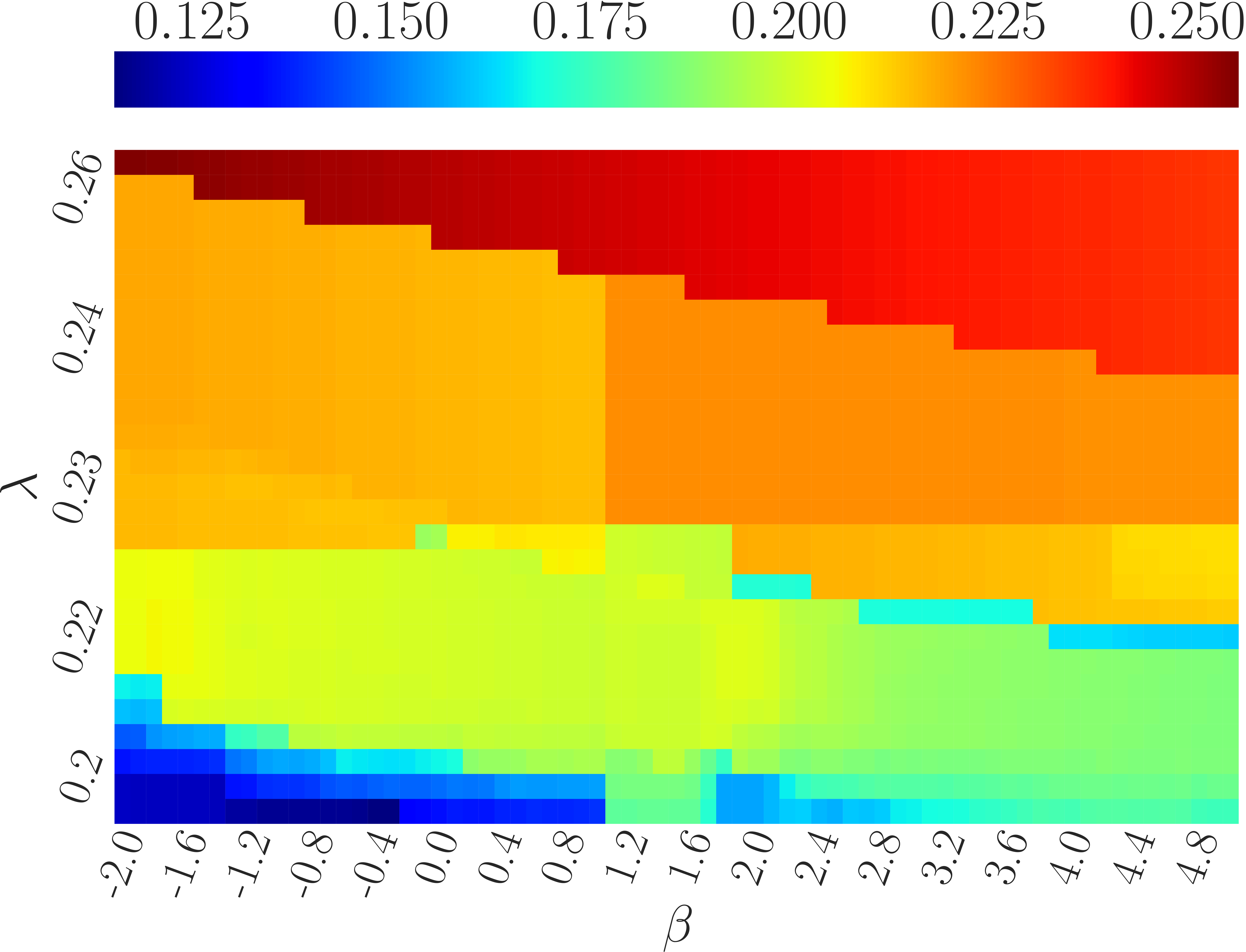}
		}
\vspace{-10pt}
		\caption{Wine, log-sum-exp \label{fig:wine_exp} }
\vspace{-10pt}
\end{figure*}

\begin{figure*}[htbp]
\centering
		\subfloat[Number of clusters \label{fig:under_lim_min_ave}]{
 			 \includegraphics[width=47mm]{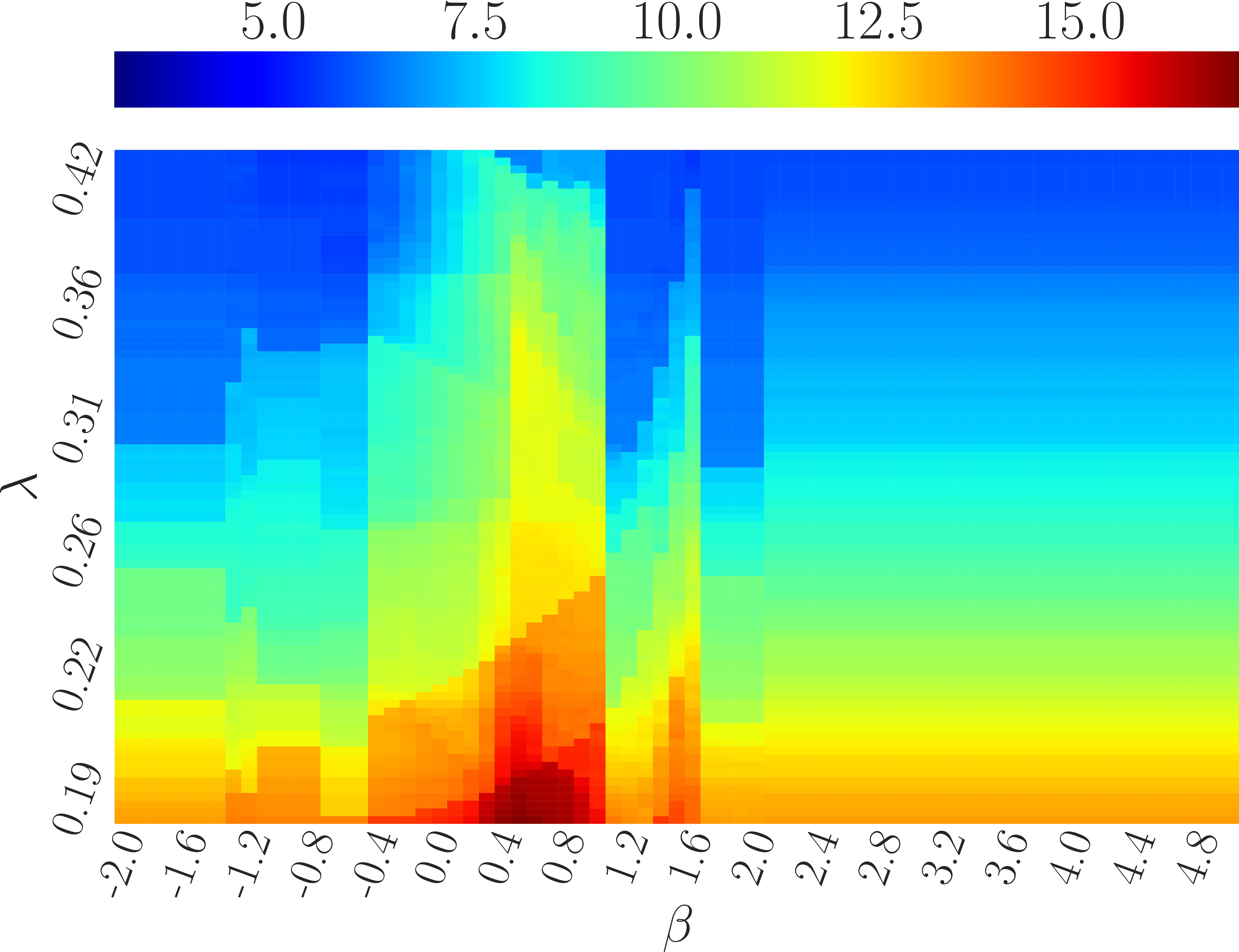}
		}
		\subfloat[NMI \label{fig:under_lim_min_max}]{
			 \includegraphics[width=47mm]{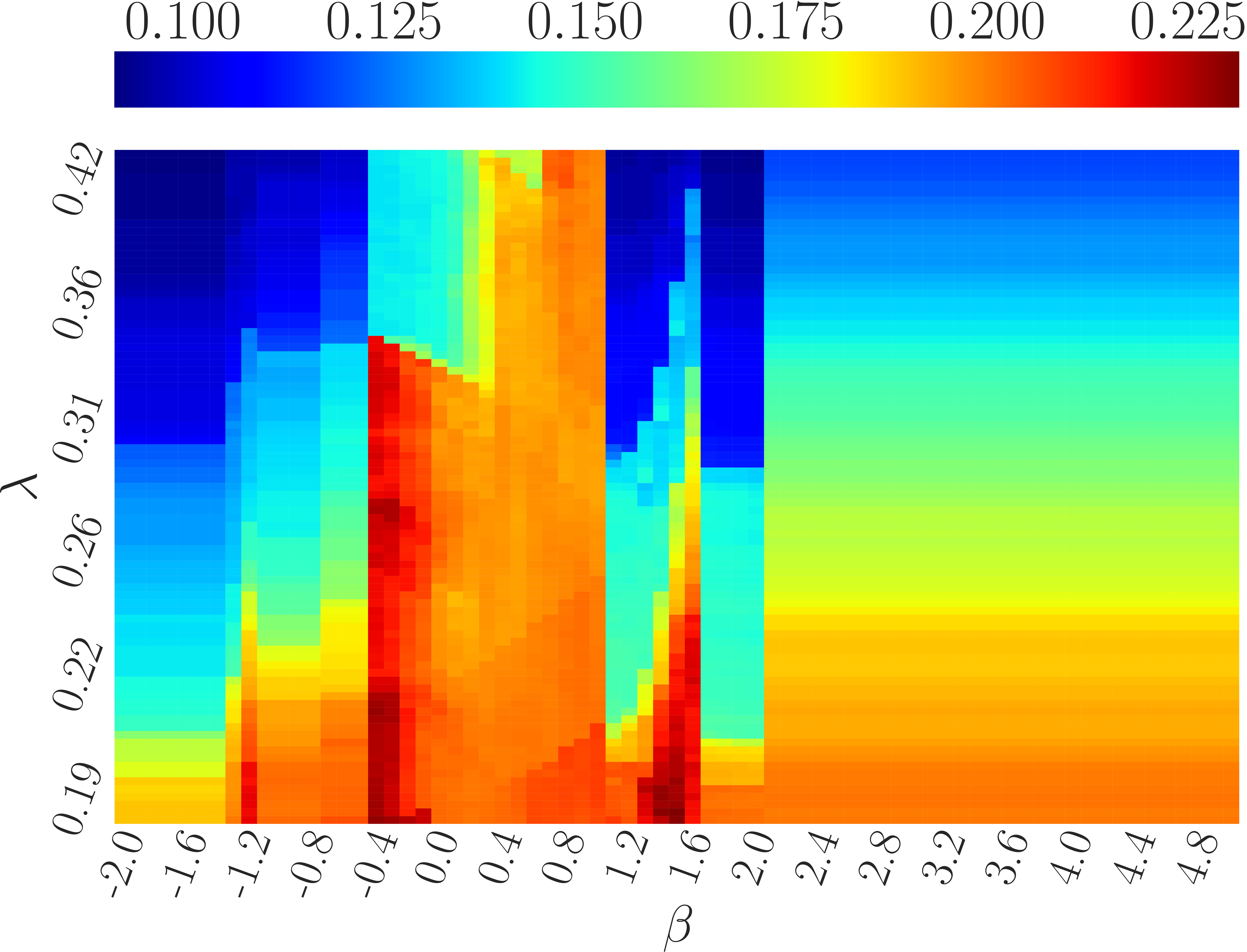}
		}
             \subfloat[Maximum distortion \label{fig:under_lim_min_max}]{
			 \includegraphics[width=47mm]{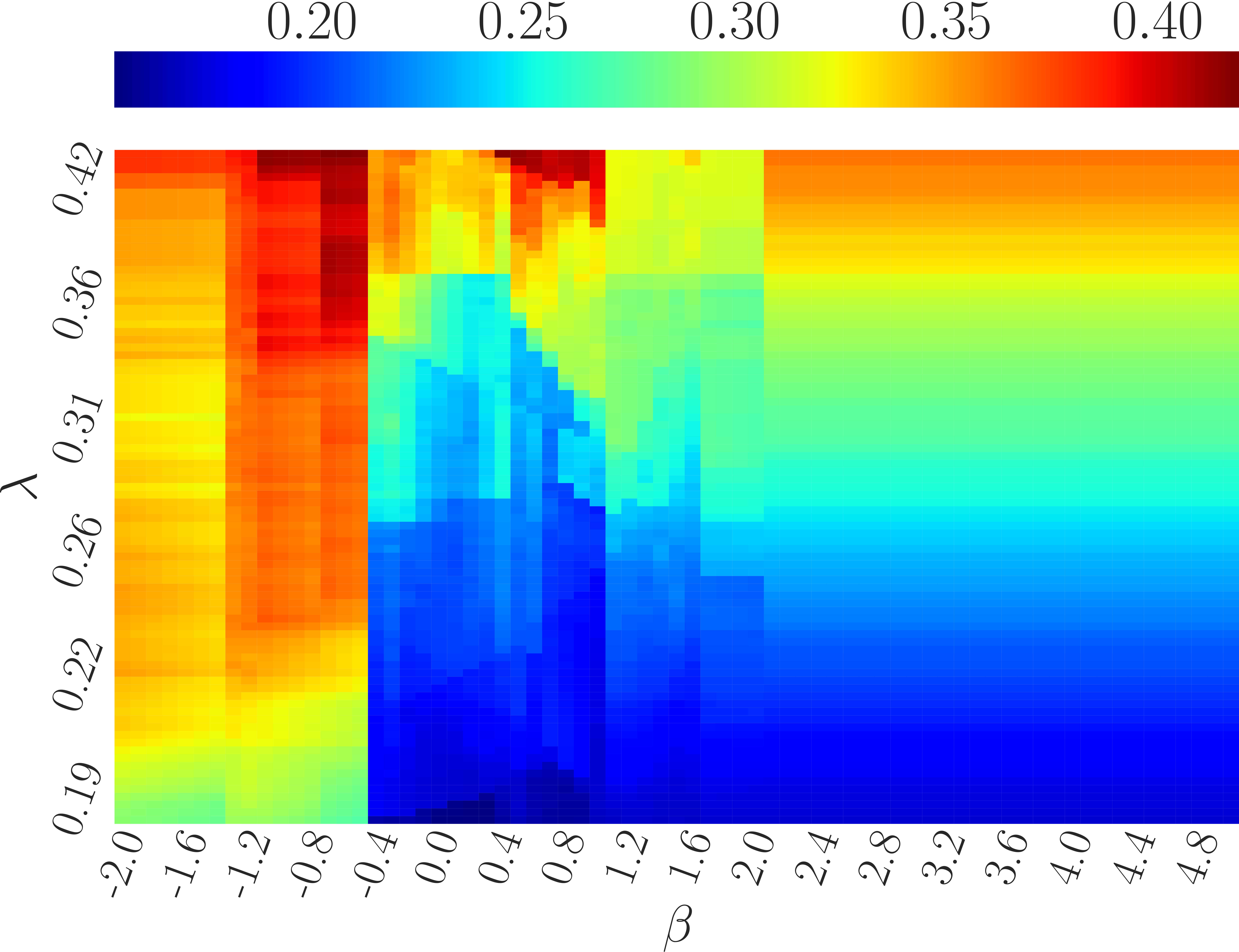}
		}
\vspace{-10pt}
		\caption{Yeast, pow mean \label{fig:yeast_pow} }
\vspace{-10pt}

\centering
		\subfloat[Number of clusters \label{fig:under_lim_min_ave}]{
 			 \includegraphics[width=47mm]{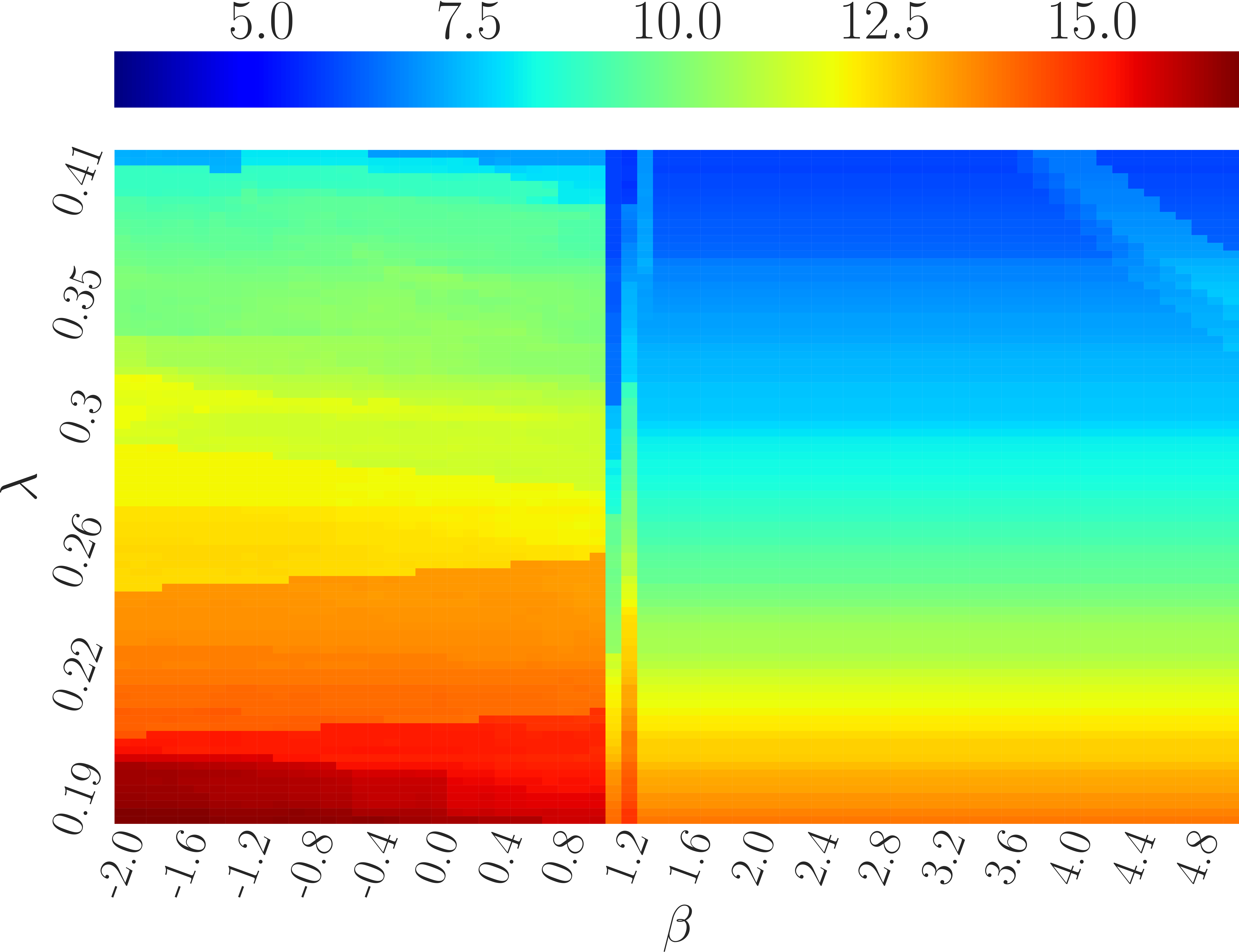}
		}
		\subfloat[NMI \label{fig:under_lim_min_max}]{ 
			 \includegraphics[width=48mm]{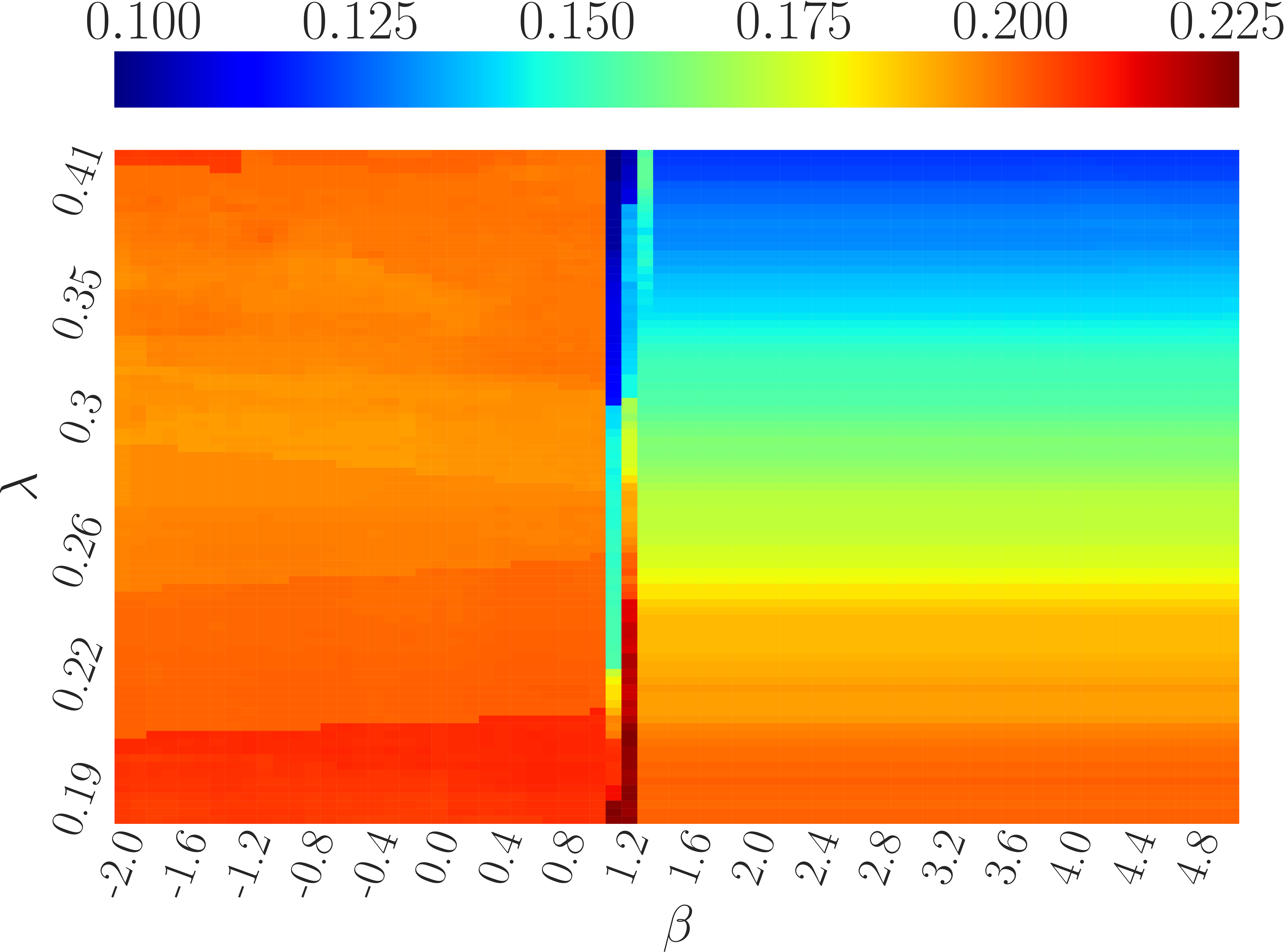}
		}
             \subfloat[Maximum distortion \label{fig:under_lim_min_max}]{
			 \includegraphics[width=47mm]{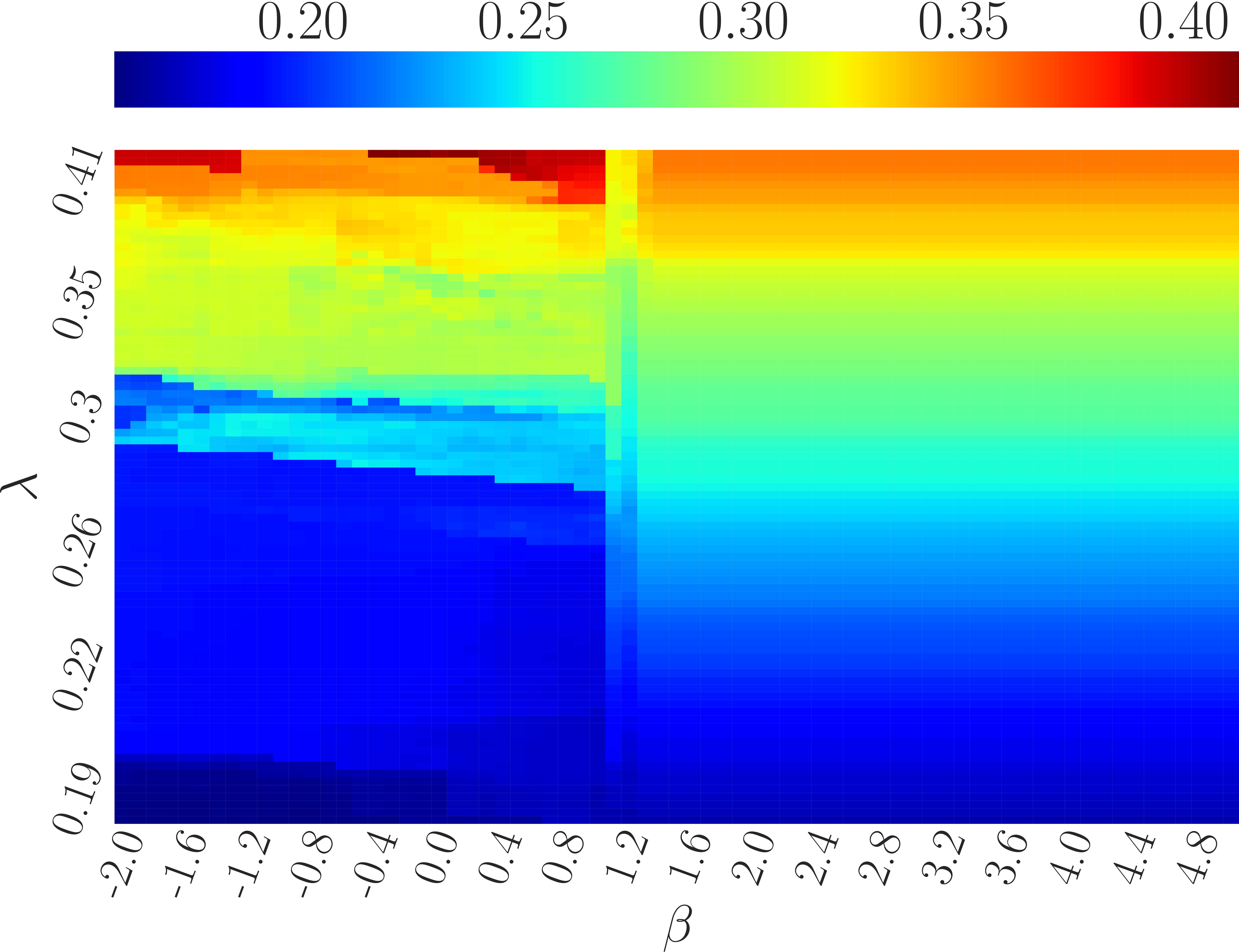}
		}
\vspace{-10pt}
		\caption{Yeast, log-sum-exp \label{fig:yeast_exp} }
\vspace{-10pt}
\end{figure*}

\subsubsection{Result}
Figures \ref{fig:bcw_pow} through \ref{fig:yeast_exp} show the number of clusters, NMI, and the maximum distortion by heat maps for the 11 datasets.
We can see that the number of clusters, NMI, and maximum distortion are correlated.
Depending on the dataset, NMI tends to be improved when the value of $\beta$ is lowered below 1.
The tendency is remarkable in ``Breast Cancer Wisconsin'' (Figure \ref{fig:bcw_pow}(b), Figure \ref{fig:bcw_exp}(b)) and ``HTRU2'' (Figure \ref{fig:htru2_pow}(b), Figure \ref{fig:htru2_exp}(b)), where NMI increases monotonically as the value of $\beta$ is lowered.
In other words, the outliers in the sense discussed in Section \ref{sec:outlier} exist implicitly in the datasets in which NMI was improved by lowering $\beta$.
However, NMI worsens if $\beta$ is lowered too much.
Since the number of clusters and the maximum distortion are in a trade-off relationship, we can see the maximum distortion tends to decrease as $\beta$ increases when the number of clusters is fixed and the maximum distortion is compared.
Theoretically, the larger $\beta$, the closer the generalized DP-means to the maximum distortion minimization and the smaller $\beta$, the it robust against outliers.
The above result supports this fact.

\begin{figure*}[htbp]
\centering
		\subfloat[Number of clusters \label{fig:under_lim_min_ave}]{
 			 \includegraphics[width=47mm]{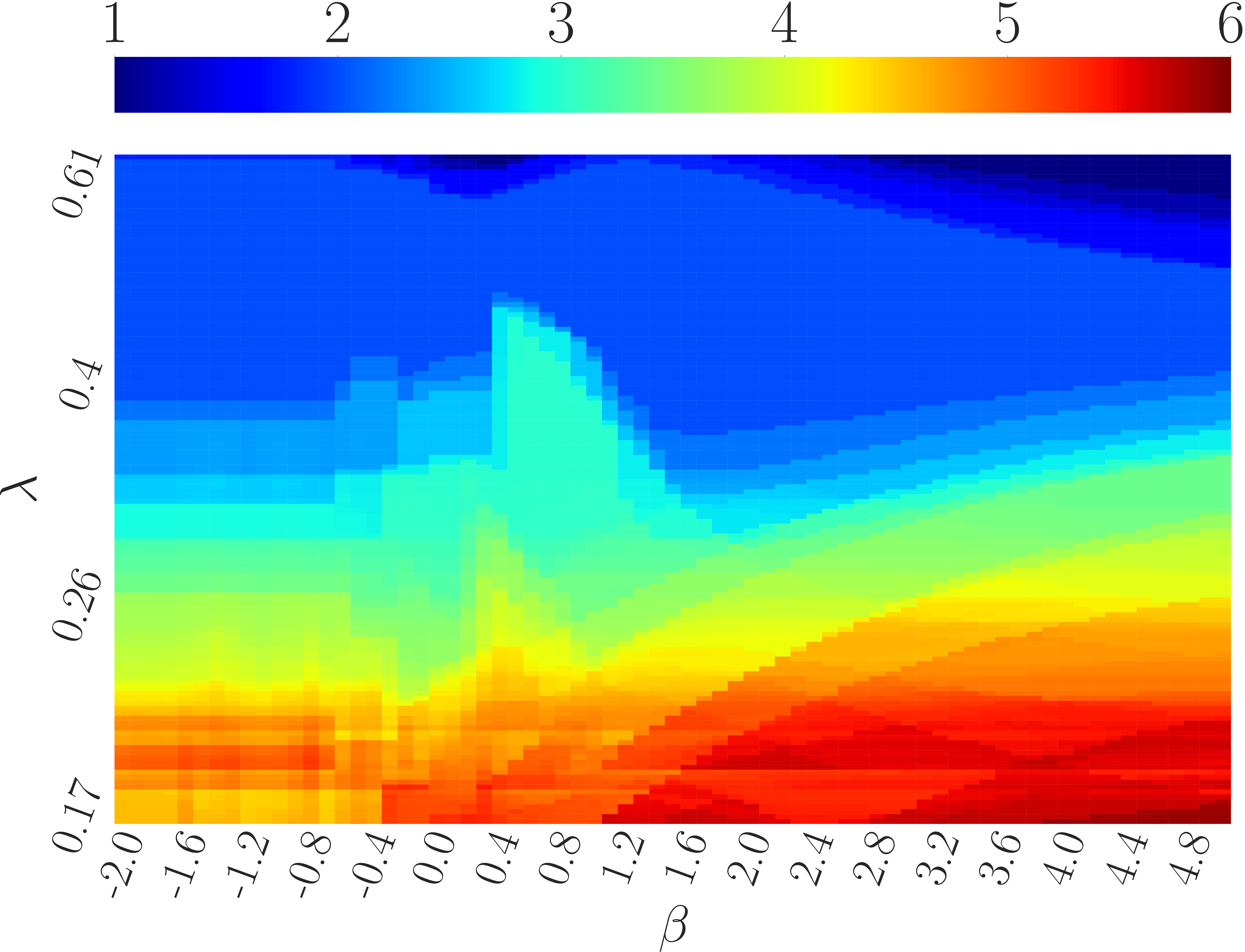}
		}
		\subfloat[NMI \label{fig:under_lim_min_max}]{
			 \includegraphics[width=47mm]{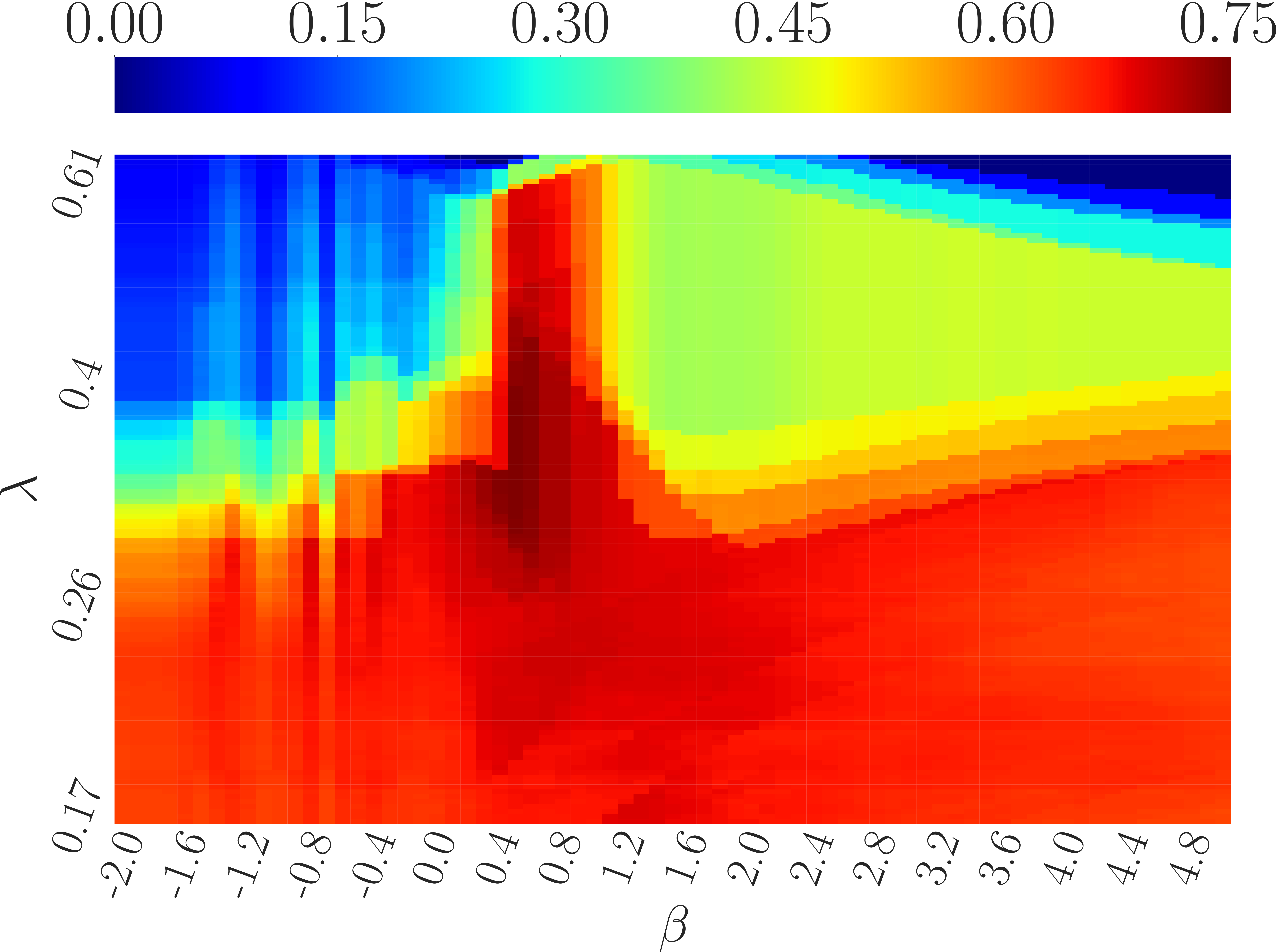}
		}
\vspace{-10pt}
		\caption{Iris contaminated with outliers, pow mean \label{fig:Iris_outlier_pow} }
\vspace{-10pt}

\centering
		\subfloat[Number of clusters \label{fig:under_lim_min_ave}]{
 			 \includegraphics[width=47mm]{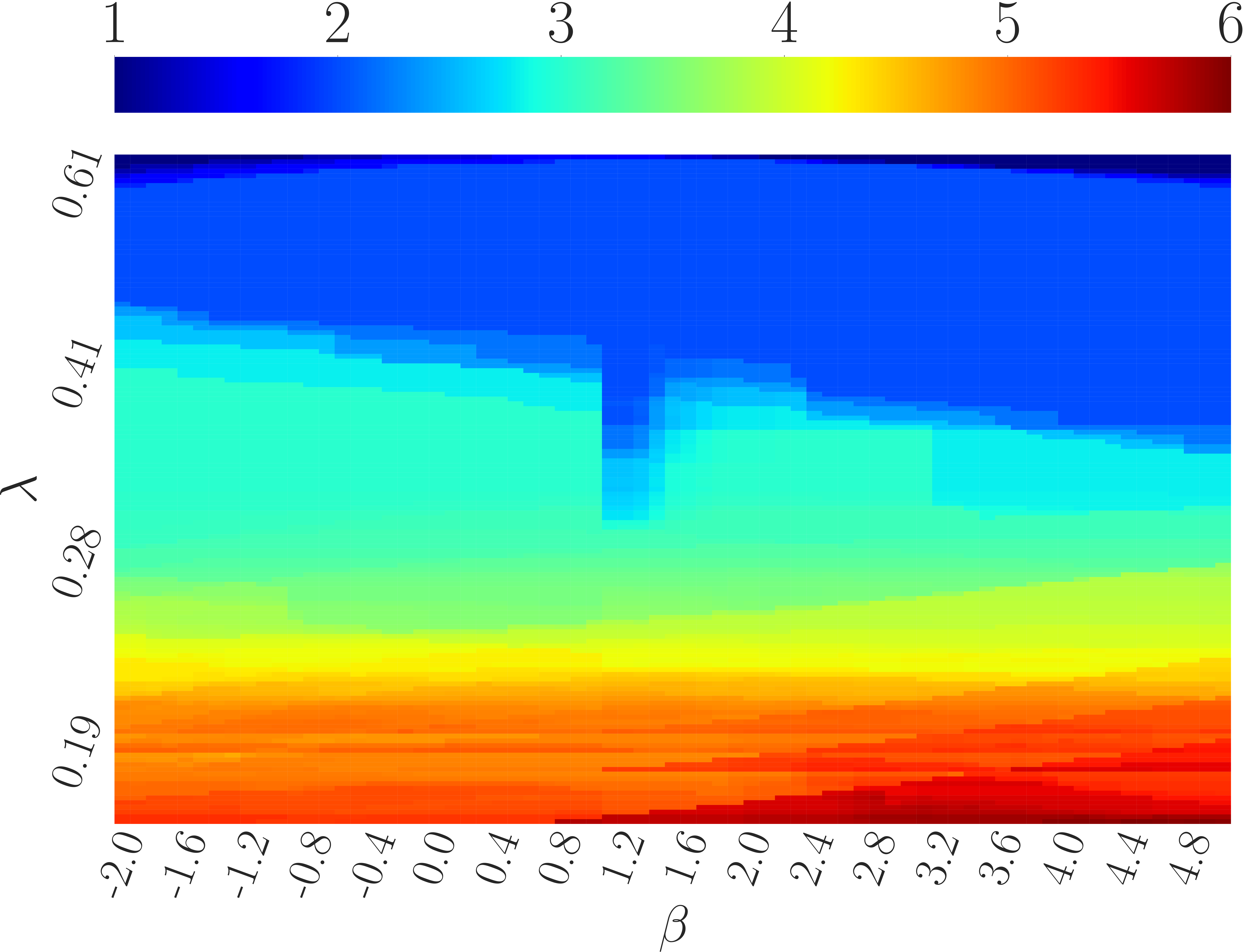}
		}
		\subfloat[NMI \label{fig:under_lim_min_max}]{ 
			 \includegraphics[width=47mm]{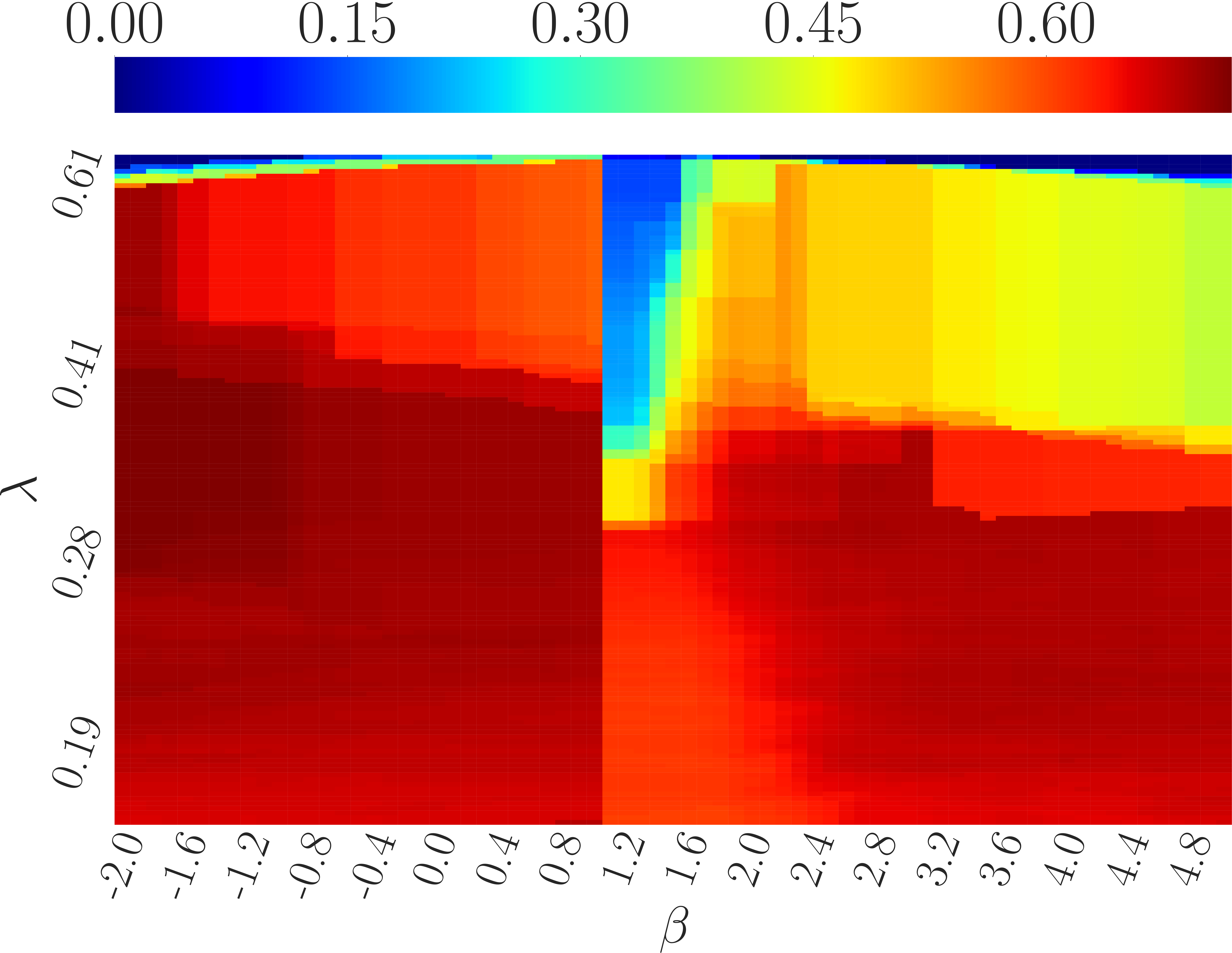}
		}
\vspace{-10pt}
		\caption{Iris contaminated with outliers, log-sum-exp \label{fig:Iris_outlier_exp} }
\vspace{-10pt}
\end{figure*}

On the other hand, the value of NMI may not change even if the value of $\beta$ is lowered than 1 as in the Iris dataset.
It means that the dataset did not include the outlier which we assumed in Section \ref{sec:outlier}.
Therefore, we explicitly added outliers to the Iris dataset and conducted an additional experiment.
The Iris dataset has 3 clusters, setosa, versicolor and virginica.
The clusters of versicolor and virginica overlap.
In such a situation, adding outliers to either cluster would adversely affect the estimation.
Hence, outliers of $30\%$ of the total original data were randomly mixed at 0.95 to 1 times the true maximum distortion from the true cluster center of the virginica cluster.
It is expected that this procedure simulates the outliers assumed in Section \ref{sec:outlier}.
Then, we applied the same preprocessing of data and experiment as described in Section \ref{sec:method}.
The result is shown in Figures \ref{fig:Iris_outlier_pow} and \ref{fig:Iris_outlier_exp}.
We can see that the value of NMI is improved when $\beta$ is lowered from 1 compared to when $\beta=1$.
This fact shows that the generalized DP-means is robust against the outliers assumed in Section \ref{sec:outlier}.
Although there are many other ways to mix outliers, it is beyond the scope of this paper and needs further study to examine comprehensively the influence of outliers to the clustering performance.

\subsection{Image compression task}
\begin{figure}[htbp]
  \begin{center}
  \includegraphics[width=60mm]{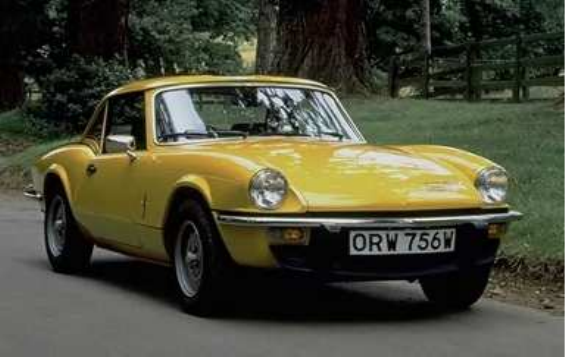}
  \caption{No compression, compression ratio:100\%, number of clusters 1536 \protect\cite{kvq}. \label{fig:original}}
  \end{center}
\vspace{-25pt}
\end{figure}

\begin{figure*}[htbp]
		\begin{center}
		\subfloat[Average distortion minimization, compression rate : 5.61\%, number of clusters 86 \label{fig:lim_min_ave}]{
 			 \includegraphics[width=60mm]{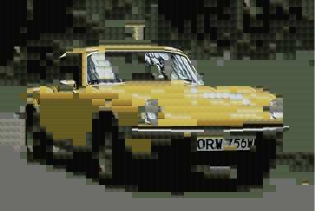}
		}
		\subfloat[Maximum distortion minimization, compression rate : 4.82\%, number of clusters 74 \label{fig:lim_min_max}]{
			  \includegraphics[width=60mm]{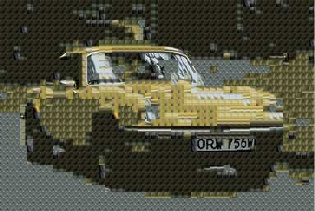}
		}
		\end{center}
\vspace{-10pt}
		\caption{Limited compression that can read the license plate. \label{fig:limitImage} }
\vspace{-10pt}

		\begin{center}
		\subfloat[Average distortion minimization\label{fig:under_lim_min_ave}]{
 			 \includegraphics[width=60mm]{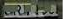}
		}
		\subfloat[Maximum distortion minimization\label{fig:under_lim_min_max}]{
			  \includegraphics[width=60mm]{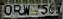}
		}
		\end{center}    
\vspace{-10pt}
		\caption{Compression below limit, compression rate : 3.91\%, number of clusters 60. \label{fig:underLimitImage} }
\vspace{-10pt}
\end{figure*}

We conducted an experiment to see if generalized DP-means is more effective than the original DP-means ($\beta = 1$) through the application of vector quantization to an image compression task.
We used generalized DP-means with power mean \eqref{eq:pow} ($a=0$).
In particular, we considered the case where the minimization measure approaches maximum distortion minimization, that is, $\beta$ is sufficiently large.
Although in this experiment, image compression was handled, the purpose was to examine the performance of the generalized DP-means.
Tipping and Sch{\"o}lkopf compared the maximum distortion minimization and the average distortion minimization in an image compression task by using a clustering method called the kernel vector quantization \cite{kvq}.
In this experiment, the same comparison was carried out using the same image (Figure \ref {fig:original}).
This is a color image size $384 \times 256$.
We obtained data points by dividing it into block images of $8 \times 8$.
Each data point consisted of $8 \times 8 \times 3 = 192$ dimensions from the block size and the color information.
The uncompressed image is represented by the dataset of $384 \times 256/64 = 1536$ points.
Image compression was performed while increasing the penalty parameter and carrying out clustering using Algorithm \ref{extAlg} with $f$ given by \eqref{eq:f_pow} and $\beta=1$ as the average distortion minimization and $\beta=200$ as the approximation of the maximum distortion minimization.
As the penalty parameter increases, the number of clusters decreases.
The preprocessing for clustering was the same as the previous experiment.
We chose the squared distance, $d_\phi(\bm{x}, \bm{\theta})={\lVert \bm{x} - \bm{\theta} \rVert}^2$ as the distance measure.
Newton's method was used for gradient descent optimization to calculate cluster centers because the convergence speed is second order.
The parameter $\beta=200$ was a relatively large value among $\beta$, which did not cause any divergence in the calculation.
Thus, $\beta=200$ was regarded as the maximum distortion minimization.
In this experiment, we focused on whether or not the letter string of the license plate in Figure \ref{fig:original} was recognizable \cite{kvq}.
Figure \ref{fig:limitImage} shows an image compressed to the limit at which the license plate letter string can be read for each of average distortion minimization and maximum distortion minimization.
Further, when the letter string of the license plate can only be read partly, comparison under the same compression ratio is shown in Figure \ref{fig:underLimitImage}.

Figure \ref{fig:limitImage} shows that the maximum distortion minimization achieved the better compression ratio compared to the average distortion minimization when the whole letter string can be read.
In Figure \ref{fig:underLimitImage}, it is possible to read several characters of the license plate in the case of the maximum distortion minimization, while it is almost impossible to read in the case of the average distortion minimization.
In the image compression task focusing on the letter string in the image, it was suggested that better performance was obtained by using the generalized DP-means  with a large value of $\beta$ that approaches the maximum distortion minimization than the original DP-means.
The reason why the generalized objective function with a large $\beta$ was effective may be as follows.
In the average distortion minimization, the license plate consisting of a small number of patterns in the entire image tended to have large distortion from the cluster center to which it belongs, whereas in the case of the maximum distortion minimization, this led to the reduction in the distortion of the blocks from the license plate.

\section{Discussion: total Bregman divergence \label{sec:tbd}}
So far we have discussed with the Bregman divergence as a prerequisite.
The same discussion can be made when the total Bregman divergence is used as the pseudo-distance.
The total Bregman divergence is invariant to the rotation of the coordinate axes, and the cluster center obtained by minimizing the average distortion has been shown to be robust to outliers \cite{total-bregman}.
The Bregman divergence is known to have a bijective relationship with the exponential family, whereas the total Bregman divergence corresponds to the lifted exponential family \cite{lifted_bregman}.
The total Bregman divergence is defined by 
\begin{align*}
{\rm tBD}\autobrac{\bm{x}, \bm{\theta}} \triangleq \frac{d_\phi(\bm{\theta}, \bm{x})}{\sqrt{1+c^2\|\bm{\nabla}\phi(\bm{x})\|^2}}, 
\end{align*}
where $c> 0$ \cite{total-bregman,information-geometry}.
Note that the arguments of $d_\phi$ in the numerator are reversed compared to the $d_\phi\autobrac{\bm{x}, \bm{\theta}}$ in \eqref{eq:def_bregman}.
In the case of $c=1$, it coincides with the definition in \cite{total-bregman}, and when $c=0$, it coincides with the case of the reversed Bregman divergence.
When the total Bregman divergence is used for the pseudo-distance, as in Section \ref{sec:update_rule}, the update rule of the cluster center can be obtained as
\begin{align}
\udrbm{\theta}{k} = \autobrac{\bm{\nabla}\phi}^{-1}\autobrac{\frac{\sum_{i=1}^n w_{ik}\bm{\nabla}\phi(\bm{x}_i)}{\sum_{j=1}^n  w_{jk}}}, \label{eq:tbd_update_rule}
\end{align}
\begin{align}
w_{ik} = \frac{r_{ik}f{\,'}\autobrac{{\rm tBD}\autobrac{\bm{x}_i, \bm{\theta}_k}}}{\sqrt{1+c^2\|\bm{\nabla}\phi(\bm{x}_i)\|^2}}. \label{eq:tbd_weight}
\end{align}
Here, $\autobrac{\bm{\nabla}\phi}^{-1}$ denotes the inverse function of $\bm{\nabla}\phi$.
As in Section \ref{sec:monotonically}, when the function $f$ is concave, Theorem \ref{teiri:tbd_monot} (\ref{sec:tbd_monoto}) holds, claims the monotonic decreasing property of the objective function.
When the function $f$ is convex, since the problem of updating the cluster center is a convex optimization problem, the gradient direction always becomes the descent direction by the gradient descent optimization.
The update rules with Newton's method are summarized in \ref{sec:newton_update}.
The influence function is derived in the same flow as in Section \ref{sec:influence}. As the theorem on the boundedness of the influence function, the following holds.
\begin{theorem}\label{thm:tbd_theorem}
The following Condition 1 on the function $f$ is a necessary and sufficient condition for its influence function to be bounded for all $\bm{x}^*$ and Condition 2 provides a necessary and sufficient condition for it to be redescending:
\begin{enumerate}
  \item $f{\,'}(z)$ is monotonically decreasing function $\iff$ $f(z)$ is a concave function or linear function,
  \item \begin{align}\lim_{z\to\infty}f{\,'}(z)=0 . \label{eq:nec_suf_tbd}\end{align}
\end{enumerate}
\end{theorem}
The proof of this theorem is in \ref{sec:tbd_theorem}.
(See Remark \ref{rmk:infl} for the discussion when $f{\,'}\autobrac{{\rm tBD} \autobrac{\bm{x}^*, \bm{\theta}}}\to\infty$ around $\bm{x}^*=\bm{\theta}$.)
Recall that Condition 2 is a necessary condition for the boundedness of the influence function when the standard Bregman divergence $d_\phi\autobrac{\bm{x}, \bm{\theta}}$ is used as the pseudo-distance (Theorem \ref{th:bound}).

\section{Conclusion \label{sec:conclusion}}
In this paper, we generalized the average distortion of DP-means to $f$-separable distortion measures by using a monotonically increasing function $f$.
If the function $f$ has an inverse function $f^{-1}$, the $f$-separable distortion measure can be expressed by $f$-mean.
We classified the function $f$ into three types, linear, convex and concave.
These three types correspond to the original average distortion, distortion measures approaching the maximum distortion, and those with robustness against outliers respectively.
We showcased two kinds of functions including the parameter $\beta$.
The objective function constituted by these functions can change the characteristics according to the value of the parameter $\beta$.
Furthermore, based on this generalized objective function, an algorithm with guaranteed convergence was constructed. 
Like the original DP-means, this algorithm has the computational complexity of the linear order of the number of data.
In order to evaluate the robustness against outliers, we derived the influence function on the general form of the function $f$ and showed the necessary condition for the influence function to be bounded.
For each concrete example of the function $f$, we examined the condition under which the boundedness of the influence function holds. 
We proved that the log-sum-exp function showed the robustness against outliers regardless of the Bregman divergence.
Although the above discussion assumes the Bregman divergence as pseudo-distance, we also showed that the same argument holds true for the total Bregman divergence.
In addition, experiments using real datasets demonstrated that the generalized DP-means improves the performance of the original DP-means.
Although the generalized DP-means is scalable to high-dimensional data computationally, its clustering performance can be under question if the Bregman divergence is naively defined for example simply additively with respect to dimension.
In such a case, it can be a remedy for high-dimensional data to combine dimension reduction techniques such as non-negative matrix factorization (NMF) \cite{nmf}, $t$-distributed stochastic neighbor embedding ($t$-SNE) \cite{t-sne}, and deep autoencoders \cite{Goodfellow-et-al-2016}.
Our future research will include analysis of the generalization error consisting of the bias and variance in the estimation of the cluster centers.
This will led to a principled design of a combination of the function $f$ and the Bregman divergence (or pseudo-distance) by investigating the trade-off between the generalization error and the robustness.
\section*{Declaration of Competing Interest}{
The authors declare that they have no known competing financial interests or personal relationships that could have appeared to influence the work reported in this paper.
}
\section*{Acknowledgments}{
We are grateful of the anonymous referees for their helpful comments and suggestions. We would like to thank Professor Michael Tipping of Bath University for providing the image used for the experiment. This work was supported in part by JSPS KAKENHI Grant numbers 19J11776 and 19K11825.
}
\appendix
\section{Derivation of objective functions through MAD-Bayes}
\subsection{Generalized Gaussian distribution \label{sec:proof_gene_gauss}}
We focus on the objective function with the function $f$ in \eqref{eq:f_pow} ($\beta>0$, $a=0$) and the Bregman divergence as the squared distance $\|\bm{x}-\bm{\theta}\|^2$.
We prove that the objective function of this case is derived from the framework of MAD-Bayes \cite{mad-bayes} when the generalized Gaussian distribution is assumed as a component.
The generalized Gaussian distribution is given by
\begin{align*}
p(\bm{x}|\bm{\theta}, \alpha, \beta)=\frac{1}{C} \exp\autobrac{-\autobrac{ \frac{\|\bm{x}-\bm{\theta}\|^2}{\alpha}}^\beta},
\end{align*}
where the normalization constant is
\begin{align*}
C = \frac{\beta\Gamma(\frac{L}{2})}{2\pi^{\frac{L}{2}}\Gamma(\frac{L}{\beta})\alpha^{\frac{L}{2}}},
\end{align*}
and $\Gamma(\cdot)$ is the gamma function.
The parameters are $\alpha>0$ and $\beta>0$.
It includes the Laplace distribution ($\beta=\frac{1}{2}$), the Gaussian distribution ($\beta=1$), and the uniform distribution ($\beta\to\infty$) as special cases.
The likelihood is given by
\begin{align*}
p(\upbm{x}{n} | \bm{r}, \autonami{\udrbm{\theta}{k}}_{k=1}^K) = \prod_{k=1}^{K} \prod_{i:r_{ik}=1} p(\bm{x}_i|\bm{\theta}_k, \alpha, \beta).
\end{align*}
The Chinese restaurant process, an example of Dirichlet processes, is given by
\begin{align*}
p(\bm{r}) = \tau^{K-1} \frac{\Gamma(\tau+1)}{\Gamma(\tau+n)} \prod_{k=1}^{K} (S_{n,k}-1)!,
\end{align*}
where $S_{n,k}=\sum_{i=1}^{n}r_{ik}$ and $\tau>0$ is the hyperparameter \cite{non-parabayse}.
When an arbitrary distribution that creates the cluster center is defined as $p(\bm{\theta}_k)$, the simultaneous distribution is expressed by $p(\autonami{\udrbm{\theta}{k}}_{k=1}^K)$.
The simultaneous distribution of data, cluster assignments and cluster centers are expressed by the following equation:
\begin{align*}
\begin{split}
\MoveEqLeft[1]
p(\upbm{x}{n}, \bm{r}, \autonami{\udrbm{\theta}{k}}_{k=1}^K)
= p(\upbm{x}{n} | \bm{r}, \autonami{\udrbm{\theta}{k}}_{k=1}^K)p(\bm{r})p(\autonami{\udrbm{\theta}{k}}_{k=1}^K)
\\={}& \prod_{k=1}^{K} \prod_{i:r_{ik}=1} \frac{1}{C} \exp\autobrac{-\autobrac{ \frac{\|\bm{x}_i-\bm{\theta}_k\|^2}{\alpha}}^\beta}
\cdot{} \tau^{K-1} \frac{\Gamma(\tau+1)}{\Gamma(\tau+n)} \prod_{k=1}^{K} (S_{n,k}-1)!
\cdot{} \prod_{k=1}^K p(\bm{\theta}_k) .
\end{split}
\end{align*}
Then, setting $\tau = \exp\autobrac{-\frac{\lambda^\beta}{\alpha^\beta}}$, we consider the limit $\alpha\to0$.
We have
\begin{align*}
-\ln p(\upbm{x}{n}, \bm{r}, \autonami{\udrbm{\theta}{k}}_{k=1}^K) = \sum_{k=1}^K \sum_{i:r_{ik}=1} \left[ O(\ln(\alpha))+\autobrac{ \frac{\|\bm{x}_i-\bm{\theta}_k\|^{2\beta}}{\alpha^\beta}} \right]
+ (K-1)\frac{\lambda^\beta}{\alpha^\beta} + O(1) .
\end{align*}
It follows that
\begin{align*}
-\alpha^\beta\ln p(\upbm{x}{n}, \bm{r}, \autonami{\udrbm{\theta}{k}}_{k=1}^K) = \sum_{k=1}^K \sum_{i:r_{ik}=1} \|\bm{x}_i-\bm{\theta}_k\|^{2\beta}
+ (K-1)\lambda^\beta + \alpha^\beta O(\ln(\alpha)).
\end{align*}
Because $\alpha^\beta O(\ln(\alpha))\to0$ as $\alpha\to0$, we obtain the objective function as follows:
\begin{align*}
\sum_{k=1}^K \sum_{i:r_{ik}=1} \|\bm{x}_i-\bm{\theta}_k\|^{2\beta}
+ (K-1)\lambda^\beta .
\end{align*}
This objective function is equivalent to the objective function \eqref{eq:generalized_f} with the function $f$ in \eqref{eq:f_pow} ($\beta\neq0$, $a=0$) as follows:
\begin{align*}
\sum_{i=1}^n \|\bm{x}_i-\bm{\theta}_{c(i)}\|^{2\beta}+ \lambda^\beta K +O(1).
\end{align*}
Note, however, that $\beta$ must be positive in the generalized Gaussian distribution.

\subsection{Deformed $t$-distribution \label{sec:proof_t_dist}}
We consider the same objective function as that in \ref{sec:proof_gene_gauss} except that $\beta=0$ and $a>0$ instead of $\beta \neq 0$.
We prove that the objective function of this case is derived from the framework of MAD-Bayes when the deformed $t$-distribution is assumed as a component.
The $t$-distribution is
\begin{align*}
p(\bm{x}|\bm{\theta}, \nu) = \frac{\Gamma(\frac{\nu}{2}+\frac{L}{2})}{\Gamma(\frac{\nu}{2})(\nu\pi)^\frac{L}{2}}\left[1+\frac{\|\bm{x}-\bm{\theta}\|^2}{\nu}\right]^{-\frac{\nu+L}{2}},
\end{align*}
where $\nu>0$ is the degree of freedom.
$L$ is the dimension of data.
It includes the Cauchy distribution ($\nu=1$) and the Gaussian distribution ($\nu\to\infty$) as special cases.
Here, we use the following distribution obtained by transforming this $t$-distribution:
\begin{align*}
p(\bm{x}|\bm{\theta}, \nu, \sigma^2) = \frac{1}{C}\left[1+\frac{\|\bm{x}-\bm{\theta}\|^2}{\nu}\right]^{-\frac{\nu+L}{2\sigma^2}},
\end{align*}
where the normalization constant is 
\begin{align*}
C = \frac{\Gamma(\frac{\nu+L}{2\sigma^2}-\frac{L}{2})(\nu\pi)^\frac{L}{2}}{\Gamma(\frac{\nu+L}{2\sigma^2})}.
\end{align*}
The likelihood is given by
\begin{align*}
p(\upbm{x}{n} | \bm{r}, \autonami{\udrbm{\theta}{k}}_{k=1}^K) = \prod_{k=1}^{K} \prod_{i:r_{ik}=1} p(\bm{x}_i|\bm{\theta}_k, \nu, \sigma^2) .
\end{align*}
As in \ref{sec:proof_gene_gauss}, the simultaneous distribution of data, cluster assignments, and cluster centers are expressed by the following equation:
\begin{align*}
\begin{split}
\MoveEqLeft[1]
p(\upbm{x}{n}, \bm{r}, \autonami{\udrbm{\theta}{k}}_{k=1}^K)
= p(\upbm{x}{n} | \bm{r}, \autonami{\udrbm{\theta}{k}}_{k=1}^K)p(\bm{r})p(\autonami{\udrbm{\theta}{k}}_{k=1}^K)
\\={}&  \prod_{k=1}^{K} \prod_{i:r_{ik}=1} p(\bm{x}_i|\bm{\theta}_k, \nu, \sigma^2)
\cdot{} \tau^{K-1} \frac{\Gamma(\tau+1)}{\Gamma(\tau+n)} \prod_{k=1}^{K} (S_{n,k}-1)!
\cdot{} \prod_{k=1}^K p(\bm{\theta}_k) .
\end{split}
\end{align*}
Then,  setting $\tau = (\nu+\lambda)^{-\frac{\nu+L}{2\sigma^2}}$, we consider the limit $\sigma^2\to0$.
We have
\begin{align*}
\begin{split}
\MoveEqLeft[1]
-\ln p(\upbm{x}{n}, \bm{r}, \autonami{\udrbm{\theta}{k}}_{k=1}^K)
\\={}&  \sum_{k=1}^K \sum_{i:r_{ik}=1} \left[ \ln C +\frac{\nu+L}{2\sigma^2}\ln\autobrac{1+\frac{\|\bm{x}_i-\bm{\theta}_k\|^{2}}{\nu}} \right]
+ \frac{\nu+L}{2\sigma^2}(K-1)\ln(\nu+\lambda) + O(1) .
\end{split}
\end{align*}
It follows that
\begin{align*}
\begin{split}
\MoveEqLeft[1]
-\frac{2\sigma^2}{\nu+L} \ln p(\upbm{x}{n}, \bm{r}, \autonami{\udrbm{\theta}{k}}_{k=1}^K)
\\={}&  \sum_{k=1}^K \sum_{i:r_{ik}=1} \left[ \frac{2\sigma^2}{\nu+L} \ln C +\ln\autobrac{1+\frac{\|\bm{x}_i-\bm{\theta}_k\|^{2}}{\nu}} \right]
+ (K-1)\ln(\nu+\lambda)
\\={}& \sum_{k=1}^K \sum_{i:r_{ik}=1} \left[ \frac{2\sigma^2}{\nu+L} \ln C +\ln\frac{1}{\nu}+ \ln\autobrac{\nu+\|\bm{x}_i-\bm{\theta}_k\|^{2}} \right]
+ (K-1)\ln(\nu+\lambda).
\end{split}
\end{align*}
Because $\sigma^2\ln C\to \rm{constant}$ as $\sigma^2\to0$, we obtain the following objective function:
\begin{align*}
\sum_{k=1}^K \sum_{i:r_{ik}=1} \ln\autobrac{\nu+\|\bm{x}_i-\bm{\theta}_k\|^{2}}
+  (K-1)\ln(\nu+\lambda) .
\end{align*}
We put $\nu=a$. This objective function is equivalent to the objective function \eqref{eq:generalized_f} with the function $f$ in \eqref{eq:f_pow} ($\beta=0$, $a\geq0$) as follows:
\begin{align*}
\sum_{i=1}^n \ln\autobrac{\|\bm{x}_i-\bm{\theta}_{c(i)}\|^{2}+a}+ \ln\autobrac{\lambda+a} K .
\end{align*}
Note, however, that $a$ must not be equal to 0 in the deformed $t$-distribution.

\section{Update rules with Newton's method \label{sec:newton_update}}
The cluster center is updated with Newton's method as follows:
\begin{align}
\bm{\theta}_k = \bm{\theta}_k - \left[\bm{\nabla\nabla}\bar{L}_f(\bm{\theta}_k)\right]^{-1}\bm{\nabla}\bar{L}_f(\bm{\theta}_k), \label{eq:newton_rule}
\end{align}
where the gradient vector and the hessian matrix are given by
\begin{align}
&\bm{\nabla}\bar{L}_f(\bm{\theta}_k) = \sum_{i=1}^n r_{ik}f{\,'}\left(d_\phi \left(\bm{x}_i, \bm{\theta}_k\right)\right) \bm{\nabla}d_\phi\left(\bm{x}_i, \bm{\theta}_k\right), \label{eq:loss_grad}\\
\begin{split}
\MoveEqLeft
\bm{\nabla\nabla}\bar{L}_f(\bm{\theta}_k) = \sum_{i=1}^n r_{ik}f{\,''}\left(d_\phi \left(\bm{x}_i, \bm{\theta}_k\right)\right)\bm{\nabla}d_\phi\left(\bm{x}_i, \bm{\theta}_k\right)\left[\bm{\nabla}d_\phi\left(\bm{x}_i, \bm{\theta}_k\right)\right]^{\rm T}\\
&+\sum_{i=1}^n r_{ik}f{\,'}\left(d_\phi \left(\bm{x}_i, \bm{\theta}_k\right)\right)\bm{\nabla\nabla}d_\phi\left(\bm{x}_i, \bm{\theta}_k\right),
\end{split} \label{eq:loss_hess}
\end{align}
respectively.
The gradient vector and the hessian matrix of the Bregman divergence are given by
\begin{align}
\bm{\nabla}d_\phi\left(\bm{x}, \bm{\theta}\right) = -\bm{\nabla\nabla}\phi(\bm{\theta})\left(\bm{x} - \bm{\theta}\right), \label{eq:breg_grad} \\
\bm{\nabla\nabla}d_\phi\left(\bm{x}, \bm{\theta}\right) = \bm{\nabla\nabla}\phi(\bm{\theta})-\bm{\nabla\nabla\nabla}\phi(\bm{\theta})(\bm{x}-\bm{\theta}) \label{eq:breg_hess} .
\end{align}
If the Bregman divergence is additive with respect to the dimension of data points, $d_\phi(\bm{x}, \bm{\theta}) = \sum_{l=1}^L d_\phi\left(x^{(l)}, \theta^{(l)}\right)
$, the gradient \eqref{eq:breg_grad} and hessian matrix \eqref{eq:breg_hess} are expressed simply.
The $l$-th element of $L$-dimensional column vector \eqref{eq:breg_grad} is given by
\begin{align*}
\frac{\partial d_\phi(\bm{x}, \bm{\theta})}{\partial \theta^{(l)}} = -\phi{\,''}\left(\theta^{(l)}\right)\left(x^{(l)}-\theta^{(l)}\right).
\end{align*}
The hessian matrix \eqref{eq:breg_hess} is diagonal matrix and its $ll$-th element is given by
\begin{align*}
\frac{\partial d_\phi(\bm{x}, \bm{\theta})}{\partial \theta^{(l)}\partial \theta^{(l)}} = -\phi{\,'''}\left(\theta^{(l)}\right)\left(x^{(l)}-\theta^{(l)}\right)+\phi{\,''}\left(\theta^{(l)}\right) .
\end{align*}

In the case of the total Bregman divergence, $d_\phi(\bm{x},\bm{\theta})$ in \eqref{eq:loss_grad} and \eqref{eq:loss_hess} is replaced with ${\rm tBD}(\bm{x}, \bm{\theta})$ and the cluster center is updated by \eqref{eq:newton_rule}.
The gradient vector and the hessian matrix of the total Bregman divergence are given by
\begin{align}
\bm{\nabla}{\rm tBD}(\bm{x}, \bm{\theta}) = \frac{\bm{\nabla}\phi(\bm{\theta})-\bm{\nabla}\phi(\bm{x})}{\sqrt{1+c^2\left\|\bm{\nabla}\phi(\bm{x})\right\|^2}}, \label{eq:tbd_grad} \\
\bm{\nabla\nabla}{\rm tBD}(\bm{x}, \bm{\theta}) = \frac{\bm{\nabla\nabla}\phi(\bm{\theta})}{\sqrt{1+c^2\left\|\bm{\nabla}\phi(\bm{x})\right\|^2}}. \label{eq:tbd_hess}
\end{align}
Similarly, the total Bregman divergence is additive with respect to the dimension of data points, \eqref{eq:tbd_grad} and \eqref{eq:tbd_hess} are expressed simply.
The $l$-th element of $L$-dimensional column vector \eqref{eq:tbd_grad} is given by
\begin{align*}
\frac{\partial {\rm tBD}(\bm{x}, \bm{\theta})}{\partial \theta^{(l)}} =\frac{\phi{\,'}(\theta^{(l)}) - \phi{\,'}(x^{(l)})}{\sqrt{1+c^2\left\|\bm{\nabla}\phi(\bm{x})\right\|^2}}.
\end{align*}
The hessian matrix \eqref{eq:tbd_hess} is diagonal matrix and its $ll$-th element is given by
\begin{align*}
\frac{\partial {\rm tBD}(\bm{x}, \bm{\theta})}{\partial \theta^{(l)}\partial \theta^{(l)}} = \frac{\phi{\,''}(\theta^{(l)})}{\sqrt{1+c^2\left\|\bm{\nabla}\phi(\bm{x})\right\|^2}} .
\end{align*}

\section{Plots of influence functions \label{sec:plot_if}}
\begin{figure*}[htbp]
		\begin{center}
		\subfloat[$\theta=50$, $a=1$, $N=100$]{
 			  \includegraphics[width=6cm]{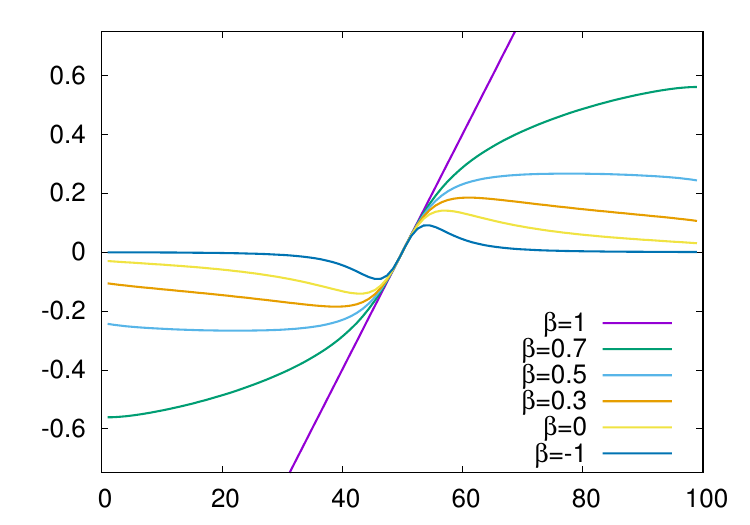}
		}
		\subfloat[$\theta=50$, $a=0$, $N=100$]{
			  \includegraphics[width=6cm]{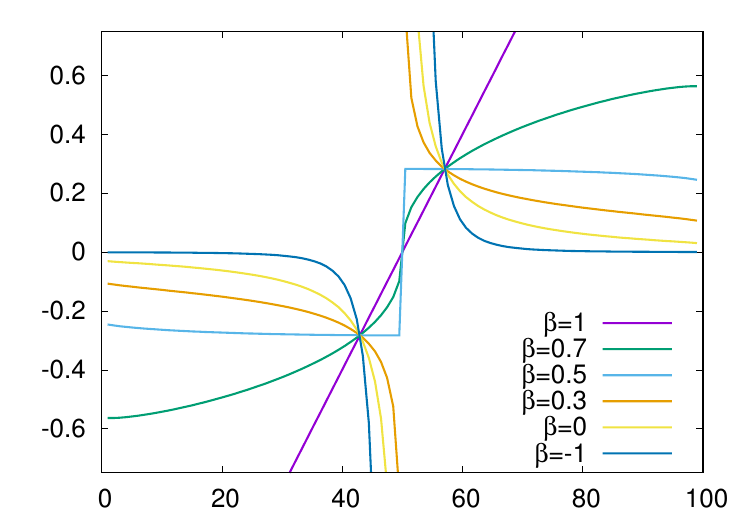}
		}
		\end{center}
\caption{Power mean and binomial-loss. \label{fig:pow_bin}}
		\begin{center}
		\subfloat[$\theta=0$, $a=1$]{
 			  \includegraphics[width=6cm]{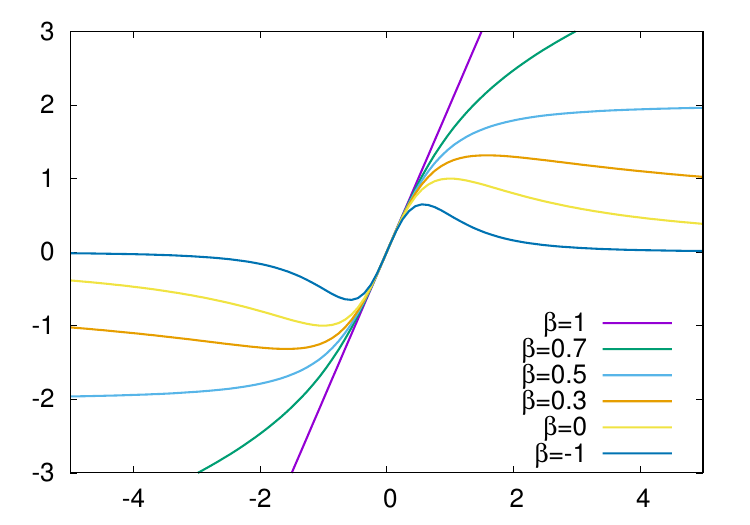}
		}
		\subfloat[$\theta=0$, $a=0$]{
			  \includegraphics[width=6cm]{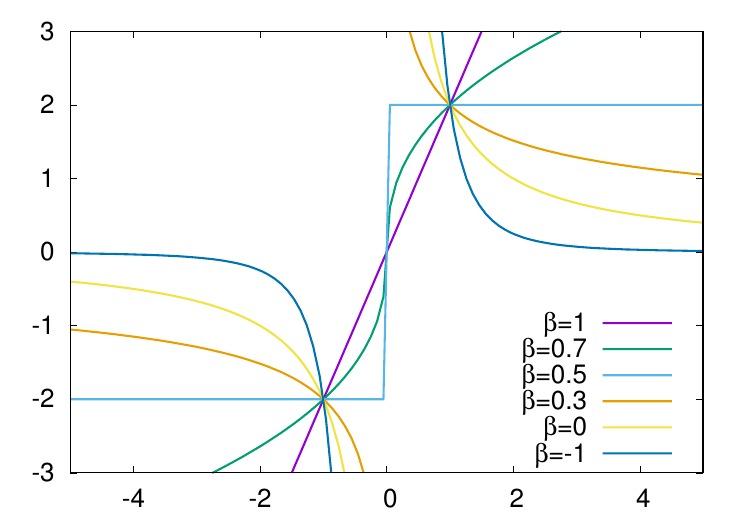}
		}
		\end{center}
\caption{Power mean and squared distance.}
		\begin{center}
		\subfloat[$\theta=100$, $a=1$]{
 			  \includegraphics[width=6cm]{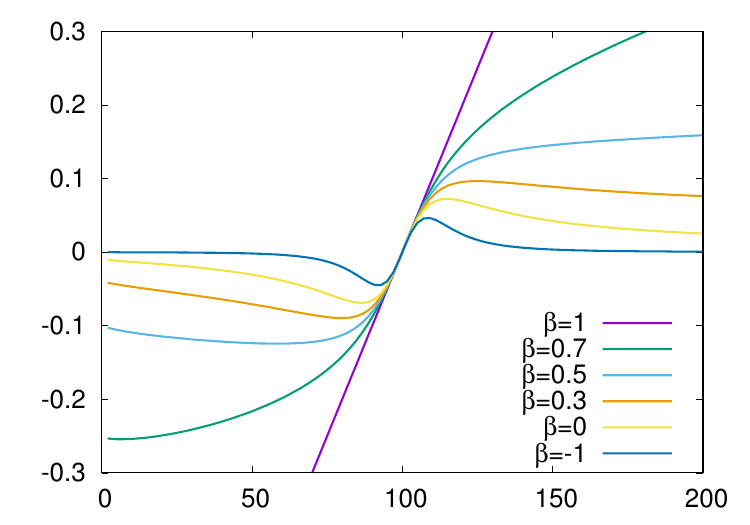}
		}
		\subfloat[$\theta=100$, $a=0$]{
			  \includegraphics[width=6cm]{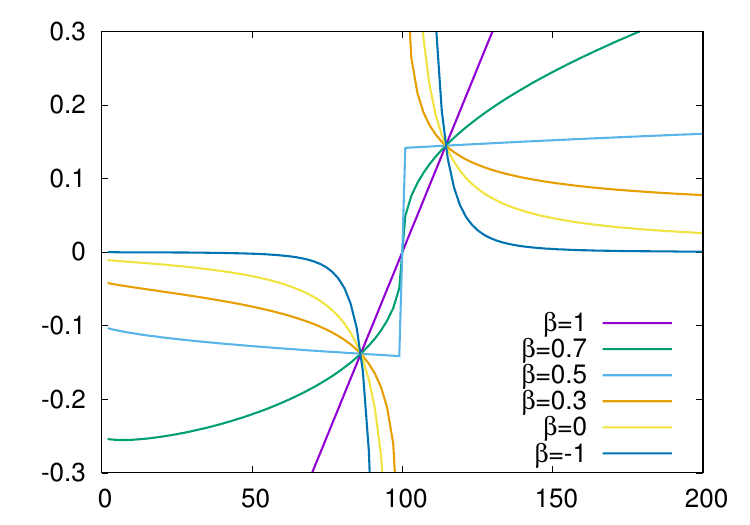}
		}
		\end{center}
\caption{Power mean and generalized KL divergence.}
\end{figure*}

\begin{figure*}[htbp]
		\begin{center}
		\subfloat[Binomial-loss, $\theta=50$, $N=100$]{
 			  \includegraphics[width=6cm]{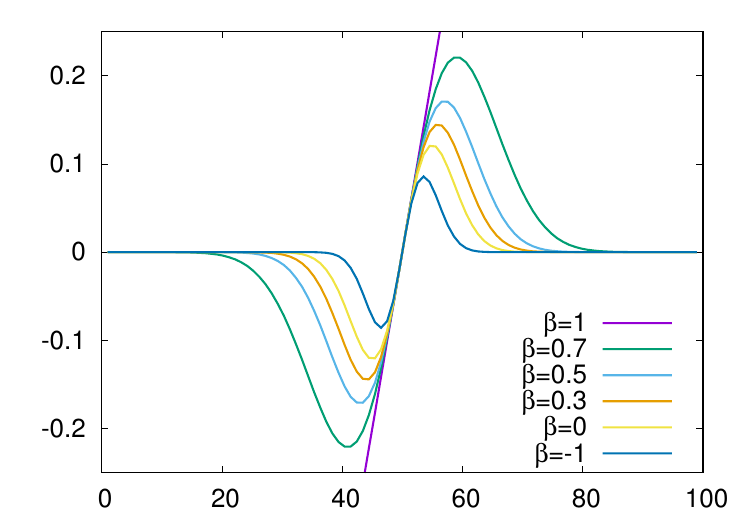}
		}
		\subfloat[Squared distance, $\theta=0$]{
			  \includegraphics[width=6cm]{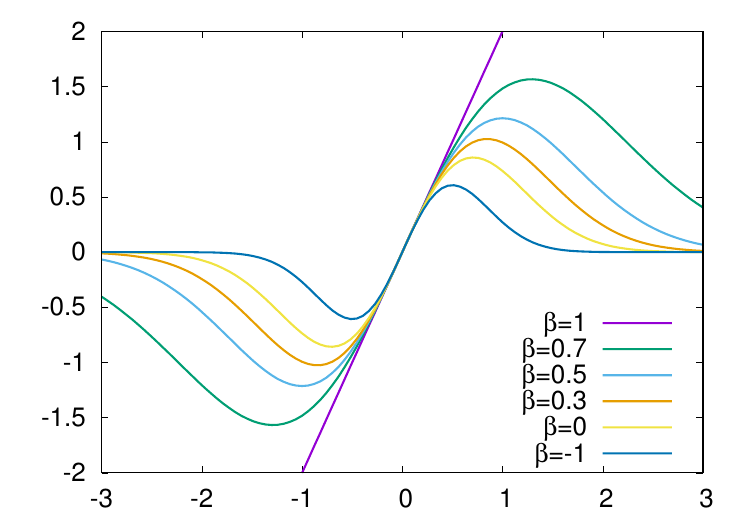}
		}
		\end{center} 
		\begin{center}
		\subfloat[Generalized KL divergence, $\theta=100$]{
 			  \includegraphics[width=6cm]{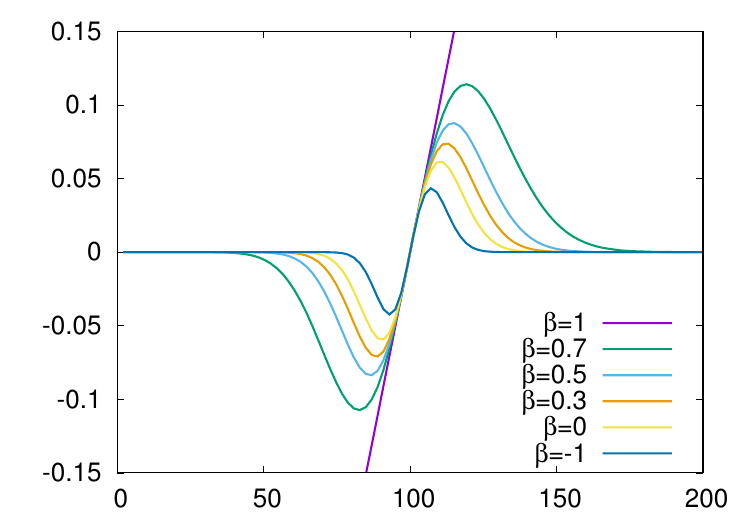}
		}
		\subfloat[Itakua Saito divergence, $\theta=1000$]{
			  \includegraphics[width=6cm]{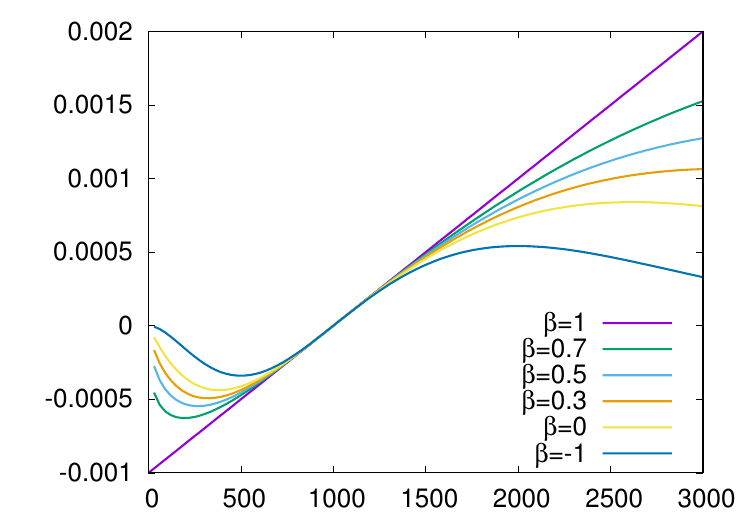}
		}
		\end{center}
		\begin{center}
		\subfloat[Exp-loss, $\theta=0$]{
			  \includegraphics[width=6cm]{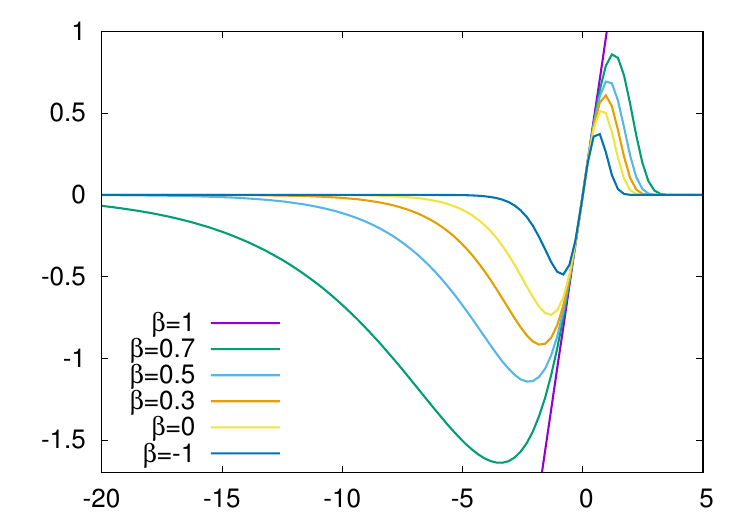}
		}
		\end{center}
\vspace{-10pt}
		\caption{Log-sum-exp. \label{fig:lse_bino}}
\vspace{-10pt}
\end{figure*}

The Bregman divergence corresponding to the binomial distribution is given by
\begin{align}
d_\phi(x,\theta) = x\ln \autobrac{\frac{x}{\theta}}+(N-x)\ln \autobrac{\frac{N-x}{N-\theta}} \label{eq:binomial}
\end{align}
where $N$ is a non-negative integer value and $x\in\{0,1,\ldots,N\}$ in \cite{clustering-bregman}.
In this Section, we call equation \eqref{eq:binomial} ``binomial-loss''.
In the following, \eqref{eq:if} in the one-dimensional case is illustrated as a function of $x^*$ for each Bregman divergence for the power mean and the log-sum-exp (Figure \ref{fig:pow_bin}-\ref{fig:lse_bino}).
It is 0 at $x^*=\theta$.
The tendencies of the influence functions as discussed in Section \ref{sec:pow} and Section \ref{sec:lse} for different $f$ and Bregman divergences can be seen.

\begin{figure*}
		\begin{center}
		\subfloat[$\theta=1000$, $a=1$]{
 			  \includegraphics[width=6cm]{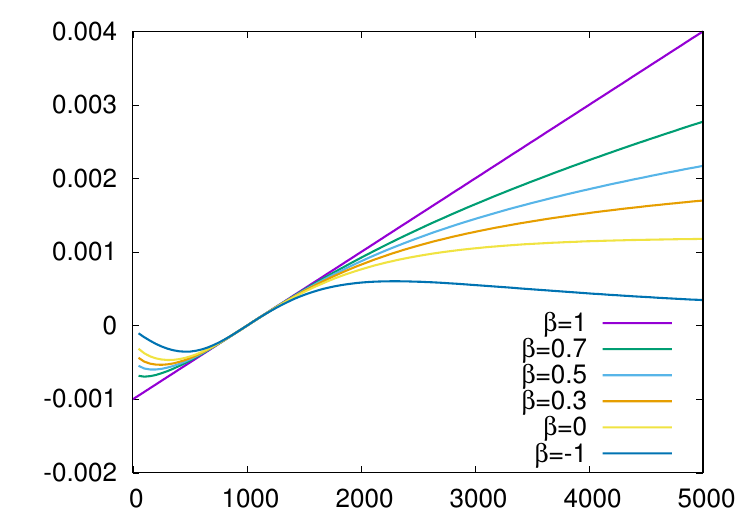}
		}
		\subfloat[$\theta=1000$, $a=0.1$]{
			  \includegraphics[width=6cm]{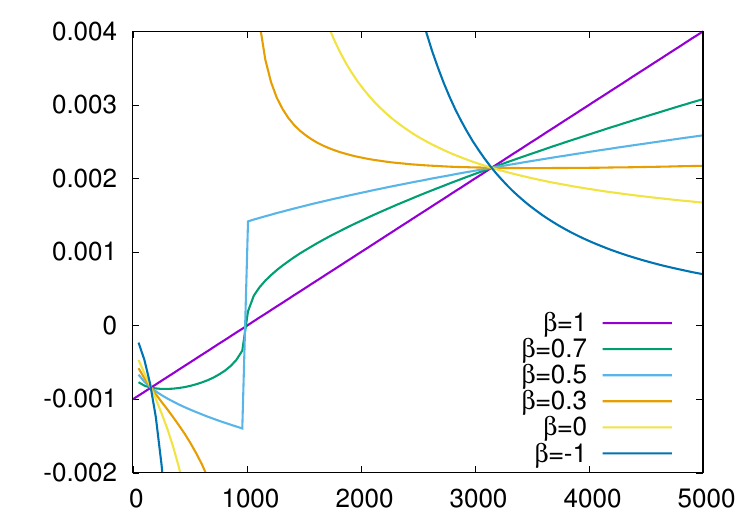}
		}
		\end{center}
\caption{Power mean and Itakura Saito divergence.}
\end{figure*}
\begin{figure*}
		\begin{center}
		\subfloat[$\theta=0$, $a=1$]{
 			  \includegraphics[width=6cm]{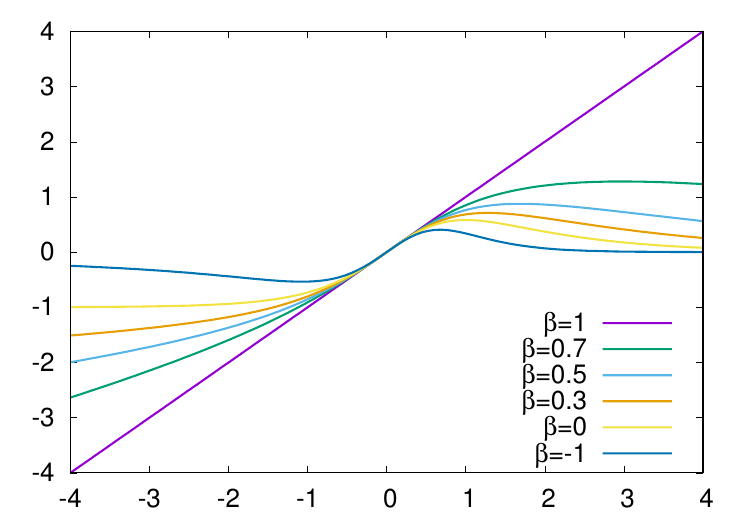}
		}
		\subfloat[$\theta=0$, $a=0$]{
			  \includegraphics[width=6cm]{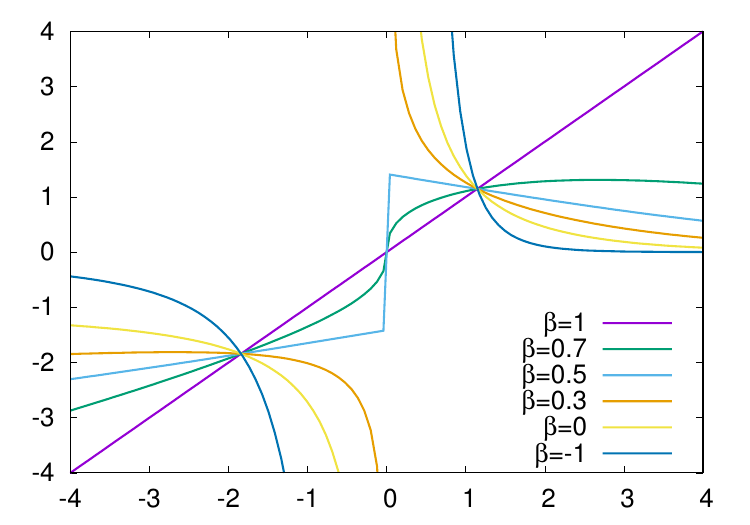}
		}
		\end{center}
\caption{Power mean and exp-loss.}
\end{figure*}

\section{Proof of bounded influence function}
\subsection{Proof of Lemma \ref{lem:lem}}
Under Assumption \ref{asm:asm}, the following lemma holds.
\begin{lemma} \label{lem:lem}
For $\tilde{\bm{x}}=(\theta^{(1)}, \cdots, \theta^{(l-1)}, {x^*}^{(l)}, \theta^{(l+1)}, \cdots, \theta^{(L)})^{\rm T}$, let 
\begin{align}
\tilde{{\rm IF}}_l = \left| \lim_{|{x^*}^{(l)}|\to\infty}  f{\,'}\autobrac{d_\phi \autobrac{\tilde{\bm{x}}, \bm{\theta}}}({x^*}^{(l)}-\theta^{(l)}) \right|, \label{eq:lemma}
\end{align}
be the influence function of the $l$-th dimension.
Then, it holds that
\begin{align*}
\lim_{\|\bm{x}^*\|\to\infty} \|\bm{{\rm IF}}(\upbm{x}{*})\|=
\begin{cases}
    \infty &  \text{{\rm if} $\exists l$, $\tilde{{\rm IF}}_l$ {\rm is divergent}} , \\
    {\rm constant} & \text{{\rm if} $\forall l$, $\tilde{{\rm IF}}_l$ {\rm is bounded}}, \\
    0 & \text{{\rm if} $\forall l$, $\tilde{{\rm IF}}_l$ {\rm is} $0$} .
  \end{cases}
\end{align*}
\end{lemma}

\begin{proof}
Let the $ij$-th component of the matrix $\bm{\nabla}\bm{\nabla}\phi\autobrac{\bm{\theta}}$ be $b_{ij}$, and $\bm{\nabla}\bm{\nabla}\phi\autobrac{\bm{\theta}}(\bm{x}^* - \bm{\theta})=\bm{u}=(u^{(1)}, \cdots, u^{(L)})^{\rm T}\in \mathbb{R}^L$.
It follows that
\begin{align*}
  \left(
    \begin{array}{ccc}
      b_{11} & \ldots & b_{1L} \\
      \vdots & \ddots & \vdots \\
      b_{L1} & \ldots & b_{LL}
    \end{array}
  \right)
  \left(
    \begin{array}{c}
      {x^{*}}^{(1)}-\theta^{(1)} \\
      \vdots \\
      {x^{*}}^{(L)}-\theta^{(L)}    \end{array}
  \right)
=
  \left(
    \begin{array}{c}
      u^{(1)} \\
      \vdots \\
      u^{(L)}    \end{array}
  \right),
\end{align*}
where
\begin{align*}
u^{(j)} = \sum_{l=1}^L b_{jl}({x^{*}}^{(l)}-\theta^{(l)}) .
\end{align*}
If the norm of \eqref{eq:if} is bounded, the norm of the influence function is bounded.
Here, the norm of \eqref{eq:if} is
\begin{align*}
\begin{split}
\MoveEqLeft
\left\|f{\,'}\autobrac{d_\phi \autobrac{\bm{x}^*, \bm{\theta}}} \bm{\nabla}\bm{\nabla}\phi\autobrac{\bm{\theta}}(\bm{x}^* - \bm{\theta})\right\|
= \sqrt{\sum_{j=1}^L \left|f{\,'}\autobrac{d_\phi \autobrac{\bm{x}^*, \bm{\theta}}} u^{(j)} \right|^2} \\
= {}&\sqrt{\sum_{j=1}^L \left|\sum_{l=1}^L b_{jl}  f{\,'}\autobrac{d_\phi \autobrac{\bm{x}^*, \bm{\theta}}} ({x^{*}}^{(l)}-\theta^{(l)})   \right|^2} .
\end{split}
\end{align*}
Hence, we have
\begin{align*}
\lim_{\|\bm{x}^*\|\to\infty} \left\|f{\,'}\autobrac{d_\phi \autobrac{\bm{x}^*, \bm{\theta}}} \bm{\nabla}\bm{\nabla}\phi\autobrac{\bm{\theta}}(\bm{x}^* - \bm{\theta})\right\| \\
=\sqrt{\sum_{j=1}^L \left|\sum_{l=1}^L b_{jl} \lim_{\|\bm{x}^*\|\to\infty} f{\,'}\autobrac{d_\phi \autobrac{\bm{x}^*, \bm{\theta}}} ({x^{*}}^{(l)}-\theta^{(l)})   \right|^2}
\end{align*}
when $\|\bm{x}^*\|\to\infty$.
If the function $f$ satisfies \eqref{eq:limz} of Theorem \ref{th:bound}, which implies the concavity of $f$, the following holds:
\begin{align*}
\left| f{\,'}\autobrac{d_\phi \autobrac{\tilde{\bm{x}}, \bm{\theta}}} ({x^{*}}^{(l)}-\theta^{(l)}) \right|
\geq
\left| f{\,'}\autobrac{d_\phi \autobrac{\bm{x}^*, \bm{\theta}}} ({x^{*}}^{(l)}-\theta^{(l)}) \right| .
\end{align*}
Thus, the bounded and redescending properties of the left hand side as $|{x^*}^{(l)}|\to\infty$ imply those of the right hand side as $\|\bm{x}^*\|\to\infty$, respectively.
This means that if $\tilde{{\rm IF}}_l$ is bounded or converging to 0 for all $l$, so is $\lim_{\|\bm{x}^*\|\to\infty} \|\bm{{\rm IF}}(\upbm{x}{*})\|$.
If $\tilde{{\rm IF}}_l = \infty$ for some $l$, putting $\bm{x}^*=\tilde{\bm{x}}$ and taking the limit $|{x^*}^{(l)}|\to\infty$, we have $\lim_{\|\bm{x}^*\|\to\infty} \|\bm{{\rm IF}}(\upbm{x}{*})\|=\infty$.
\qed
\end{proof}

\subsection{Power mean}
\subsubsection{Proof of Theorem \ref{th:pow}: redescending property for power mean \label{sec:pow_redec}}
Evaluate the following expression \eqref{eq:lemma} of Lemma \ref {lem:lem} for the function $f$ in \eqref{eq:f_pow} ($\beta<0$).
It follows from l'Hopital's rule that
\begin{align*}
\begin{split}
\MoveEqLeft
\tilde{{\rm IF}}_l=
\left| \lim_{|{x^*}^{(l)}| \to \infty} \frac{{x^*}^{(l)} - \theta^{(l)}}{\left[d_\phi \autobrac{\tilde{\bm{x}}, \bm{\theta}}+a\right]^{1-\beta}} \right| \\
= {}& \left| \lim_{|{x^*}^{(l)}| \to \infty} \frac{1}{(1-\beta)\left[d_\phi \autobrac{\tilde{\bm{x}}, \bm{\theta}}+a\right]^{-\beta} \autobrac{\frac{\partial \phi(\tilde{\bm{x}})}{\partial {x^{*}}^{(l)}}-\frac{\partial \phi(\bm{\theta})}{\partial {\theta}^{(l)}}}} \right|= 0 .
\end{split} 
\end{align*}
Therefore, from Lemma \ref{lem:lem}, the redescending property holds.

\subsubsection{$\alpha$-divergence \label{sec:beta_proof}}
Here, since the $\alpha$-divergence is additively defined, it holds that $d_\phi\autobrac{\tilde{\bm{x}}, \bm{\theta}} = d_\alpha({x^{(l)}}^*, \theta^{(l)})$.
Then,the $\alpha$-divergence is expressed as
\begin{align*}
d_\alpha(x^{(l)}, \theta^{(l)}) =
\begin{cases}
x^{(l)} + O({x^{(l)}}^{\alpha}) & \alpha<1,\\
{x^{(l)}}\ln(x^{(l)}) + O(x^{(l)}) & \alpha=1,\\
{x^{(l)}}^\alpha + O(x^{(l)}) & \alpha>1 .
\end{cases}
\end{align*}
Evaluate the following expression \eqref{eq:lemma} of Lemma \ref {lem:lem} for the function $f$ in \eqref{eq:f_pow} :
\begin{align}
\MoveEqLeft
\tilde{{\rm IF}}_l= \left| \lim_{|{x^*}^{(l)}| \to \infty} \frac{{x^*}^{(l)} - \theta^{(l)}}{\left[d_\alpha({x^*}^{(l)}, \theta^{(l)})+a\right]^{1-\beta}} \right|  \nonumber \\
= {}& \left| \left[ \lim_{|{x^*}^{(l)}| \to \infty} \frac{\autobrac{{x^*}^{(l)} - \theta^{(l)}}^{\frac{1}{1-\beta}} } {d_\alpha({x^*}^{(l)}, \theta^{(l)})+a}\right]^{1-\beta}\right|.  \label{eq:beta_eval}
\end{align}

\paragraph{1. $\alpha<1$} $\\$ \noindent
It follows from \eqref{eq:beta_eval} that
\begin{align*}
\MoveEqLeft
\tilde{{\rm IF}}_l=\left|  \left[ \lim_{|{x^*}^{(l)}| \to \infty} \frac{\alpha(\alpha-1) \autobrac{{x^*}^{(l)} - \theta^{(l)}}^{\frac{1}{1-\beta}} } { {{x^*}^{(l)}}+O({{{x^*}^{(l)}}}^\alpha)}\right]^{1-\beta}\right| \\
= {}& \left| \left[ \alpha(\alpha-1) \lim_{|{x^*}^{(l)}| \to \infty} \frac{\autobrac{{x^*}^{(l)} - \theta^{(l)}}^{\frac{1}{1-\beta}} } { {x^*}^{(l)}} \lim_{|{x^*}^{(l)}| \to \infty} \frac{1} {1+ O({{{x^*}^{(l)}}}^{(\alpha-1)})}\right]^{1-\beta}\right| \\
= {}& \left|  \left[ \alpha(\alpha-1) \lim_{|{x^*}^{(l)}| \to \infty} \left[\frac{{x^*}^{(l)} - \theta^{(l)} } { {{x^*}^{(l)}}^{(1-\beta)}}\right]^{\frac{1}{1-\beta}} \right]^{1-\beta}\right| .
\end{align*}
Therefore, it holds that
\begin{align*}
&\begin{cases}
1<1-\beta & \Rightarrow \|\bm{{\rm IF}}(\upbm{x}{*})\| \; {\rm is} \; {\rm divergent}, \\
1=1-\beta & \Rightarrow \|\bm{{\rm IF}}(\upbm{x}{*})\| \; {\rm is} \; {\rm bounded}, \\
1>1-\beta & \Rightarrow \|\bm{{\rm IF}}(\upbm{x}{*})\| \; {\rm is} \; {\rm redescending}, 
\end{cases}
\\ \iff
&\begin{cases}
\beta>0 & \Rightarrow \|\bm{{\rm IF}}(\upbm{x}{*})\| \; {\rm is} \; {\rm divergent}, \\
\beta = 0 & \Rightarrow \|\bm{{\rm IF}}(\upbm{x}{*})\| \; {\rm is} \; {\rm bounded}, \\
\beta <0 & \Rightarrow \|\bm{{\rm IF}}(\upbm{x}{*})\| \; {\rm is} \; {\rm redescending} . 
\end{cases}
\end{align*}

\paragraph{2. $\alpha=1$ (generalized Kullback-Leibler divergence)} $\\$ \noindent
It follows from \eqref{eq:beta_eval} that
\begin{align*}
\MoveEqLeft
\tilde{{\rm IF}}_l=\left|  \left[ \lim_{|{x^*}^{(l)}| \to \infty} \frac{ \autobrac{{x^*}^{(l)} - \theta^{(l)}}^{\frac{1}{1-\beta}} } { {x^{(l)}}^* \ln({x^{(l)}}^*) + O({x^{(l)}}^*)}\right]^{1-\beta}\right| \\
= {}& \left|  \left[ \lim_{|{x^*}^{(l)}| \to \infty} \frac{\autobrac{{x^*}^{(l)} - \theta^{(l)}}^{\frac{1}{1-\beta}} } {  {x^{(l)}}^* \ln({x^{(l)}}^*)} \lim_{|{x^*}^{(l)}| \to \infty} \frac{1} { 1+O(\ln({{x^*}^{(l)}})^{-1})}\right]^{1-\beta}\right| \\
= {}& \left|  \left[ \lim_{|{x^*}^{(l)}| \to \infty} \left[\frac{{x^*}^{(l)} - \theta^{(l)} } {  {{x^{(l)}}^*}^{1-\beta} {\ln({x^{(l)}}^*)}^{1-\beta} }\right]^{\frac{1}{1-\beta}} \right]^{1-\beta}\right| .
\end{align*}
Therefore, it holds that
\begin{align*}
\begin{cases}
\beta >0 & \Rightarrow \|\bm{{\rm IF}}(\upbm{x}{*})\| \; {\rm is} \; {\rm divergent}, \\
\beta \leq 0 & \Rightarrow \|\bm{{\rm IF}}(\upbm{x}{*})\| \; {\rm is} \; {\rm redescending} .
\end{cases}
\end{align*}

\paragraph{3. $\alpha>1$} $\\$ \noindent
It follows from \eqref{eq:beta_eval} that
\begin{align*}
\MoveEqLeft
\tilde{{\rm IF}}_l=\left|  \left[ \lim_{|{x^*}^{(l)}| \to \infty} \frac{\alpha(\alpha-1) \autobrac{{x^*}^{(l)} - \theta^{(l)}}^{\frac{1}{1-\beta}} } { {{x^*}^{(l)}}^\alpha+O({{x^*}^{(l)}})}\right]^{1-\beta}\right| \\
= {}& \left|  \left[ \alpha(\alpha-1) \lim_{|{x^*}^{(l)}| \to \infty} \frac{\autobrac{{x^*}^{(l)} - \theta^{(l)}}^{\frac{1}{1-\beta}} } { {{x^*}^{(l)}}^\alpha} \lim_{|{x^*}^{(l)}| \to \infty} \frac{1} { 1+ O({{{x^*}^{(l)}}}^{(1-\alpha)})}\right]^{1-\beta}\right| \\
= {}& \left|  \left[ \alpha(\alpha-1) \lim_{|{x^*}^{(l)}| \to \infty} \left[\frac{{x^*}^{(l)} - \theta^{(l)} } { {{x^*}^{(l)}}^{\alpha(1-\beta)}}\right]^{\frac{1}{1-\beta}} \right]^{1-\beta}\right| .
\end{align*}
Therefore, it holds that
\begin{align*}
&\begin{cases}
\alpha(1-\beta)<1 & \Rightarrow \|\bm{{\rm IF}}(\upbm{x}{*})\| \; {\rm is} \; {\rm divergent}, \\
\alpha(1-\beta)=1 & \Rightarrow \|\bm{{\rm IF}}(\upbm{x}{*})\| \; {\rm is} \; {\rm bounded}, \\
\alpha(1-\beta)>1 & \Rightarrow \|\bm{{\rm IF}}(\upbm{x}{*})\| \; {\rm is} \; {\rm redescending}, 
\end{cases}
\\ \iff
&\begin{cases}
\beta>1-\frac{1}{\alpha} & \Rightarrow \|\bm{{\rm IF}}(\upbm{x}{*})\| \; {\rm is} \; {\rm divergent}, \\
\beta = 1-\frac{1}{\alpha} &\Rightarrow \|\bm{{\rm IF}}(\upbm{x}{*})\| \; {\rm is} \; {\rm bounded}, \\
\beta <1-\frac{1}{\alpha} & \Rightarrow \|\bm{{\rm IF}}(\upbm{x}{*})\| \; {\rm is} \; {\rm redescending} . 
\end{cases}
\end{align*}

\subsubsection{Exp-loss \label{sec:exp-loss}}
When the function $f$ is given by \eqref{eq:f_pow} and the convex function constituting the Bregman divergence is given by the exponential function, we investigate the boundedness of the influence function.
For the convex function
\begin{align*}
\phi(x)=\exp(x),
\end{align*}
the corresponding Bregman divergence is given by \cite{clustering-bregman},
\begin{align*}
d_\phi\autobrac{x, \theta} = \exp(x)-\exp(\theta)-\exp(\theta)(x-\theta) .
\end{align*}
For multidimensional data, we additively define the divergence as follows:
\begin{align}
\MoveEqLeft
\tilde{{\rm IF}}_l=\left| \lim_{|{x^*}^{(l)}| \to \infty} \frac{{x^*}^{(l)} - \theta^{(l)}}{\left[d_\phi\autobrac{\tilde{\bm{x}}, \bm{\theta}}+a\right]^{1-\beta}} \right| \nonumber \\
= {}& \left| \lim_{|{x^*}^{(l)}| \to \infty} \frac{\left[d_\phi\autobrac{\tilde{\bm{x}}, \bm{\theta}}+a\right]^{\beta}}{(1-\beta) \autobrac{\exp({x^{*}}^{(l)})-\exp({\theta}^{(l)})}} \right| \label{eq:exp_bregman}\\
= {}& \left| \lim_{{x^*}^{(l)} \to \infty} \frac{\beta\left[d_\phi\autobrac{\tilde{\bm{x}}, \bm{\theta}}+a\right]^{\beta-1}\autobrac{\exp({x^{*}}^{(l)})-\exp({\theta}^{(l)})}}{(1-\beta) \exp({x^{*}}^{(l)})} \right| \nonumber \\
= {}& \left| \lim_{{x^*}^{(l)} \to \infty} \frac{\beta}{(1-\beta)\left[d_\phi\autobrac{\tilde{\bm{x}}, \bm{\theta}}+a\right]^{1-\beta}} 
  - \lim_{{x^*}^{(l)} \to \infty} \frac{\beta \exp({\theta}^{(l)})}{(1-\beta)\left[d_\phi\autobrac{\tilde{\bm{x}}, \bm{\theta}}+a\right]^{1-\beta}\exp({x^{*}}^{(l)})} \right| \label{eq:exp_bregman0} .
\end{align}
When $\beta=0$, it follows from \eqref{eq:exp_bregman} that:
\begin{align*}
\tilde{{\rm IF}}_l =\frac{1}{\exp({x^{*}}^{(l)})-\exp({\theta}^{(l)})} 
=\begin{cases}
0 & {\rm if} \; {x^*}^{(l)} \to \infty, \\
-\exp(-{\theta}^{(l)}) & {\rm if} \; {x^*}^{(l)} \to -\infty . \\
\end{cases}
\end{align*}
When $0<\beta<1$, if the $l$-th elements of $\bm {x}^*$ satisfies ${x^*}^{(l)}\to-\infty$ then, $\tilde{{\rm IF}}_l\to\infty$ from \eqref{eq:exp_bregman}.
On the other hand, if the $l$-th elements of $\bm {x}^*$ satisfies ${x^*}^{(l)}\to\infty$ then, $\tilde{{\rm IF}}_l=0$ from \eqref{eq:exp_bregman0}.
That is, when $0<\beta<1$, it follows that:
\begin{align*}
\tilde{{\rm IF}}_l
=\begin{cases}
0 & {\rm if} \; {x^*}^{(l)} \to \infty, \\
\infty & {\rm if} \; {x^*}^{(l)} \to -\infty . \\
\end{cases}
\end{align*}
When $\beta<0$, $\tilde{{\rm IF}}_l$ is 0 from \eqref{eq:exp_bregman}.
The results are summarized as follows:
\begin{align*}
\begin{cases}
\beta>0 & \Rightarrow \|\bm{{\rm IF}}(\upbm{x}{*})\| \; {\rm is} \; {\rm divergent}, \\
\beta=0 & \Rightarrow \|\bm{{\rm IF}}(\upbm{x}{*})\| \; {\rm is} \; {\rm bounded}, \\
\beta<0 & \Rightarrow \|\bm{{\rm IF}}(\upbm{x}{*})\| \; {\rm is} \; {\rm redescending} . \\
\end{cases}
\end{align*}

\subsection{Proof of Theorem \ref{th:lse}: redscending property for log-sum-exp \label{sec:lse_redec}}
Evaluate the following expression \eqref{eq:lemma} of Lemma \ref {lem:lem} for the function $f$ in \eqref{eq:f_exp} ($\beta<1$).
It follows from l'Hopital's rule that
\begin{align*}
\begin{split}
\MoveEqLeft
\tilde{{\rm IF}}_l=
\left| \lim_{|{x^*}^{(l)}| \to \infty} \frac{{{x^*}^{(l)} - \theta^{(l)}}}{\exp\left( (1-\beta)d_\phi \autobrac{\tilde{\bm{x}}, \bm{\theta}}\right)}  \right| \\
= {}& \left| \lim_{|{x^*}^{(l)}| \to \infty} \frac{1}{(1-\beta)\exp\left((1-\beta)d_\phi \autobrac{\tilde{\bm{x}}, \bm{\theta}}\right) \autobrac{\frac{\partial \phi(\tilde{\bm{x}})}{\partial {x^{*}}^{(l)}}-\frac{\partial \phi(\bm{\theta})}{\partial {\theta}^{(l)}}}}  \right| = 0 .
\end{split} 
\end{align*}
Therefore, from Lemma \ref{lem:lem}, the redescending property holds.

\section{Properties of total Bregman divergence}
\subsection{Guarantee of monotonic decreasing property \label{sec:tbd_monoto}}
\begin{theorem}\label{teiri:tbd_monot}
If the function $f$ is concave, the updating of the cluster center using \eqref{eq:tbd_update_rule} monotonically decreases the objective function for general total Bregman divergence.
\end{theorem}
\begin{proof}
The flow of the proof is almost the same as the proof of Theorem \ref{teiri:monot}.
We show that the objective function \eqref{eq:generalized_f} monotonically decreases when the $k$-th cluster center $\udrbm{\theta}{k}$ is newly updated to $\tilde{\bm{\theta}}_k$ by \eqref{eq:tbd_update_rule}.
More specifically, we prove that, $\bar{L}_f(\udrbm{\theta}{k})\geq \bar{L}_f(\tilde{\bm{\theta}}_k) $,  where $\bar{L}_f(\udrbm{\theta}{k})$ is the sum of the terms related to $\udrbm{\theta}{k}$ in \eqref{eq:generalized_f}, that is,
\begin{align*}
\bar{L}_f({\bm{\theta}}_k) = \sum_{i=1}^n r_{ik} f\autobrac{{\rm tBD}\autobrac{\bm{x}_i, \bm{\theta}_k}} .
\end{align*}
From the inequality in \eqref{eq:tangent_inquality}, the following holds:
\begin{align}
\MoveEqLeft[1]
\bar{L}_f(\udrbm{\theta}{k})-\bar{L}_f(\tilde{\bm{\theta}}_k) \nonumber
\\={}& \sum_{i=1}^n r_{ik} \left[f\autobrac{{\rm tBD}\autobrac{\bm{x}_i, \bm{\theta}_k}}-f\autobrac{{\rm tBD}\autobrac{\bm{x}_i, \tilde{\bm{\theta}}_k}}\right]  \nonumber
\\ \geq{}& \sum_{i=1}^n r_{ik} f{\,'}\autobrac{{\rm tBD}\autobrac{\bm{x}_i, \bm{\theta}_k}} \left[{\rm tBD}\autobrac{\bm{x}_i, \bm{\theta}_k} - {\rm tBD}\autobrac{\bm{x}_i, \tilde{\bm{\theta}}_k}\right] \nonumber
\\={}& \sum_{i=1}^n r_{ik} \frac{f{\,'}\autobrac{{\rm tBD}\autobrac{\bm{x}_i, \bm{\theta}_k}}}{\sqrt{1+c^2\|\bm{\nabla}\phi(\bm{x}_i)\|^2}}\left[d_\phi\left(\bm{\theta}_k, \bm{x}_i\right) - d_\phi\left(\tilde{\bm{\theta}}_k, \bm{x}_i \right) \right] \nonumber
\\={}& \sum_{i=1}^n r_{ik} \frac{f{\,'}\autobrac{{\rm tBD}\autobrac{\bm{x}_i, \bm{\theta}_k}}}{\sqrt{1+c^2\|\bm{\nabla}\phi(\bm{x}_i)\|^2}}\left[ \phi\left(\bm{\theta}_k\right) - \phi\left(\tilde{\bm{\theta}}_k\right) - \bm{\nabla}\phi\left(\bm{x}_i\right) \left(\bm{\theta}_k - \tilde{\bm{\theta}}_k\right) \right] \label{eq:use_update_tbd}
\\={}& d_\phi \autobrac{\bm{\theta}_k, \tilde{\bm{\theta}}_k} \sum_{i=1}^n r_{ik} \frac{f{\,'}\autobrac{{\rm tBD}\autobrac{\bm{x}_i, \bm{\theta}_k}}}{\sqrt{1+c^2\|\bm{\nabla}\phi(\bm{x}_i)\|^2}} \geq 0 , \nonumber
\end{align}
where 
\begin{align*}
\bm{\nabla}\phi\left(\tilde{\bm{\theta}}_k \right)\sum_{i=1}^n r_{ik} \frac{f{\,'}\autobrac{{\rm tBD}\autobrac{\bm{x}_i, \bm{\theta}_k}}}{\sqrt{1+c^2\|\bm{\nabla}\phi(\bm{x}_i)\|^2}} = \sum_{i=1}^n r_{ik} \frac{f{\,'}\autobrac{{\rm tBD}\autobrac{\bm{x}_i, \bm{\theta}_k}}}{\sqrt{1+c^2\|\bm{\nabla}\phi(\bm{x}_i)\|^2}}\bm{\nabla}\phi\left(\bm{x}_i\right) ,
\end{align*}
which is derived from \eqref{eq:tbd_update_rule} and \eqref{eq:tbd_weight} was used in \eqref{eq:use_update_tbd}.
\qed
\end{proof}
The following corollary immediately follows from Theorem \ref{teiri:tbd_monot}.
\begin{corollary}
When the objective function is constructed by the power mean \eqref{eq:pow} or the log-sum-exp function \eqref{eq:lse}, the following holds.
When $\beta\leq1$, the updating of the cluster center  using \eqref{eq:tbd_update_rule} monotonically decreases the objective function for general total Bregman divergence.
\end{corollary}

\subsection{Influence function}
When we derive the influence function as in Section \ref{sec:influence}, it is given by
\begin{align*}
&\bm{{\rm IF}}(\upbm{x}{*}) = - m\bm{{\rm G}}^{-1}\bm{\nabla}f\autobrac{{\rm tBD} \autobrac{\bm{x}^*, \bm{\theta}}} ,  \\
&\bm{{\rm G}} = \sum_{i=1}^m \bm{\nabla}\bm{\nabla}f\autobrac{{\rm tBD}\autobrac{\bm{x}_i, \bm{\theta}}}.
\end{align*}
Because the matrix $\bm{{\rm G}}$ does not depend on $\upbm{x}{*}$, the robustness against outliers is evaluated by
\begin{align}
\begin{split}
\MoveEqLeft
-\bm{\nabla}f\autobrac{{\rm tBD} \autobrac{\bm{x}^*, \bm{\theta}}} = -f{\,'}\autobrac{{\rm tBD} \autobrac{\bm{x}^*, \bm{\theta}}} \bm{\nabla}{\rm tBD} \autobrac{\bm{x}^*, \bm{\theta}} \\
= {}&f{\,'}\autobrac{{\rm tBD} \autobrac{\bm{x}^*, \bm{\theta}}} \frac{\bm{\nabla}\phi(\bm{x}^*)-\bm{\nabla}\phi(\bm{\theta})}{\sqrt{1+c^2\|\bm{\nabla}\phi(\bm{x}^*)\|^2}}. \\
\end{split} \label{eq:if_tbd}
\end{align}

\subsection{Proof of Theorem \ref{thm:tbd_theorem} \label{sec:tbd_theorem}}
Evaluate the following expression, which is the norm of \eqref{eq:if_tbd}:
\begin{align}
f{\,'}\autobrac{{\rm tBD} \autobrac{\bm{x}^*, \bm{\theta}}} \frac{\| \bm{\nabla}\phi(\bm{x}^*)-\bm{\nabla}\phi(\bm{\theta})\| }{\sqrt{1+c^2\|\bm{\nabla}\phi(\bm{x}^*)\|^2}}. \label{eq:norm_total_breg}
\end{align}
Even if $\|\bm{x}^*\|$ has any value, $\frac{\| \bm{\nabla}\phi(\bm{x}^*)-\bm{\nabla}\phi(\bm{\theta})\| }{\sqrt{1+c^2\|\bm{\nabla}\phi(\bm{x}^*)\|^2}}$ it is bounded \cite{total-bregman}.
Therefore, it is a necessary and sufficient condition for the influence function to be bounded that $f'(z)$ is bounded for all $z$.
As $\|\bm{\theta}\|<\infty$, when $\|\bm{x}^*\|$ is large, ${\rm tBD} \autobrac{\bm{x}^*, \bm{\theta}}$ is large.
When the function $f(z)$ is convex, $f{\,'}(z)$ is a monotonically increasing function.
As $\|\bm{x}^*\|$ is increased, \eqref{eq:norm_total_breg} grows unboundedly, and the influence function is not bounded.
In order to reduce \eqref{eq:norm_total_breg} when $\|\bm{x}^*\|$ is large, $f{\,'}(z)$ must be a monotonically decreasing function.
The function $f(z)$ that satisfies this condition is concave or linear (Condition 1).
As $\|\bm{\theta}\|<\infty$, when $\|\bm{x}^*\|\to\infty$, $\frac{\| \bm{\nabla}\phi(\bm{x}^*)-\bm{\nabla}\phi(\bm{\theta})\| }{\sqrt{1+c^2\|\bm{\nabla}\phi(\bm{x}^*)\|^2}}$ does not become 0.
Therefore, the necessary and sufficient condition for satisfying the redescending property \eqref{eq:redescending} is \eqref{eq:nec_suf_tbd} (Condition 2).



\bibliographystyle{elsarticle-num} 


\bibliography{Refj}

\begin{thebibliography}{10}
\expandafter\ifx\csname url\endcsname\relax
  \def\url#1{\texttt{#1}}\fi
\expandafter\ifx\csname urlprefix\endcsname\relax\def\urlprefix{URL }\fi
\expandafter\ifx\csname href\endcsname\relax
  \def\href#1#2{#2} \def\path#1{#1}\fi

\bibitem{affinity}
B.~J. Frey, D.~Dueck, Clustering by passing messages between data points,
  Science 315~(5814) (2007) 972--976.

\bibitem{mean_shift}
Y.~Cheng, Mean shift, mode seeking, and clustering, IEEE Transactions on
  Pattern Analysis and Machine Intelligence 17~(8) (1995) 790--799.

\bibitem{gamma-clust}
A.~Notsu, S.~Eguchi, Robust clustering method in the presence of scattered
  observations, Neural Computation 28~(6) (2016) 1141--1162.

\bibitem{dpmeans}
B.~Kulis, M.~I. Jordan, Revisiting k-means: New algorithms via {B}ayesian
  nonparametric, in: Proceedings of the 29th {I}nternational {C}onference on
  {M}achine {L}earning ({ICML}), 2012, pp. 513--520.

\bibitem{non-parabayse}
S.~J. Gershman, D.~M. Blei, A tutorial on {B}ayesian nonparametric models,
  Journal of Mathematical Psychology 56~(1) (2012) 1--12.

\bibitem{occ}
X.~Pan, J.~E. Gonzalez, S.~Jegelka, T.~Broderick, M.~I. Jordan, Optimistic
  concurrency control for distributed unsupervised learning, in: {A}dvances in
  {N}eural {I}nformation {P}rocessing {S}ystems 26 ({NIPS}), 2013, pp.
  1403--1411.

\bibitem{coreset-dpmeans}
O.~Bachem, M.~Lucic, A.~Krause, Coresets for nonparametric estimation - the
  case of {DP}-means, in: Proceedings of the 32th {I}nternational {C}onference
  on {M}achine {L}earning ({ICML}), 2015, pp. 209--217.

\bibitem{dace}
L.~Jiang, Y.~Dong, N.~Chen, T.~Chen, {D}{A}{C}{E}: a scalable {D}{P}-means
  algorithm for clustering extremely large sequence data, Bioinformatics 33~(6)
  (2017) 834--842.

\bibitem{split-dp}
S.~Odashima, M.~Ueki, N.~Sawasaki, A split-merge {D}{P}-means algorithm to
  avoid local minima, in: Proceedings of the {E}uropean {C}onference on
  {M}achine {L}earning and {P}rinciples and {P}ractice of {K}nowledge
  {D}iscovery in {D}atabases ({ECML} {PKDD}), 2016, pp. 63--78.

\bibitem{clustering-bregman}
A.~Banerjee, S.~Merugu, I.~S. Dhillon, J.~Ghosh, Clustering with {B}regman
  divergences, Journal of Machine Learning Research 6~(Oct) (2005) 1705--1749.

\bibitem{dp-bregman}
K.~Jiang, B.~Kulis, M.~I. Jordan, Small-variance asymptotics for exponential
  family {D}irichlet process mixture models, in: {A}dvances in {N}eural
  {I}nformation {P}rocessing {S}ystems 25 ({NIPS}), 2012, pp. 3158--3166.

\bibitem{mypaper1}
M.~Kobayashi, K.~Watanabe, A rate-distortion theoretic view of {D}irichlet
  process means clustering, IEICE Transactions on Fundamentals J100-A~(12)
  (2017) 475--486, in Japanese.

\bibitem{k-center}
T.~F. Gonzalez, Clustering to minimize the maximum intercluster distance,
  Theoretical Computer Science 38 (1985) 293--306.

\bibitem{seb}
M.~B\u{a}doiu, K.~L. Clarkson, Smaller core-sets for balls, in: Proceedings of
  the 14th {A}nnual {ACM-SIAM} {S}ymposium on {D}iscrete {A}lgorithms ({SODA}),
  2003, pp. 801--802.

\bibitem{bregman-ball}
R.~Nock, F.~Nielsen, Fitting the smallest enclosing {B}regman ball, in:
  Proceedings of the 16th {E}uropean {C}onference on {M}achine {L}earning
  ({ECML}), 2005, pp. 649--656.

\bibitem{mypaper2}
M.~Kobayashi, K.~Watanabe, Generalized {D}irichlet-process-means for robust and
  maximum distortion criteria, in: Proceedings of {I}nternational {S}ymposium
  on {I}nformation {T}heory and {I}ts {A}pplications ({ISITA}), 2018, pp.
  45--49.

\bibitem{f-separable}
Y.~Shkel, S.~Verd\'u, A coding theorem for f-separable distortion measures,
  Entropy 20~(2) (2018).

\bibitem{f-mean}
V.~M. Tikhomirov, On the notion of mean, in: Selected Works of {A}. {N}.
  {K}olmogorov, Netherlands:Springer, 1991, pp. 144--146.

\bibitem{psi-div}
S.~Eguchi, Y.~Kano, Robustifing maximum likelihood estimation by
  psi-divergence, Tech. rep. (2001).

\bibitem{tsallis_book}
C.~Tsallis, Introduction to nonextensive statistical mechanics, New York:
  Springer, 2009.

\bibitem{mad-bayes}
T.~Broderick, B.~Kulis, M.~I. Jordan, {M}{A}{D}-{B}ayes: {M}{A}{P}-based
  asymptotic derivations from {B}ayes, in: Proceedings of the 30th
  {I}nternational {C}onference on {M}achine {L}earning ({ICML}), 2013, pp.
  226--234.

\bibitem{convex}
S.~Boyd, L.~Vandenberghe, Convex optimization, {C}ambridge: {C}ambridge
  {U}niversity {P}ress, 2004.

\bibitem{entropic_risk}
K.~Watanabe, S.~Ikeda, Entropic risk minimization for nonparametric estimation
  of mixing distributions, Machine Learning 99~(1) (2015) 119--136.

\bibitem{ei-means}
K.~Watanabe, Vector quantization based on $\varepsilon$-insensitive mixture
  models, Neurocomputing 165 (2015) 32--37.

\bibitem{mm-alg}
D.~R. Hunter, K.~Lange, A tutorial on {MM} algorithms, The American
  Statistician 58~(1) (2004) 30--37.

\bibitem{total-bregman}
B.~C. Vemuri, M.~Liu, S.~Amari, F.~Nielsen, Total {B}regman divergence and its
  applications to {D}{T}{I} analysis, IEEE Transactions on Medical Imaging
  30~(2) (2011) 475--483.

\bibitem{information-geometry}
S.~Amari, Robust cluster center, in: Information geometry and its applications,
  Japan: Springer, 2016, pp. 238--240.

\bibitem{robust}
F.~R. Hampel, E.~M. Ronchetti, P.~J. Rousseeuw, W.~A. Stahel, Robust
  statistics: the approach based on influence functions, John Wiley \& Sons,
  2005.

\bibitem{beta-divergence}
R.~Hennequin, B.~David, R.~Badeau, Beta-divergence as a subclass of {B}regman
  divergence, IEEE Signal Processing Letters 18~(2) (2011) 83--86.

\bibitem{kvq}
M.~Tipping, B.~Sch{\"o}lkopf, A kernel approach for vector quantization with
  guaranteed distortion bounds, in: Proceedings of {A}rtificial {I}ntelligence
  and {S}tatistics ({AISTATS}), 2001, pp. 129--134.

\bibitem{lifted_bregman}
M.~Liu, B.~C. Vemuri, S.~Amari, F.~Nielsen, Shape retrieval using hierarchical
  total {B}regman soft clustering, IEEE Transactions on Pattern Analysis and
  Machine Intelligence 34~(12) (2012) 2407--2419.

\bibitem{nmf}
A.~Cichocki, R.~Zdunek, A.~H. Phan, S.~Amari, Nonnegative Matrix and Tensor
  Factorizations: Applications to Exploratory Multi-way Data Analysis and Blind
  Source Separation, John Wiley \& Sons, 2009.

\bibitem{t-sne}
L.~van~der Maaten, G.~E. Hinton, Visualizing data using t-{SNE}, Journal of
  Machine Learning Research 9~(Nov) (2008) 2759--2605.

\bibitem{Goodfellow-et-al-2016}
I.~Goodfellow, Y.~Bengio, A.~Courville, Deep Learning, MIT Press, 2016.

\end{thebibliography}



\end{document}